\documentclass{article}
\usepackage{array}

\usepackage[final]{corl_2026} 
\usepackage{graphicx}
\usepackage{booktabs}
\usepackage{enumitem}
\usepackage{prettyref}
\newrefformat{fig}{Figure~\ref{#1}}
\newrefformat{tab}{Table~\ref{#1}}
\newrefformat{sec}{Section~\ref{#1}}
\newrefformat{eq}{Equation~\ref{#1}}
\newrefformat{appen}{Appendix~\ref{#1}}
\usepackage{amsmath}
\usepackage{amssymb}
\usepackage{wrapfig}
\usepackage{titletoc}

\usepackage{xcolor}
\usepackage[most]{tcolorbox}

\usepackage{xcolor}
\usepackage[most]{tcolorbox}

\definecolor{apiBlue}{RGB}{30,80,150}
\definecolor{apiGray}{RGB}{245,245,245}
\usepackage{pifont}

\newtcolorbox{apibox}[1]{
  colback=apiGray,
  colframe=apiBlue,
  coltitle=white,
  colbacktitle=apiBlue,
  title=\texttt{#1},
  fonttitle=\bfseries,
  boxrule=0.6pt,
  arc=2pt,
  left=5pt,
  right=5pt,
  top=4pt,
  bottom=4pt,
  before skip=0.6em,
  after skip=0.6em
}

\usepackage{xcolor}
\usepackage{listings}
\usepackage{hyperref}
\usepackage[most]{tcolorbox}

\definecolor{codebg}{RGB}{248,248,248}
\definecolor{codeframe}{RGB}{220,220,220}
\definecolor{inputbg}{RGB}{245,248,255}

\lstdefinestyle{mipcode}{
    language=Python,
    basicstyle=\ttfamily\scriptsize,
    keywordstyle=\bfseries\color{blue!60!black},
    commentstyle=\itshape\color{gray!70!black},
    stringstyle=\color{green!40!black},
    numbers=left,
    numberstyle=\tiny\color{gray},
    stepnumber=1,
    numbersep=6pt,
    showstringspaces=false,
    breaklines=true,
    breakatwhitespace=true,
    frame=single,
    framerule=0.3pt,
    rulecolor=\color{codeframe},
    backgroundcolor=\color{codebg},
    tabsize=4,
    xleftmargin=1.5em,
    framexleftmargin=1.2em
}

\newtcolorbox{codeexample}[2]{
    enhanced,
    breakable,
    colback=white,
    colframe=black!20,
    boxrule=0.4pt,
    arc=2pt,
    left=4pt,
    right=4pt,
    top=4pt,
    bottom=4pt,
    title=\textbf{#1},
    fonttitle=\small,
    coltitle=black,
    colbacktitle=black!5,
    attach boxed title to top left={yshift=-1mm, xshift=2mm},
    boxed title style={boxrule=0pt, arc=2pt},
    before upper={\textbf{Input instruction.}\par\smallskip},
    segmentation style={solid, black!15},
    lower separated=true,
    before lower={\textbf{Generated executable program.}\par\smallskip}
}

\title{M$^3$P-R1: Reinforcement Learning for Large Language Model Guided Multi-Modal Motion Planning via MIP Code Generation}

\author{
  Xingpeng Sun\\
  Purdue University\\
  United States\\
  \texttt{sun1223@purdue.edu} \\
  \And
  Zherong Pan \\
  LightSpeed Studios \\
  United States \\
  \texttt{zherong.pan.usa@gmail.com} \\
  \AND
  Kai Cheng \\
  Purdue University\\
  United States\\
  \texttt{cheng753@purdue.edu} \\
  \And
  Xindi Tang \\
  Purdue University \\
  United States \\
  \texttt{tang666@purdue.edu} \\
  \And
  Syed Talha Bukhari \\
  Purdue University \\
  United States\\
  \texttt{bukhars@purdue.edu} \\
  \And
  Aniket Bera \\
  Purdue University \\
  United States\\
  \texttt{aniketbera@purdue.edu} \\
}

\begin{document}

\maketitle


\begin{abstract}
Multi-Modal Motion Planning (M$^3$P) requires joint reasoning over continuous motions and discrete mode transitions, making it difficult to solve efficiently. For instance, a bipedal robot may walk to a target location and then use its arms to grasp an object. This scenario captures both mode transitions and continuous dynamics, yielding feasible paths that neither purely discrete nor continuous planners can handle. While Mixed-Integer Programming (MIP) offers a principled framework, constructing tractable formulations for non-convex problems is typically manual and domain-specific, especially in the \emph{approximate, discretization-based} MIP regime needed for non-convex robotic tasks.
We propose M$^3$P-R1, a reinforcement learning method that fine-tunes large language models (LLMs) to decompose M$^3$P tasks into MIP variables, constraints, and objectives. Instead of directly outputting answers—often prone to hallucination—the model generates executable Python code using MIP optimization libraries and constraint interfaces. This enables solver-backed execution for robust and verifiable solutions.
Trained with an outcome-driven reward against the solver, M$^3$P-R1 learns to compose modality-level discretization primitives and synthesize cross-modal coupling constraints, producing executable MIP programs for complex M$^3$P tasks. We further introduce LLMM$^3$P Bench, a benchmark of 1{,}316 problems spanning 5 single-modal and 8 multi-modal compositions, and use it to evaluate the proposed method. Experiments show that M$^3$P-R1 achieves 91.7\% single-modal and 80.9\% multi-modal task success rate, improving over the strongest prompt-only LLM baseline (GPT-5.2) by 26.4 and 40.6 percentage points, respectively. M$^3$P-R1 further transfers to real-world hardware at 87.5\% success rate. These results demonstrate the effectiveness of solver-grounded LLM reasoning for reliable motion planning. Project page: \url{https://github.com/XingpengSun0/CoRL-26-M3P-R1}
\end{abstract}



\section{Introduction}

\begin{figure*}[t]
\centering
\includegraphics[width=\textwidth, trim=0 2.8cm 0 0, clip]{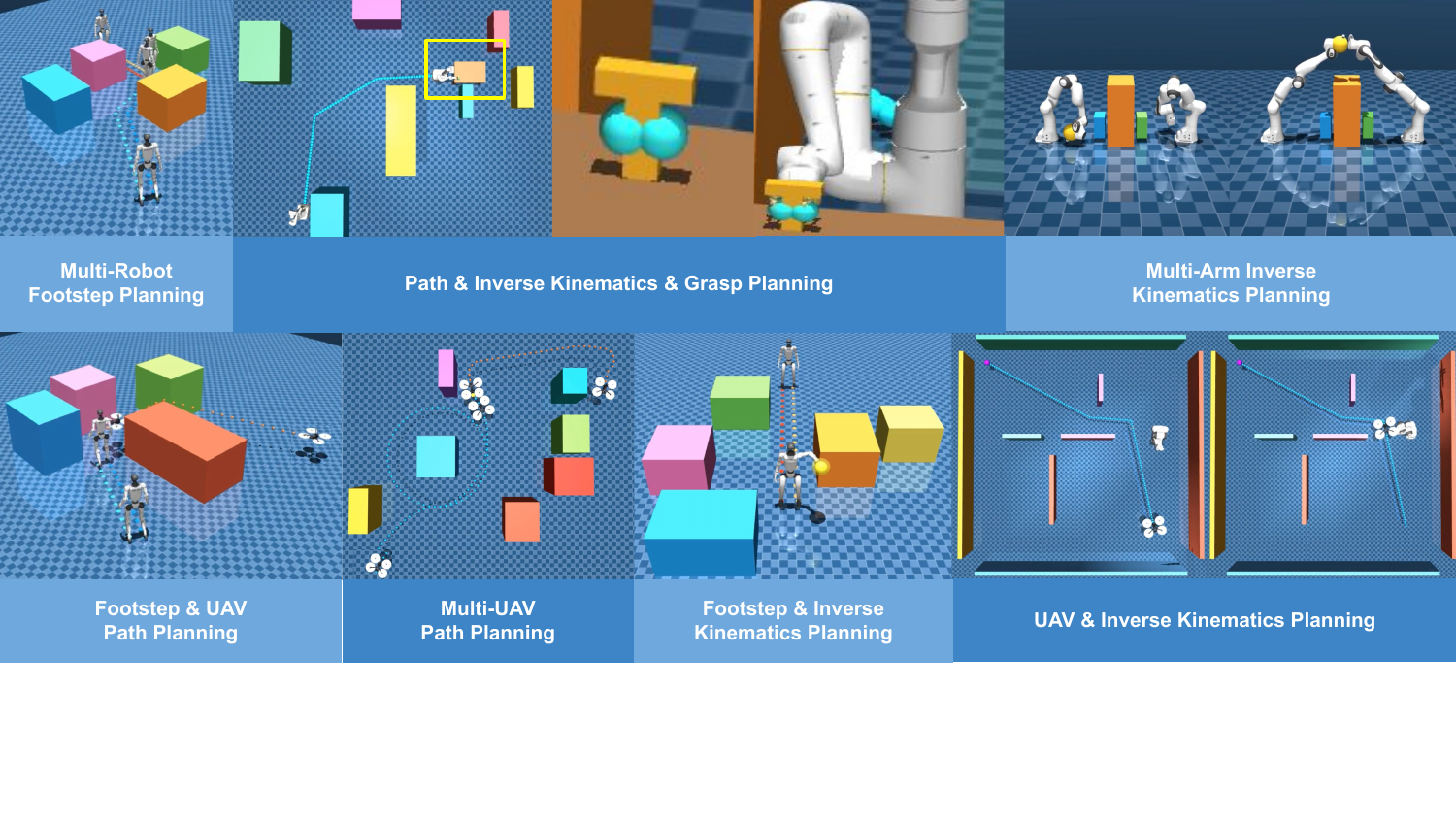}
\caption{\small M$^3$P-R1 enables unified planning for diverse M$^3$P tasks: multi-robot footstep planning, unmanned aerial vehicle (UAV) navigation, inverse kinematics (IK), and grasping, as well as their compositional combinations. By generating solver-grounded MIP programs, the method produces feasible and robust trajectories across heterogeneous robotic domains. Trajectories are shown as dotted lines, with blue spheres on the T-shaped object marking predicted grasp contact points.}
\label{fig:teaser}
\vspace{-1.3em}

\end{figure*}

In this paper, we address Multi-Modal Motion Planning (M$^3$P)~\citep{hauser2010multia,hauser2010multib}, where robots must transition between distinct modes of operation (e.g., grasping, locomotion, navigation), as illustrated in \prettyref{fig:teaser}. As a specialized form of Task and Motion Planning (TAMP), M$^3$P requires joint reasoning over discrete decisions and continuous motions, leading to a large hybrid decision space with tightly coupled symbolic and geometric constraints. While recent advances in TAMP~\citep{dantam2016incremental,toussaint2018differentiable,garrett2021integrated,jiao2022sequential} and M$^3$P~\citep{kingston2020informing,beyer2021multi,kingston2022scaling} improve scalability, many approaches rely on domain-specific heuristics or assumptions such as local optimality~\citep{zhao2024survey,manchester2019variational} or differentiability~\citep{toussaint2018differentiable,envall2023differentiable}, and often decouple task and motion planning, resulting in weak integration.

To address these limitations, unified formulations have been explored. Symbolic approaches based on the Planning Domain Definition Language (PDDL)~\citep{helmert2006fast,piotrowski2024nyx} enable efficient high-level reasoning but lack geometric grounding and depend on external motion planners~\citep{garrett2021integrated}. In contrast, Mixed-Integer Programming (MIP) provides a principled framework for jointly modeling discrete and continuous variables within a single optimization problem. MIP has been successfully applied to grasp synthesis~\citep{liu2020new}, caging~\citep{aceituno2019convex}, inverse kinematics~\citep{dai2019global}, UAV planning~\citep{deits2015efficient,yu2013planning,marcucci2023motion}, and footstep planning~\citep{deits2014footstep}, enabling tighter integration of task--motion reasoning.

With the progress of LLMs, their use in robotics has expanded to interpreting high-level task descriptions and generating plans~\citep{ahn2022can,ding2023task,driess2023palm,huang2022language}. However, prompt-based methods remain unreliable for long-horizon structured reasoning because their outputs are difficult to verify and prone to inconsistency. Prior LLM+P methods~\citep{liu2023llm+,wang2024llm,silver2024generalized} improve reliability by producing symbolic plans, but remain limited to discrete abstractions. More broadly, Reinforcement Learning with Verifiable Rewards (RLVR) shows that grounding LLM reasoning in executable systems can improve reliability through objective, tool-based feedback. Motivated by this paradigm, we propose M$^3$P-R1, where LLMs generate MIP formulations directly from natural-language M$^3$P descriptions. We call this paradigm \emph{LLM+MIP}, extending LLM+P~\citep{liu2023llm+} from symbolic PDDL planning to solver-grounded mixed-integer optimization. This setting requires semantic understanding, mathematical modeling, and program synthesis over hybrid decision spaces. MIP solvers provide feasibility and constraint-satisfaction signals for verification, but generating valid programs remains challenging due to non-convex planning and problem-specific discretization~\citep{sherali2001global,deits2015computing,amice2022finding}. Building on recent advances in LLM mathematical reasoning~\citep{yu2025chain,xiong2024metamath,shi2024math}, our experiments in \prettyref{sec:experiment} show that LLMs can learn such solver-grounded formulations.

To address this challenge, we propose a reinforcement fine-tuning (RFT) framework that trains LLMs to decompose robotic planning tasks into structured optimization components, including decision variables, constraints, and objective functions. By leveraging solver-derived feedback signals such as feasibility status and objective value, the model is directly optimized to generate valid MIP formulation code. This solver-in-the-loop training process enables the LLM to learn to decompose M$^3$P tasks into various MIP constraints and objectives, and naturally extends to multi-modal motion planning scenarios. Note that we focus on \emph{approximate}, discretization-based MIP formulations for non-convex robotic planning, in contrast to exact MIP/MILP formulations targeted by prior LLM-for-optimization work. To evaluate this framework, we introduce LLMM$^3$P Bench, a benchmark dataset and evaluation protocol for LLM+MIP, consisting of diverse single- and multi-modal M$^3$P instances. Experiments show that RFT enables M$^3$P-R1 to outperform the strongest prompt-only LLM baseline (GPT-5.2) by 26.4/40.6 pp in single-/multi-modal task success rate.

Our core contributions are as follows:
\begin{enumerate}[leftmargin=*,topsep=2pt,itemsep=1pt]
\item \textbf{LLM+MIP method:} Building on prior LLM-to-optimization work that targets exact MIP/MILP formulations~\citep{li2023synthesizing,ahmaditeshnizi2024optimus,li2025towards}, we introduce a solver-grounded method that enables LLMs to generate executable \emph{approximate} MIP formulations for M$^3$P, bridging language-based reasoning with optimization-based planning over hybrid discrete--continuous decision spaces.
\item \textbf{RFT for structured decomposition:} We propose a RFT framework that leverages solver feedback to train LLMs to decompose robotic planning tasks into variables, constraints, and objectives, improving correctness, executability, and generalization to held-out task difficulties.
\item \textbf{Benchmark and empirical validation:} We present a new benchmark for LLM+MIP and demonstrate that RFT enables reliable generation of MIP programs across diverse single- and multi-modal planning tasks.
\end{enumerate}

\section{Related Work}
\vspace{-0.5em}

\textbf{TAMP, M$^3$P, and MIP-based planning.}
Task and Motion Planning (TAMP) integrates high-level symbolic reasoning with continuous motion feasibility~\citep{garrett2021integrated,dantam2016incremental}, while Multi-Modal Motion Planning (M$^3$P) focuses on transitions between locomotion, contact, and manipulation modes~\citep{hauser2010multia,hauser2010multib}. Unified TAMP approaches, including Logic-Geometric Programming (LGP), jointly reason over discrete mode sequences and continuous geometric feasibility, while mixed-integer formulations such as ScottyActivity provide related unified optimization frameworks under convex continuous constraints. Thus, jointly coupling task-level choices with continuous feasibility is not itself our novelty. Mixed-Integer Programming (MIP) has also been applied to grasping, caging, inverse kinematics, UAV navigation, multi-robot planning, and footstep planning~\citep{liu2020new,aceituno2019convex,dai2019global,deits2015efficient,yu2013planning,marcucci2023motion,deits2014footstep}. In contrast to these human-specified planning formulations, our focus is on generating executable optimization programs directly from natural-language task descriptions, including approximate formulations for non-convex geometric constraints using IRIS-style convex decomposition. Sampling-based and trajectory-optimization methods such as RRT*, BIT*, PRM, and TrajOpt~\citep{karaman2011sampling,gammell2015batch,kavraki1996probabilistic,schulman2014motion} handle continuous motion but typically decouple geometric planning from discrete mode reasoning. PDDLStream~\citep{garrett2020pddlstream} connects symbolic planning with continuous feasibility through hand-designed streams, but requires a manually specified symbolic model and continuous samplers.

\textbf{LLMs for planning.}
LLMs have been used for robotic planning through direct prompting, symbolic translation, and code generation~\citep{ahn2022can,ding2023task,driess2023palm,huang2022language}. Methods such as ReAct~\citep{yao2022react} improve interactive reasoning but lack formal optimization grounding. LLM+P-style approaches translate language into PDDL or related symbolic representations~\citep{liu2023llm+,wang2024llm,silver2024generalized,agarwal2025l3m+}, while Code as Policies and ProgPrompt generate executable robot programs~\citep{liang2023code,singh2022progprompt}. Prior work has also explored richer language and multimodal interfaces for robot planning. TrustNavGPT models uncertainty in audio-guided LLM navigation~\citep{sun2024trustnavgpt}, while Beyond Text incorporates vocal cues into LLM decision making for robot navigation~\citep{sun2024beyond}. In inspection planning, Sun et al.~\citep{sun2025textguided} use a VLM to interpret language-specified inspection goals and combine VLM-selected viewpoints with TSP and trajectory optimization to generate executable paths. These methods improve language-conditioned planning but rely on predefined planning or optimization pipelines rather than synthesizing the underlying optimization formulation itself. Similarly, waypoint-based approaches that combine LLM outputs with RRT*, BIT*, or TrajOpt still separate semantic decomposition from continuous optimization and therefore struggle with tightly coupled M$^3$P constraints.

Foundation models have also been combined with classical geometric optimization outside robot planning. For example, Sun et al.~\citep{sun2026mesh} use a VLM to interpret natural-language deformation instructions, select deformation handles, and couple these predictions with a classical handle-based mesh deformation algorithm. Such systems demonstrate the benefit of using foundation models as semantic interfaces to structured geometric solvers, but the downstream optimization model remains fixed and manually designed. In contrast, our goal is to train an LLM to construct the optimization program itself, including task-specific variables, objectives, constraints, discretization primitives, and cross-modal coupling constraints.

\textbf{LLMs for MIP and solver-grounded reinforcement fine-tuning.}
Recent work has explored LLMs for optimization, including diagnosing infeasible models, translating natural language into MIP scripts, and learning foundation models for MILP~\citep{li2023large,chen2024diagnosing,liu2024evolution,li2023synthesizing,ahmaditeshnizi2024optimus,li2025towards}. However, these methods primarily target exact MIP or operations-research-style MILP instances. In contrast, robotic M$^3$P requires approximate MIP formulations with geometry-specific discretization choices. Reinforcement fine-tuning methods such as RLHF, PPO, DPO, and GRPO~\citep{ouyang2022training,schulman2017proximal,rafailov2023direct,shao2024deepseekmath} show that outcome-based rewards can improve structured reasoning, especially when external tools provide verifiable feedback. Solver-grounded and tool-integrated LLM systems further demonstrate the value of executable verification~\citep{chen2025solver,feng2025retool,li2025start}.

Unlike classical TAMP and MIP-based M$^3$P methods~\citep{garrett2021integrated,liu2020new,deits2014footstep} and prior language/VLM-guided planning systems~\citep{sun2025textguided,sun2026mesh}, which rely on human-specified models or fixed downstream optimizers, M$^3$P-R1 maps natural-language tasks directly to executable optimization programs by generating task-specific variables, constraints, objectives, and cross-modal couplings. Unlike LLM planning and program-generation methods that use execution or solver feedback mainly for test-time refinement~\citep{liu2023llm+,liang2023code,singh2022progprompt}, we use solver feedback as a reinforcement fine-tuning signal, enabling single-pass program generation at inference.

\vspace{-1.0em}
\section{Method}
\begin{figure*}[t]
\centering
\includegraphics[width=\textwidth, trim=0.2cm 0 0.2cm 0, clip]{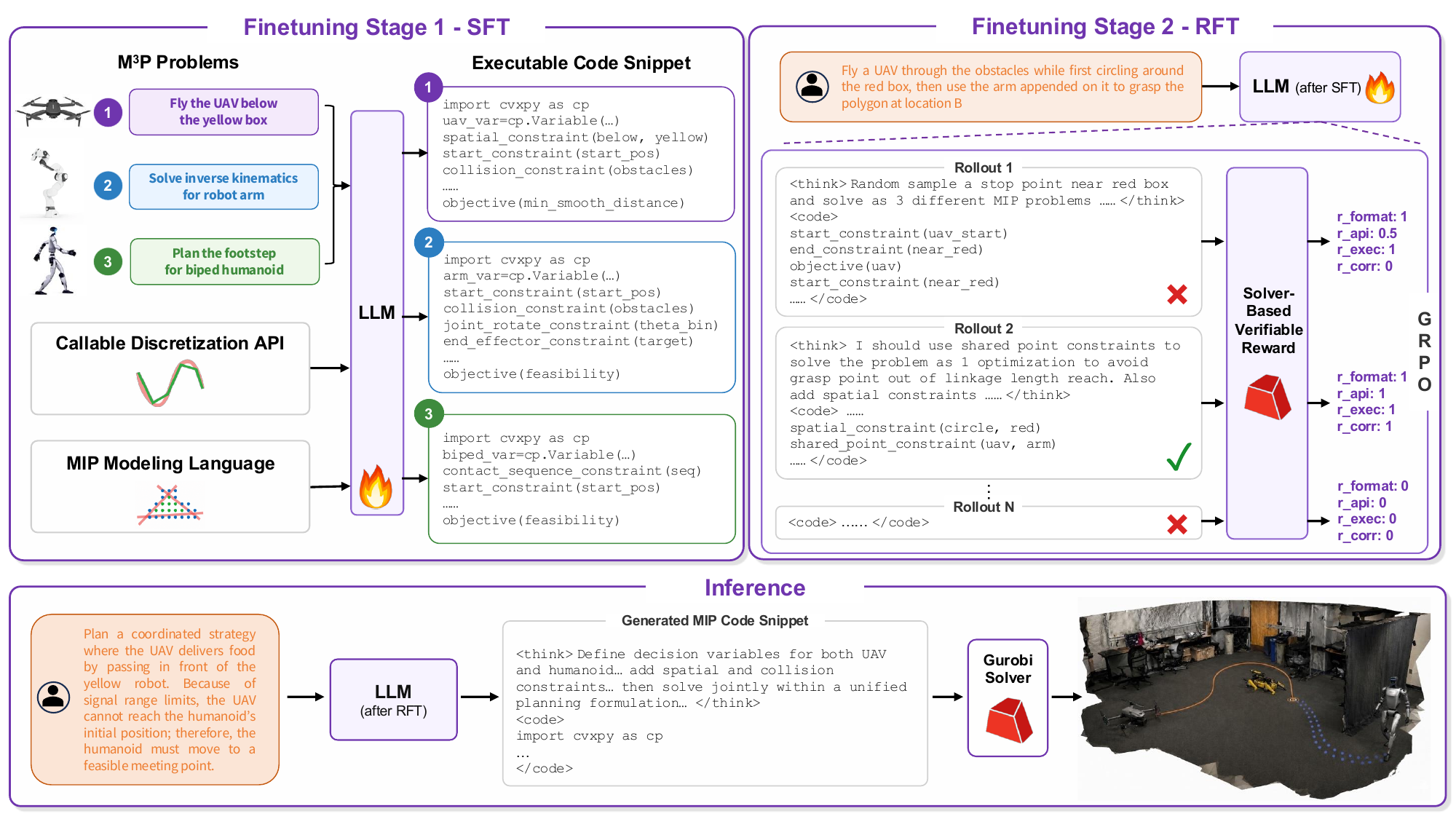}
\caption{\small M$^3$P-R1 pipeline. The model is first supervised to generate structured MIP programs via a discretization API, then refined with GRPO using solver-verifiable rewards (format, API coverage, execution, and correctness). At inference, it maps natural language tasks to executable MIP code with intermediate reasoning, which is solved to yield feasible multi-modal plans.}
\label{fig:paradigm}
\vspace{-1.0em}
\end{figure*}
\vspace{-0.5em}
Our pipeline is shown in \prettyref{fig:paradigm}. To translate M$^3$P descriptions into executable MIP formulations, we expose an existing MIP modeling language as an LLM-callable Application Programming Interface (API). We use CVXPY~\citep{diamond2016cvxpy}, a high-level optimization package supporting multiple backend solvers, as the modeling interface, with Gurobi~\citep{pedroso2011optimization} as the default solver.

Approximate MIP for M$^3$P requires problem-specific discretization, such as Iterative Regional Inflation by Semidefinite Programming (IRIS)~\citep{deits2015computing} and related methods~\citep{sherali2001global,amice2022finding}, to approximate non-convex constraints with disjoint convex representations. Since deriving such discretizations from scratch is beyond our scope, we provide CVXPY-integrated discretization primitives as LLM-callable APIs. The library contains only 23 modality-level functions, reused unchanged across all five single-modal and eight multi-modal task families. It contains no multi-modal template or coupling primitive: cross-modal interfaces such as meeting points, contact locations, shared states, and reachability constraints must be generated by the LLM. Thus, multi-modal planning still requires the LLM to synthesize a joint MIP formulation by composing single-modality primitives and adding coupling constraints, such as shared meeting points, contact locations, reachability constraints, and synchronized intermediate states. The fine-tuned LLM therefore learns to generate executable code that uses these primitives while constructing task-specific MIP constraints and objectives for M$^3$P. The system prompt and representative \texttt{<think>}/\texttt{<code>} output are shown in \prettyref{appen:llm_prompt}; the API specification, MIP formulations, and data samples are provided in \prettyref{appen:api}, \ref{appen:math_detail}, and \ref{appen:data_example}. We introduce the dataset and GRPO training below.

\begin{wrapfigure}{r}{0.5\textwidth}
\centering
\includegraphics[width=\linewidth]{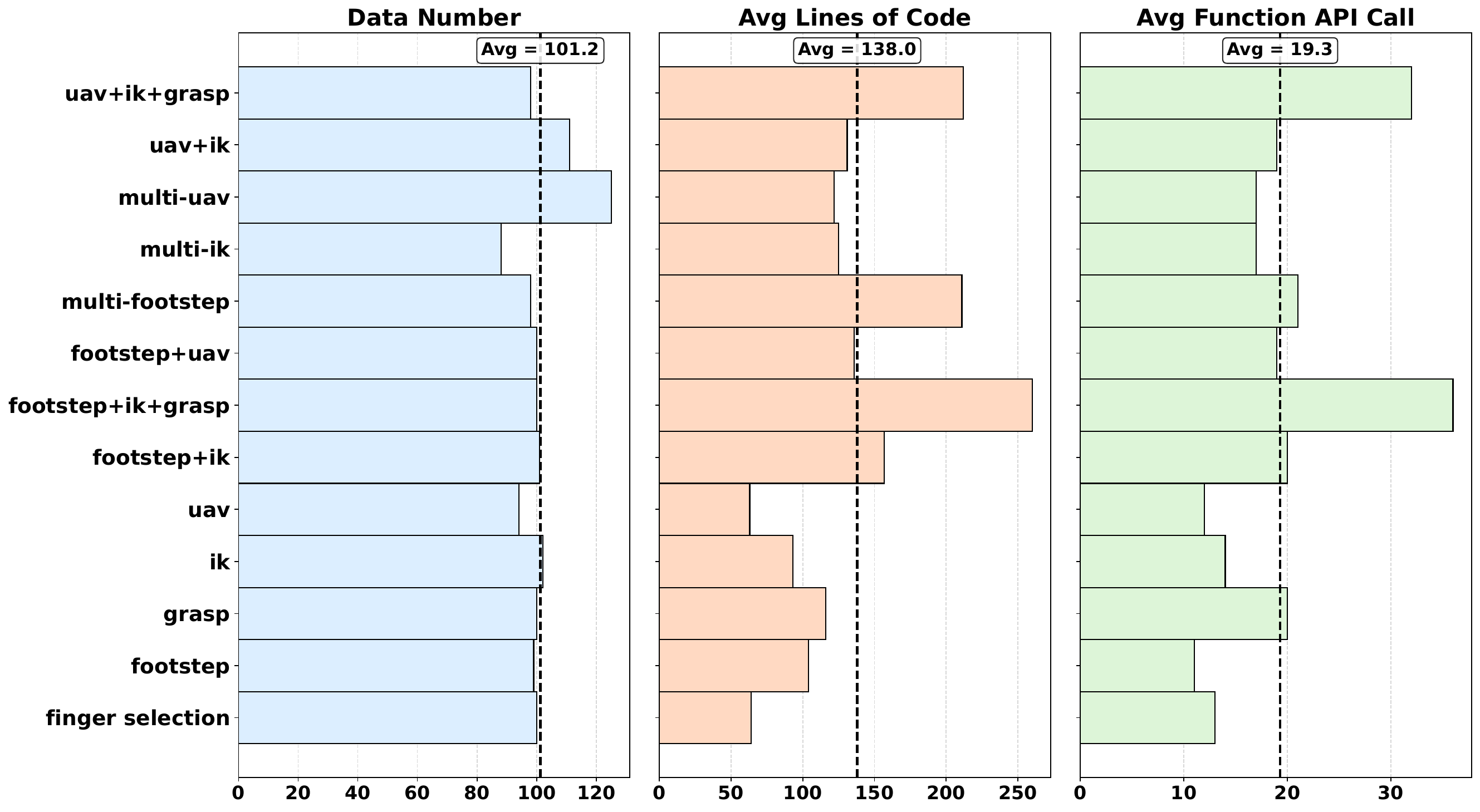}
\caption{\small Dataset statistics. Left: total samples per modality; Middle: LoC distribution; Right: function-call distribution. Bars are grouped by modality (single-modal vs.\ multi-modal compositions); single-modal tasks are smaller, while multi-modal compositions are more complex.}
\label{fig:dataset}
\end{wrapfigure}

\vspace{-0.8em}
\paragraph{LLMM$^3$P Bench}\label{sec:dataset}
We introduce LLMM$^3$P Bench, a dataset of 1,316 sampled M$^3$P problems covering five single-modal tasks and eight multi-modal compositions that integrate locomotion, manipulation, and aerial navigation (\prettyref{appen:task}). Each sample consists of: (i) a natural language task description, including an obstacle map specifying the environment layout and object boundaries, and (ii) fully executable MIP trajectory code. A statistical summary is shown in \prettyref{fig:dataset}. In total, LLMM$^3$P Bench comprises 16.8K lines of code (LoC), with an average of 138 LoC per sample (range: 63--260), and 23K function calls, averaging 138 LoC and 19.3 API calls per program (range: 11--36). These statistics highlight both the diversity and structural complexity of the dataset, making it a challenging benchmark for evaluating whether LLMs can generalize to generate correct, executable, and scalable MIP formulations for robotic motion planning tasks. Of the 1,316 samples, 100 are used for supervised fine-tuning (SFT), 1,040 for GRPO training (including the SFT subset), and 276 for evaluation. We detail the dataset collection pipeline in \prettyref{appen:data_generation}. We will release the LLMM$^3$P Bench dataset, training code, and model checkpoints upon acceptance.
\vspace{-0.8em}

\paragraph{Training Pipeline}\label{sec:training}
Let \(f\) denote a generic base LLM with parameters \(\theta_0\) adapted into policy \(\pi_\theta\); given a mission specification \(x\), its task environment \(\mathcal{E}_x\) (workspace bounds, obstacles, start and goal states; full tuple in \prettyref{appen:data_generation}), and a predefined MIP discretization tool library \(\mathcal{T}=\{t_k\}\), the policy \(\pi_\theta(y|x)\) generates an executable program \(y\) that composes API calls \(t_k\in\mathcal{T}\) with variable, constraint, and objective definitions. The program defines a MIP instance executed by solver \(\mathcal{S}\), producing \((\tau,\mathcal{L})=\mathcal{S}(y,\mathcal{E}_x)\), where \(\tau\) is the trajectory and \(\mathcal{L}\) is the API execution log; learning therefore optimizes \(\pi_\theta\) to both select modality-level tools and compose them with newly generated variables, constraints, and objectives into a unified MIP formulation through two stages: SFT on single-modal demonstrations to obtain \(\pi_{\theta_{\text{SFT}}}\), followed by GRPO refinement with solver-verifiable rewards, instantiating RLVR with deterministic solver feedback.

In Stage 1, we perform supervised fine-tuning on single-modal demonstrations \(\mathcal{D}_{\text{single}}\) by maximizing the log-likelihood objective:
\[
\mathcal{J}_{\text{SFT}}(\theta)
=
\operatorname{E}_{(x,y)\sim\mathcal{D}_{\text{single}}}
\left[\log \pi_\theta(y|x)\right].
\]
This yields a bootstrapped policy \(\pi_{\theta_{\text{SFT}}}\), whose output \(y \sim \pi_{\theta_{\text{SFT}}}(\cdot|x)\) is a syntactically valid program that follows the required format, but is not yet optimized for solver success or task feasibility.

In Stage 2, starting from \(\pi_{\theta_{\text{SFT}}}\), we optimize tool usage and solution quality via Group Relative Policy Optimization (GRPO)~\citep{shao2024deepseekmath}. GRPO computes advantages by normalizing rewards within each sampled group, eliminating the need for a learned value function.
For each \(x\), we sample a group of \(G\) programs \(\{y_i\}_{i=1}^G \sim \pi_\theta(\cdot|x)\) (group size \(G\)), execute them with \(\mathcal{S}\), let \(\tau_i\) be the trajectory component of \(\mathcal{S}(y_i,\mathcal{E}_x)\), and compute reward \(r_i=\mathcal{R}_\text{rew}(x,y_i,\tau_i)\):
\[
\mathcal{R}_\text{rew}(x,y,\tau)
=
\lambda_1 R_{\text{format}}(y)
+
\lambda_2 R_{\text{API}}(x,y)
+
\lambda_3 R_{\text{exec}}(y)
+
\lambda_4 R_{\text{corr}}(x,y,\tau),
\]
where \(\lambda_1,\ldots,\lambda_4 \ge 0\) are scalar weighting coefficients balancing the four reward terms (values reported in \prettyref{appen:finetune_detail}, \prettyref{tab:grpo_details}). The format reward is defined as the indicator:
$R_{\text{format}}(y) = \mathbf{1}[\,y\ \text{matches the}\ \texttt{<think>}\dots\texttt{</think>}\texttt{<code>}\dots\texttt{</code>}\ \text{schema}\,]\in\{0,1\},
$
enforcing the required structured output schema by checking the presence and proper closure of the reasoning \texttt{<think>} \dots \texttt{</think>} and code \texttt{<code>} \dots \texttt{</code>} blocks. The execution reward is the indicator (with the fixed solver \(\mathcal{S}\) and environment \(\mathcal{E}_x\) treated as hyper-parameters):
$
R_{\text{exec}}(y) = \mathbf{1}[\,y_{\texttt{code}}\ \text{executes successfully on}\ \mathcal{S}\,]\in\{0,1\},$
which verifies that the generated code \(y_{\texttt{code}}\)---the content inside the \texttt{<code>}\dots\texttt{</code>} block of \(y\)---is syntactically valid and executable in Python, when executed in \(\mathcal{E}_x\) against the fixed solver \(\mathcal{S}\). To teach the model to use the correct MIP discretization APIs, let \(\mathcal{A}(x)\subseteq \mathcal{T}\) denote the required discretization APIs and \(\hat{\mathcal{A}}(y)\subseteq \mathcal{T}\) those invoked in \(y\); we define:
$
R_{\text{API}}(x,y) = \frac{|\hat{\mathcal{A}}(y)\cap\mathcal{A}(x)|}{|\mathcal{A}(x)|}.
$
This coverage-based reward provides a smoother optimization signal than binary correctness, assigning partial credit to incomplete but valid API usage, thereby improving training stability. The required API set is used only during training as an auxiliary reward and is obtained from the executable expert traces in LLMM$^3$P Bench; it is not provided at inference time. For trajectory correctness, let \(\tau=(s_t)_{t=1}^T\) denote the solver-generated trajectory of length \(T\), with \(s_t\in\mathcal{X}\) for a task-dependent state space \(\mathcal{X}\) (concrete instantiation per modality in \prettyref{appen:math_detail}); let \(\text{stat}(y, \mathcal{E}_x)\) denote the solver status induced by executing \(y\) in \(\mathcal{E}_x\), and let \(\text{stat}^*(x)\) denote the ground-truth solver status for task \(x\). Here \(\Phi_{\text{init}}\) and \(\Phi_{\text{goal}}\) verify that all agents' start and goal states match the mission specification \(x\), and \(\Phi_{\text{dyn}}\) enforces that the trajectory satisfies all spatial and task constraints, including collision-free motion for each agent and consistent coordination at shared states for multi-agent tasks. Each \(\Phi_*\) may also use the environment \(\mathcal{E}_x\); we treat \(\mathcal{E}_x\) as packaged inside \(x\). 
We define $R_{\text{corr}}(x,y,\tau)=\mathbf{1}\!\left[\text{stat}(y,\mathcal{E}_x)=\text{stat}^*(x)\wedge \Phi_{\text{init}}(x,\tau)=1 \wedge \Phi_{\text{goal}}(x,\tau)=1 \wedge \Phi_{\text{dyn}}(x,\tau)=1\right]$.
Trajectory correctness is evaluated by executing the generated program with the solver and checking task constraints, rather than comparing against a fixed reference trajectory. This reward assigns \(1\) only when the generated program yields a solver-consistent and fully valid trajectory, and \(0\) otherwise.
Rewards are normalized within each group: with \(\bar{r}\) and \(\sigma_r\) denoting the mean and standard deviation of \(\{r_i\}_{i=1}^{G}\), we define the group-normalized advantage as$
\hat{A}_i=\frac{r_i-\bar{r}}{\sigma_r}.$
Let \(\pi_{\theta_{\mathrm{old}}}\) denote the behavior policy used to generate each sampled group. We optimize the standard clipped GRPO objective

$$
\mathcal{L}_{\mathrm{GRPO}}
=
-\mathbb{E}_{x\sim\mathcal{D}_{\mathrm{GRPO}}}
\left[
\frac{1}{G}\sum_{i=1}^{G}
\frac{1}{T_i}\sum_{t=1}^{T_i}
\min\!\left(
\rho_{i,t}\hat{A}_i,\,
\operatorname{clip}\!\left(\rho_{i,t},1-\epsilon,1+\epsilon\right)\hat{A}_i
\right)
\right],
$$

where$
\rho_{i,t}
=
\frac{\pi_{\theta}(y_{i,t}\mid h_{i,t})}
{\pi_{\theta_{\mathrm{old}}}(y_{i,t}\mid h_{i,t})},
$
\(h_{i,t}\) denotes the generation history preceding token \(t\), \(T_i\) is the length of program \(y_i\), and \(\epsilon\) is the clipping coefficient. In our implementation, we use no additional KL-divergence penalty (\(\beta=0\)). The solver \(\mathcal{S}\), training set \(\mathcal{D}_{\mathrm{GRPO}}\), and reward functions \(\{R_{\mathrm{format}},R_{\mathrm{API}},R_{\mathrm{exec}},R_{\mathrm{corr}}\}\) are fixed during training. This objective refines \(\pi_{\theta_{\mathrm{SFT}}}\) into a policy that learns to compose MIP discretization APIs into executable formulations whose solver outputs satisfy the task specification.

\section{Experiment}\label{sec:experiment}
\vspace{-0.5em}

Our evaluation focuses on answering four questions: \textbf{(i)} \textit{Can LLMs correctly generate code that uses the provided API functions to assemble a MIP formulation that directly solves an M$^3$P task described in natural language?} \textbf{(ii)} \textit{How does M$^3$P-R1 compare to symbolic and sampling-based planning baselines?} \textbf{(iii)} \textit{What is the contribution of each training stage and reward component?} \textbf{(iv)} \textit{Does the approach transfer to real-world hardware?}
We use Qwen3-8B as our base model. Detailed training parameters are provided in \prettyref{appen:finetune_detail}. 
\vspace{-1.0em}
\paragraph{Experimental Setup.}
\setlength{\intextsep}{-0.5pt}
\begin{wraptable}{r}{0.63\textwidth}
\vspace{-1em}
\centering
\small
\caption{\small Average performance for LLMs across single-modal and multi-modal settings over three runs. Values are mean $\pm$ standard deviation.}
\label{tab:llm_avg_performance}

\resizebox{\linewidth}{!}{%
\begin{tabular}{lcccccc}
\toprule
& \multicolumn{3}{c}{Single-modal}
& \multicolumn{3}{c}{Multi-modal} \\
\cmidrule(lr){2-4}
\cmidrule(lr){5-7}

Model
& API (\%) & Exec (\%) & SR (\%)
& API (\%) & Exec (\%) & SR (\%) \\
\midrule

\textbf{Ours}
& $\mathbf{98.0 \pm 1.7}$
& $\mathbf{94.3 \pm 1.2}$
& $\mathbf{91.7 \pm 0.6}$
& $\mathbf{94.5 \pm 0.3}$
& $\mathbf{88.8 \pm 1.8}$
& $\mathbf{80.9 \pm 2.3}$ \\

GPT-5.2
& $89.3 \pm 4.0$
& $88.3 \pm 4.5$
& $65.3 \pm 6.5$
& $83.9 \pm 3.1$
& $60.0 \pm 3.7$
& $40.3 \pm 4.0$ \\

Gemini 2.5 Flash
& $86.3 \pm 6.0$
& $56.3 \pm 4.6$
& $49.0 \pm 11.0$
& $84.8 \pm 6.3$
& $50.9 \pm 7.7$
& $35.8 \pm 6.2$ \\

Qwen3-30B
& $68.7 \pm 10.5$
& $81.7 \pm 10.5$
& $56.7 \pm 9.5$
& $38.8 \pm 6.2$
& $34.7 \pm 6.8$
& $30.3 \pm 6.0$ \\

Gemma3-12B
& $81.7 \pm 7.6$
& $39.7 \pm 12.0$
& $35.7 \pm 7.5$
& $35.8 \pm 6.8$
& $28.6 \pm 6.0$
& $14.2 \pm 5.1$ \\

Qwen3-8B
& $67.0 \pm 11.8$
& $76.0 \pm 6.1$
& $58.3 \pm 9.8$
& $60.0 \pm 8.2$
& $33.1 \pm 7.1$
& $25.4 \pm 5.2$ \\

LLaMA3.1-8B
& $52.7 \pm 12.5$
& $75.7 \pm 7.5$
& $52.7 \pm 7.5$
& $32.4 \pm 5.7$
& $38.8 \pm 6.2$
& $27.7 \pm 6.5$ \\

Qwen3-Coder-Next
& $52.3 \pm 8.7$
& $83.3 \pm 9.3$
& $51.7 \pm 9.2$
& $42.0 \pm 6.8$
& $43.9 \pm 8.2$
& $33.9 \pm 7.1$ \\

Qwen3-4B
& $34.3 \pm 21.4$
& $27.3 \pm 8.3$
& $23.0 \pm 8.2$
& $21.2 \pm 5.4$
& $15.0 \pm 4.8$
& $10.8 \pm 4.0$ \\

Gemma3-4B
& $49.0 \pm 6.2$
& $4.7 \pm 5.0$
& $4.3 \pm 5.1$
& $21.8 \pm 4.6$
& $7.0 \pm 4.3$
& $1.9 \pm 2.0$ \\

\bottomrule
\end{tabular}%
}

\end{wraptable}
\textit{Metrics.} We report three complementary metrics throughout: \textbf{API (\%)} measures structured output correctness (how often the generated code uses the discretization APIs with valid signatures); \textbf{Exec (\%)} measures whether the generated program executes successfully on the solver; and \textbf{SR (\%)} measures full task-level correctness against the ground-truth solver status and trajectory constraints (\prettyref{appen:finetune_detail}).
\textit{LLM-baseline evaluation protocol.} All prompt-only LLM baselines, as well as the LLM component of LLM+P, LLM+RRT*, LLM+BIT*, and LLM+TrajOpt, use \textbf{GPT-5.2} as their underlying LLM, with sampling temperature $T_{\text{samp}}=0.6$ and one in-context example for format alignment. ReAct uses Qwen3-8B as the backbone, with the same API handbook provided in context. Parse failures are retried once before being recorded as a failure. The full protocol is documented in \prettyref{appen:baselines_setup}. PDDLStream operates on pre-parsed (gold) symbolic input rather than free-form language.

\begin{table}[t]
\centering
\small
\caption{\small Comparison with symbolic/prompt (left) and sampling-based (right) baselines. Values are reported as mean $\pm$ standard deviation over three runs. \textit{SR (\%)} is task-level success rate (API/Exec/SR metrics defined in \prettyref{tab:llm_avg_performance}). \textit{Time} denotes end-to-end planning time (LLM inference + solver solve) in seconds. \textit{Qual} is the ratio of path length and trajectory smoothness relative to our method (1.00/1.00), computed only on instances where both Ours and the baseline succeed; lower is better. PDDLStream operates on pre-parsed (gold) symbolic input rather than free-form language (see \prettyref{appen:baselines_setup}).}
\label{tab:combined_comparison}
\resizebox{\linewidth}{!}{
\begin{tabular}{lcccccc|lcccc}
\toprule
\multicolumn{7}{c|}{\textbf{Symbolic \& Prompt Baselines}}
& \multicolumn{5}{c}{\textbf{Sampling-based Baselines}} \\
\cmidrule(lr){1-7} \cmidrule(lr){8-12}

& \multicolumn{3}{c}{Single} & \multicolumn{3}{c}{Multi}
& & \multicolumn{2}{c}{Single} & \multicolumn{2}{c}{Multi} \\
\cmidrule(lr){2-4} \cmidrule(lr){5-7} \cmidrule(lr){9-10} \cmidrule(lr){11-12}

Method
& SR & Time & Qual
& SR & Time & Qual
& Method & Path & Smooth & Path & Smooth \\

\midrule

\textbf{Ours}
& $91.7 \pm 0.6$
& $20.2 \pm 1.1$
& $\mathbf{1.00/1.00}$
& $\mathbf{80.9 \pm 2.3}$
& $\mathbf{99.7 \pm 7.8}$
& $\mathbf{1.00/1.00}$
& \textbf{Ours}
& $1.00 \pm 0.05$
& $\mathbf{1.00 \pm 0.04}$
& $\mathbf{1.00 \pm 0.17}$
& $\mathbf{1.00 \pm 0.22}$ \\

PDDLStream~\cite{garrett2020pddlstream}
& $\mathbf{95.0 \pm 3.5}$
& $\mathbf{18.1 \pm 2.4}$
& $1.21/1.36$
& $57.6 \pm 5.8$
& $101.5 \pm 8.2$
& $1.42/1.29$
& LLM+RRT*~\cite{karaman2011sampling}
& $0.95 \pm 0.11$
& $1.48 \pm 0.19$
& $1.36 \pm 0.14$
& $1.73 \pm 0.27$ \\

LLM+P~\cite{liu2023llm+}
& $85.9 \pm 5.7$
& $33.8 \pm 4.9$
& $1.39/1.48$
& $39.4 \pm 6.3$
& $118.6 \pm 11.5$
& $1.78/1.61$
& LLM+BIT*~\cite{gammell2015batch}
& $\mathbf{0.93 \pm 0.08}$
& $1.46 \pm 0.21$
& $1.24 \pm 0.12$
& $1.57 \pm 0.42$ \\

ReAct~\cite{yao2022react}
& $69.8 \pm 8.1$
& $64.7 \pm 12.2$
& $2.51/2.44$
& $24.6 \pm 6.9$
& $149.8 \pm 20.3$
& $3.11/1.89$
& LLM+TrajOpt~\cite{schulman2014motion}
& $1.09 \pm 0.14$
& $1.34 \pm 0.11$
& $1.18 \pm 0.16$
& $1.34 \pm 0.08$ \\

\bottomrule
\end{tabular}
}
\vspace{-1.2em}
\end{table}

\begin{wrapfigure}{r}{0.5\textwidth}
    \centering
    \includegraphics[width=\linewidth, trim=0 0 1cm 0, clip]{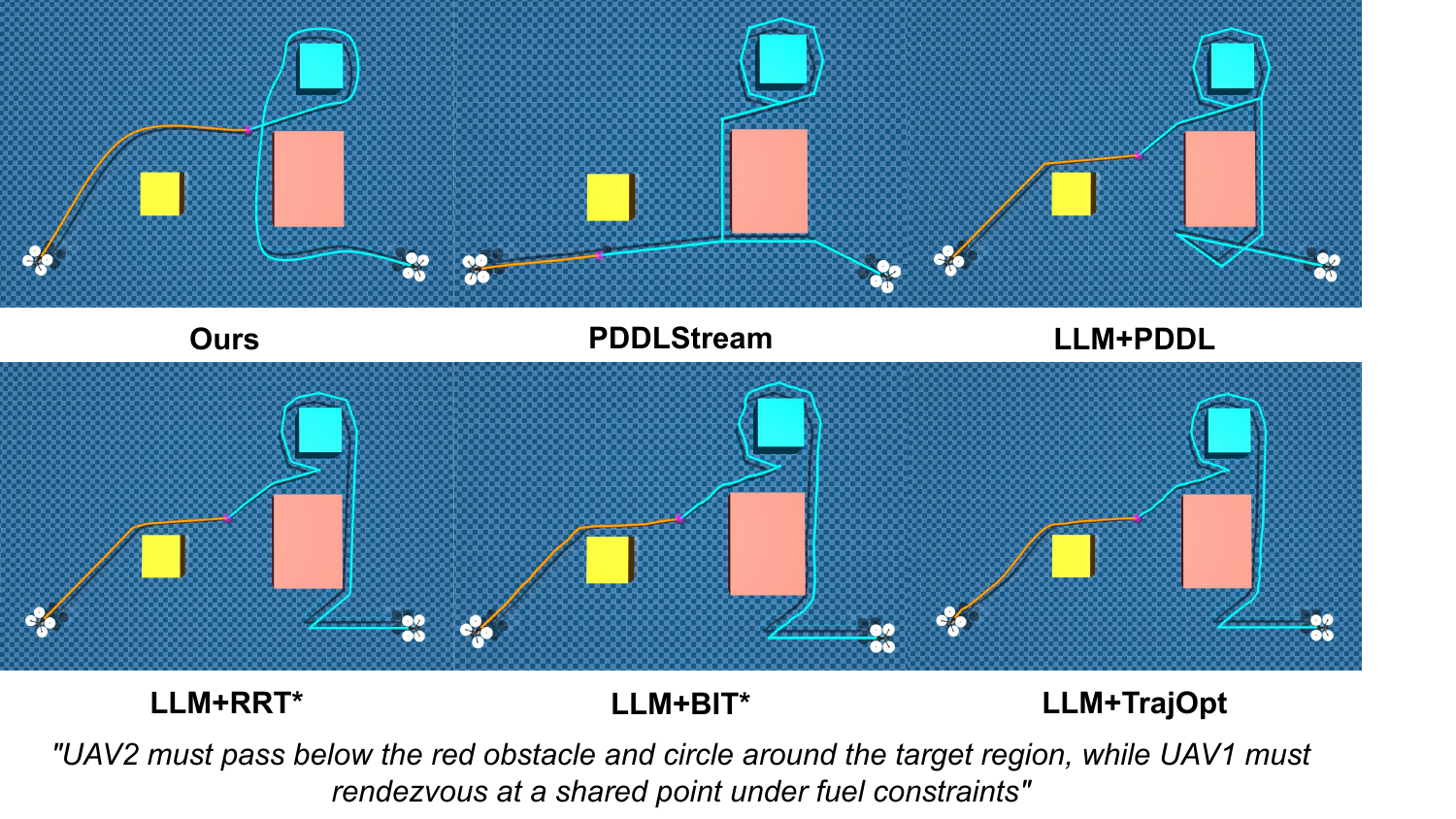}
    \caption{\small Comparison of multi-UAV path planning. UAV trajectories are shown in orange (UAV1) and blue (UAV2). Our method generates smoother and shorter trajectories that better satisfy spatial requirements. The bottom three sampling-based methods do not complete a full circle around the blue box, and therefore fail the spatial requirement.}
    \label{fig:comparison}
\end{wrapfigure}
\vspace{-1.0em}
\paragraph{Quantitative analysis}
In \prettyref{tab:llm_avg_performance}, we compare M$^3$P-R1 with proprietary LLMs (GPT-5.2, Gemini-2.5-Flash), open-source LLMs of different scales (Qwen3, LLaMA3.1, Gemma3, 4B--30B), and a coding-specialized model (Qwen3-Coder-Next). We report API/Exec/SR separately for \textit{single-modal} tasks (UAV, IK, footstep, grasp, and finger selection) and \textit{multi-modal} compositions (e.g., UAV+IK, multi-UAV, multi-IK, multi-footstep, and higher-order combinations), following the protocol in \prettyref{appen:baselines_setup}. All baselines receive one in-context example for format alignment. Although this improves API and execution rates, prompt-only methods remain unreliable on task correctness, especially for multi-modal tasks (40.3\% SR), motivating RFT with solver-verifiable rewards. In \prettyref{tab:combined_comparison}, we compare against six symbolic and sampling-based baselines. We report success rate (SR), end-to-end planning time (LLM inference + solver solve), and solution quality (Qual). Qual is evaluated only on successful single- and multi-UAV instances where both methods succeed. For a trajectory $\tau=(s_t)_{t=1}^T$, $s_t\in\mathbb{R}^3$, smoothness is measured by discrete curvature $\sum_{t=2}^{T-1}\|s_{t+1}-2s_t+s_{t-1}\|$. Qual reports path length and smoothness normalized by our method (1.00/1.00), where lower is better. PDDLStream~\cite{garrett2020pddlstream} uses symbolic PDDL with black-box streams for continuous feasibility but struggles with tightly coupled multi-modal tasks. LLM+P~\cite{liu2023llm+} generates PDDL instances under fixed symbolic abstractions, while ReAct~\cite{yao2022react} performs ungrounded iterative prompting. LLM+RRT*~\cite{karaman2011sampling}, LLM+BIT*~\cite{gammell2015batch}, and LLM+TrajOpt~\cite{schulman2014motion} use LLM-proposed waypoints or shared points followed by classical planners, and are limited to UAV settings.

\begin{wraptable}{r}{0.48\textwidth}
    \centering
    \small
    \caption{\small Ablation on training stages and reward terms. Results are data-count-weighted averages over all single- and multi-modal tasks, reported as mean $\pm$ standard deviation over three runs. \textit{Ours (800 steps)} is the final M$^3$P-R1 reference from \prettyref{tab:llm_avg_performance}; reward ablations are compared against the matched \textit{GRPO-200} checkpoint.}
    \label{tab:rft_ablation}
    \resizebox{\linewidth}{!}{
    \begin{tabular}{lccc}
    \toprule
    Model & API(\%) & Exec(\%) & SR(\%) \\
    \midrule
    \textbf{Ours (800 steps)}
& $\mathbf{95.3 \pm 0.4}$
& $\mathbf{90.6 \pm 1.8}$
& $\mathbf{84.8 \pm 1.6}$ \\
    \midrule
    \multicolumn{4}{l}{\textit{Training stages}} \\
    GRPO-200
    & $89.5 \pm 2.5$
    & $62.7 \pm 2.5$
    & $58.7 \pm 3.3$ \\
    GRPO-200 w/o SFT
    & $75.7 \pm 3.3$
    & $56.9 \pm 2.9$
    & $53.3 \pm 2.9$ \\
    SFT only
    & $80.8 \pm 2.2$
    & $64.5 \pm 2.5$
    & $56.3 \pm 3.1$ \\
    \midrule
    \multicolumn{4}{l}{\textit{Reward terms}} \\
    w/o $R_{\mathrm{format}}$
    & $58.8 \pm 2.7$
    & $27.2 \pm 3.3$
    & $29.7 \pm 2.9$ \\
    w/o $R_{\mathrm{API}}$
    & $69.9 \pm 3.3$
    & $53.9 \pm 2.9$
    & $46.4 \pm 2.9$ \\
    w/o $R_{\mathrm{exec}}$
    & $85.7 \pm 2.0$
    & $43.5 \pm 3.6$
    & $41.1 \pm 3.4$ \\
    w/o $R_{\mathrm{corr}}$
    & $82.0 \pm 2.7$
    & $70.3 \pm 2.5$
    & $40.6 \pm 4.0$ \\
    \bottomrule
    \end{tabular}
    }
\end{wraptable}
As shown in \prettyref{tab:combined_comparison}, M$^3$P-R1 achieves the best overall performance, with high SR and low runtime across both settings. PDDLStream attains higher single-modal SR (95.0\% vs.\ 91.7\%) because it operates on fully grounded symbolic specifications without language understanding, but degrades substantially on multi-modal tasks where discrete-continuous coordination is tightly coupled. LLM+P and ReAct underperform due to brittle symbolic abstractions and ungrounded reasoning. RRT* and BIT* produce shorter single-UAV paths, but our method is more effective in multi-UAV settings by jointly optimizing meeting points and smoothness, yielding shorter and smoother trajectories. These results show that solver-grounded MIP formulations better capture continuous objectives than sampling-based pipelines, and that composing discretization APIs with solver-verifiable rewards enables robust and scalable planning across diverse M$^3$P tasks. 

\begin{wraptable}{r}{0.48\textwidth}
\centering
\small
\caption{\small Expert-MIP control. The LLM selects fixed expert single-modal MIP modules and predicts the cross-modal interface variables required to compose.}
\label{tab:expert_mip}
\resizebox{\linewidth}{!}{
\begin{tabular}{lcc}
\toprule
Method & Single-modal SR (\%) & Multi-modal SR (\%) \\
\midrule
Expert-MIP + Qwen3-8B
& $\mathbf{100.0}$
& $55.6$ \\

Expert-MIP + GPT-5.2
& $\mathbf{100.0}$
& $63.2$ \\

M$^3$P-R1
& $91.7$
& $\mathbf{80.9}$ \\
\bottomrule
\end{tabular}
}
\end{wraptable}
To test whether our gains can be explained simply by access to expert optimization modules, we construct an Expert-MIP baseline in which an LLM selects fixed single-modal MIP modules and predicts only the cross-modal interface variables required to compose them. As shown in \prettyref{tab:expert_mip}, the expert modules achieve 100.0\% success on single-modal tasks with either Qwen3-8B or GPT-5.2. However, multi-modal success drops to 55.6\% and 63.2\%, respectively, compared with 80.9\% for M$^3$P-R1. Predicted meeting points, contact locations, or intermediate states can make downstream IK, grasp, or motion-planning subproblems infeasible, indicating that access to expert modules alone does not resolve the cross-modal coupling learned through solver-grounded reinforcement fine-tuning.

In \prettyref{tab:rft_ablation}, GRPO-200 (GRPO for 200 steps) improves over SFT-only training, while removing SFT degrades performance, highlighting that SFT is beneficial for tool-augmented RL~\cite{zhang2025thinking}. The \textit{Ours (800 steps)} row reports the final M$^3$P-R1 model as a reference upper bound, while all four reward ablations (\textit{w/o $R_\bullet$}, with $\bullet \in \{\text{format}, \text{API}, \text{exec}, \text{corr}\}$) share the matched 200-step GRPO budget and should be compared directly against the \textit{GRPO-200} row. Additional held-out generalization studies are provided in Appendix~\ref{appen:heldout-1} and~\ref{appen:heldout_difficulty}, including unseen modality compositions and out-of-distribution (OOD) shifts in obstacle density, agent count, and spatial-relation complexity. M$^3$P-R1 maintains strong success under these shifts, e.g., 85.7/70.5 SR on Obstacle-OOD, 89.9/75.9 SR on Agent-OOD, and 87.4/76.8 SR on Spatial-OOD for single-/multi-modal tasks, suggesting that it learns reusable MIP-programming abstractions and cross-modal constraint patterns rather than memorizing fixed templates. To further control for training budget, we rerun all reward-component ablations for the full 800 GRPO steps used by the final model. At this matched budget, removing \(R_{\mathrm{corr}}\) produces the largest degradation, reducing SR from 84.8\% to 50.3\%, while removing \(R_{\mathrm{exec}}\) and \(R_{\mathrm{API}}\) lowers SR to 70.0\% and 73.5\%, respectively. In contrast, removing \(R_{\mathrm{format}}\) has a smaller effect at convergence, yielding 78.4\% SR. These results confirm that solver-verified task correctness provides the strongest training signal, while execution and API rewards remain important for learning reliable executable formulations.

\begin{wrapfigure}{r}{0.6\textwidth}
    \centering
    \includegraphics[width=\linewidth, trim=0 3cm 0 0, clip]{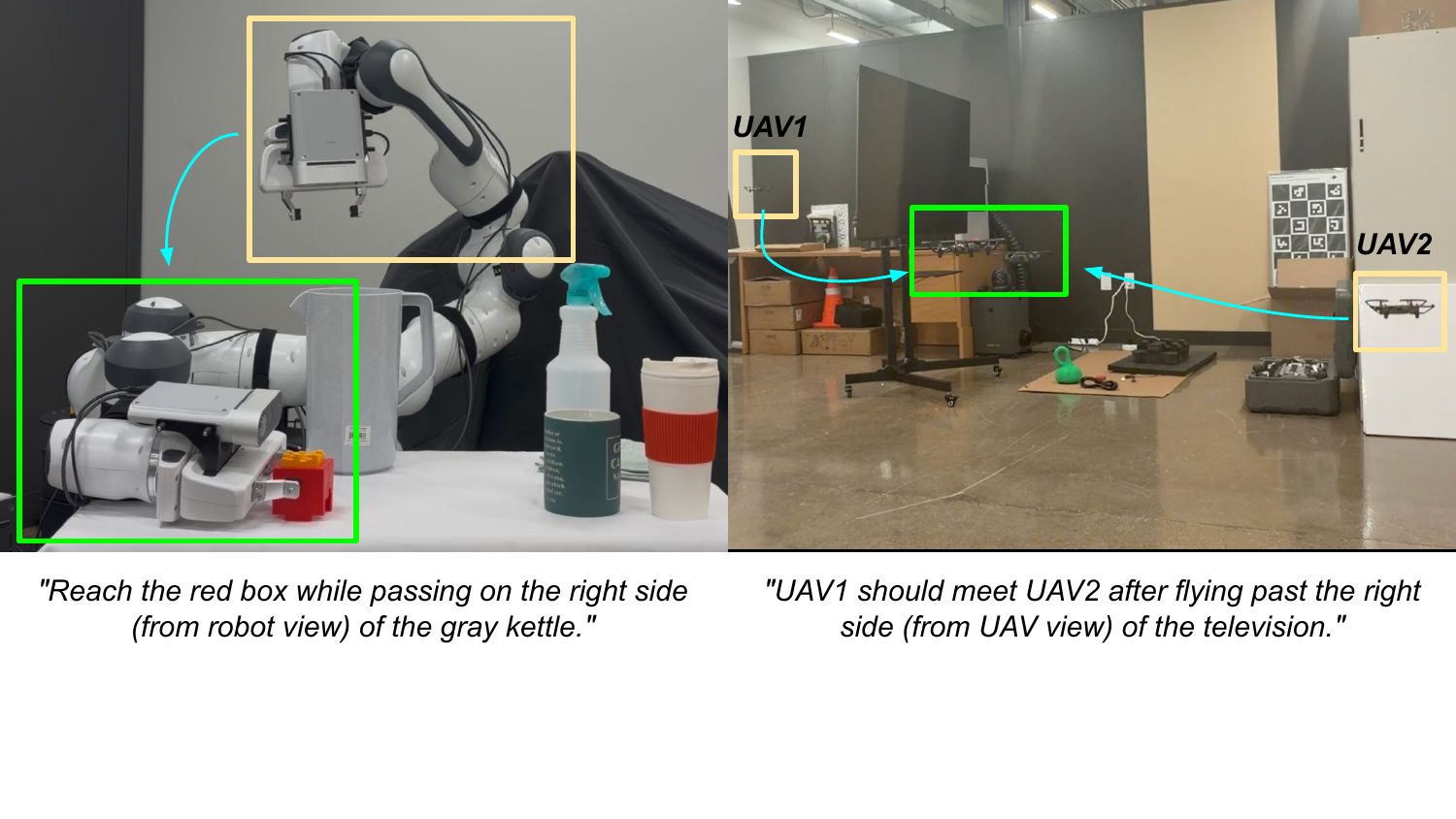}
    \caption{\small Real-world execution. (a) The robot arm performs obstacle-aware IK by following a constrained trajectory. (b) Two UAVs coordinate to satisfy spatial constraints and rendezvous. These results demonstrate reliable execution of solver-generated plans in real environments.}
    \label{fig:real}
    \vspace{-0.6em}
\end{wrapfigure}
\vspace{-1.0em}
\paragraph{Qualitative analysis}
\prettyref{fig:comparison} compares our method with representative planning baselines. Symbolic planners (e.g., PDDL) and sampling-based methods (e.g., RRT-style) often produce inefficient or zigzag trajectories due to discrete abstractions or lack of global optimality, while trajectory optimization (TrajOpt) improves smoothness but remains suboptimal under complex multi-modal constraints. We further validate our approach in physics simulation (MuJoCo), as shown in \prettyref{fig:result} and \ref{fig:supp-result} (\prettyref{appen:more_qualitative_results}). Across both single- and multi-modal tasks, our method generates feasible, collision-free, and coordinated trajectories that satisfy geometric, kinematic, and task-level constraints. Detailed per-modality MIP formulations are provided in \prettyref{appen:math_detail} (see \prettyref{appen:novel_spatial} for our novel spatial constraints and \ref{appen:math_grasp} for the grasp formulation).
\vspace{-1.0em}
\paragraph{Real-world robotic experiment}
We validate M$^3$P-R1 on real hardware using a Franka Panda robot arm and DJI Tello drones across 40 scenarios, achieving an overall success rate of 87.5\% (35/40). Failures are primarily attributed to sim-to-real mismatch (\prettyref{appen:real_world_results}). The Franka Panda operates at 1\,kHz over Ethernet, while the Tello communicates via 2.4\,GHz WiFi. As shown in \prettyref{fig:real}, our method successfully executes both (a) IK and (b) multi-UAV tasks, demonstrating that M$^3$P-R1 transfers reliably from simulation to real-world deployment.

\vspace{-1.0em}
\section{Conclusion, Limitations, and Future Work}
\vspace{-1.0em}
\paragraph{Conclusion.} We introduce M$^3$P-R1, a novel method for LLM-guided robot motion planning. Our approach generates solver-grounded MIP formulations, enabling joint reasoning over discrete and continuous decision spaces. Crucially, through GRPO with solver-verifiable rewards, the model learns to compose discretization APIs and construct MIP decompositions that solve M$^3$P tasks, rather than relying on memorized patterns. We demonstrate that M$^3$P-R1 effectively handles diverse single- and multi-modal M$^3$P tasks, a subset of TAMP. Empirically, M$^3$P-R1 reaches 91.7\%/80.9\% single-/multi-modal success rate---a 26.4/40.6 pp gain over the strongest prompt-only LLM baseline (GPT-5.2)---and transfers to real Franka+Tello hardware at 87.5\% success.
\vspace{-1.0em}
\paragraph{Limitation.}
Despite promising results, our work has several limitations that motivate future research.
First, due to the inherent limitations of MIP formulation, our pipeline is primarily designed for static environments and does not yet support dynamic scenes with moving obstacles or time-varying constraints.
Second, our study focuses on MIP-based robot planning problems already developed by human experts. A promising future direction is to enable LLMs to propose, formulate, and solve previously unmodeled M$^3$P problems, such as whole-body control and more general contact-rich robotic behaviors.



\clearpage
\acknowledgments{We thank the anonymous reviewers for their valuable feedback and the IDEAS Lab at Purdue University for providing robotic hardware support.}


\bibliography{example}  
\clearpage
\appendix

\section*{Appendix Contents}
\addcontentsline{toc}{section}{Appendix Contents}
\vspace{-0.5em}
\startcontents[appendix]
{\small\printcontents[appendix]{}{1}{}}
\clearpage

\section{\label{appen:task}Task Description}

We provide detailed descriptions of our five single-modal M$^3$P tasks:

\begin{itemize}[leftmargin=*]
\item \textbf{Footstep Planning}~\citep{deits2014footstep}:
The goal is for a bipedal robot to generate a sequence of footsteps, subject to reachability and obstacle-avoidance constraints, such that its center of mass reaches a desired 2D target position.

\item \textbf{Finger Selection}~\citep{hang2017framework}:
The goal is for a robotic gripper to select fingertip contact points that maximize the 2D grasp quality of an object.

\item \textbf{Inverse Kinematics}~\citep{dai2019global}:
The goal is for an articulated robotic arm to reach a target configuration while avoiding a set of obstacles.

\item \textbf{Collision-Free UAV Trajectory Planning (UAV)}~\citep{deits2015efficient}:
The goal is for a UAV to reach a target position without colliding with obstacles, while satisfying additional user-defined constraints. We support various constraint types, such as ``circling around an obstacle'' and ``flying from the left of an obstacle.''

\item \textbf{Grasp}~\citep{liu2020new}:
A combination of finger selection and IK, where the gripper must select contact points and compute the corresponding arm configuration to reach them.
\end{itemize}

\vspace{0.3em}
\noindent
We further provide detailed descriptions of our multi-modal M$^3$P tasks:

\begin{itemize}[leftmargin=*]

\item \textbf{Footstep+IK}:
A bipedal robot equipped with a 2D articulated arm must reach a distant target. It first walks closer to the target via footstep planning, then uses IK to reach it.

\item \textbf{UAV+IK}:
Similar to Footstep+IK, but with a UAV equipped with a 2D articulated arm, which must reach a distant target while avoiding obstacles.

\item \textbf{Footstep+UAV}:
A bipedal robot and a UAV must meet at a known or unknown location while avoiding obstacles.

\item \textbf{Multi-UAV}:
Multiple UAVs must meet at a known or unknown location while avoiding obstacles.

\item \textbf{Multi-Footstep}:
Multiple bipedal robots must meet at a known or unknown location while avoiding obstacles.

\item \textbf{Multi-IK}:
Multiple robotic arms must reach a shared target configuration or location while avoiding obstacles.

\item \textbf{Footstep+IK+Grasp}:
A bipedal robot must grasp a distant object. It first walks close to the object, then selects grasp points, and finally reaches the object using IK.

\item \textbf{UAV+IK+Grasp}:
Similar to Footstep+IK+Grasp, but with a UAV equipped with a gripper.
\end{itemize}

\section{LLMM$^3$P Bench Data Generation Pipeline}\label{appen:data_generation}
\begin{figure}[ht]
    \centering
    \includegraphics[width=\textwidth, trim=0 0 0 1.5cm,
        clip]{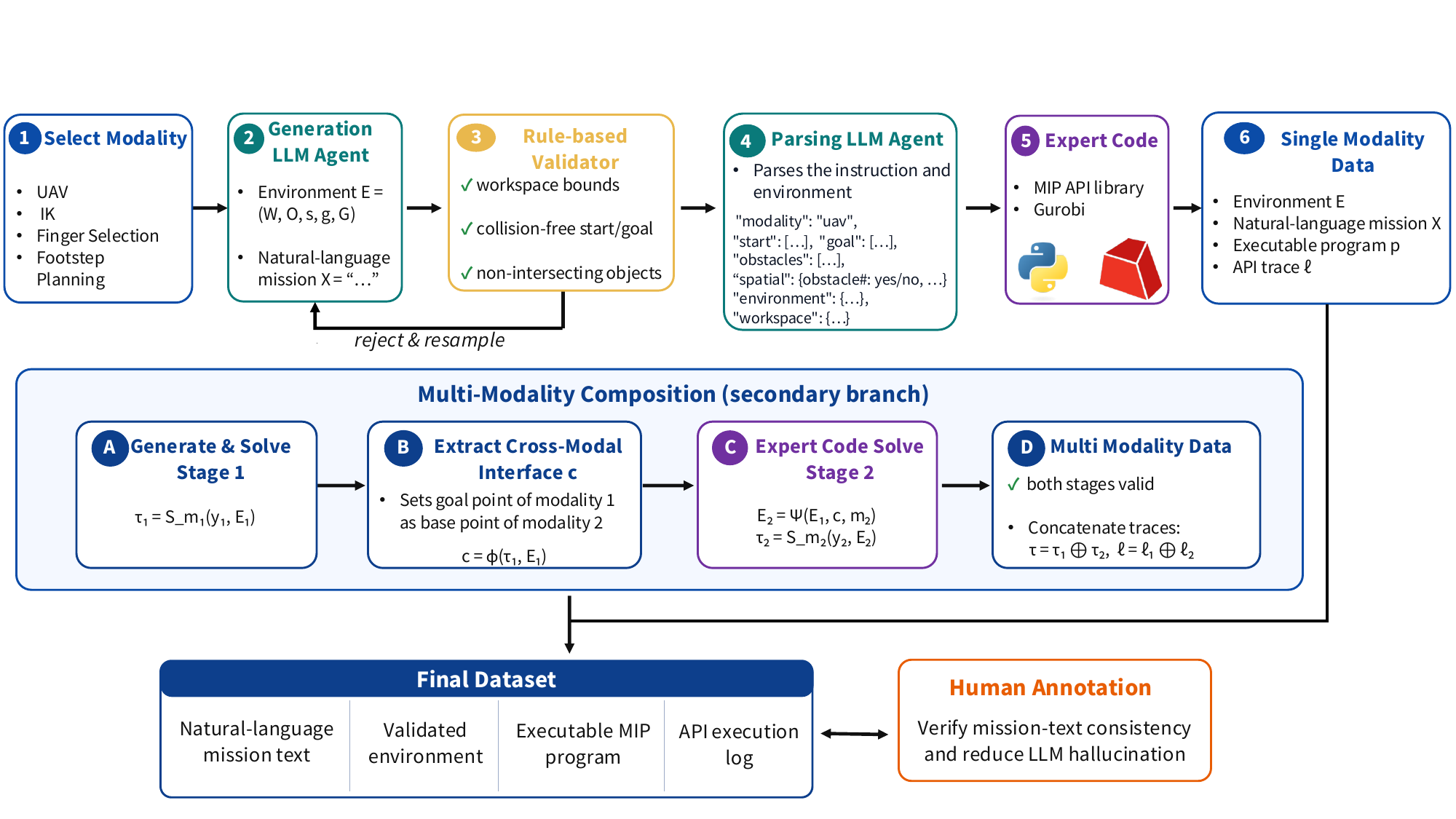}
    \vspace{-0.5em}
    \caption{
    LLMM$^3$P Bench dataset preparation pipeline.
    An agentic generation pipeline constructs single-modality M$^3$P data by sampling task environments and natural-language missions, validating geometric consistency, parsing instructions into structured representations, and solving them with modality-specific MIP expert code.
    Multi-modality tasks are generated through a secondary composition branch that couples feasible single-modality solutions via cross-modal interface constraints and concatenates their executable traces.
    Human annotation further verifies mission--data consistency before inclusion in the final dataset.
    }
    \label{fig:data_pipeline}
    \vspace{-0.5em}
\end{figure}
As illustrated in \prettyref{fig:data_pipeline}, we construct an \emph{agentic data generation pipeline} that automatically produces paired natural-language missions and solver-grounded executable programs for M$^3$P tasks. The key challenge is that expert code based on previously published papers~\cite{liu2020new,dai2019global} is only available for individual motion-planning modalities, such as UAV navigation, inverse kinematics, finger selection, and footstep planning. There is no hand-designed expert solver that directly generates complete programs for arbitrary multi-modal compositions. Therefore, our dataset is generated by combining validated single-modality experts with LLM-based task synthesis, structured grounding, and cross-modal composition.

\subsection{Single-Modal Data Generation}
For each selected modality $m \in \mathcal{M}$, where $\mathcal{M}=\{\text{UAV},\text{IK},\text{footstep},\text{finger selection},\text{grasp}\}$ denotes the supported single-modal task types, a generation LLM agent $\mathcal{A}_{\text{gen}}$ first samples a task instance
\[
\mathcal{E} = (\mathcal{W}, \mathcal{O}, s, g, \mathcal{G}_{\text{mod}}).
\]
Here, $\mathcal{W}$ denotes workspace bounds, $\mathcal{O}$ is the obstacle map, $s$ and $g$ are start and goal states, and $\mathcal{G}_{\text{mod}}$ denotes modality-specific geometry, such as robot kinematics, UAV navigation parameters, footstep terrain regions, finger-selection candidates, or grasp geometry. The same agent also generates a natural-language mission instruction $x$ that describes the task constraints, including spatial relations, footstep planning, or finger selection, and goal location. To encourage diverse instruction wording and environment variation, $\mathcal{A}_{\text{gen}}$ uses a relatively high sampling temperature, e.g., temperature $= 0.7$.

A deterministic rule-based validation module $\mathcal{V}$ then checks environment-internal geometric and semantic consistency, $\mathcal{V}(\mathcal{E}) = \mathbf{1}[\mathrm{valid}]$. Specifically, it verifies that all entities lie within the workspace bounds, start and goal states are distinct and collision-free, and objects in $\mathcal{O} \cup \mathcal{G}_{\text{mod}}$ do not intersect; cross-modal coupling is enforced separately by the interface constraint factor in the multi-modal composition stage below. Invalid samples are rejected and resampled.

Next, a parsing LLM agent $\mathcal{A}_{\text{parse}}$ (a low-temperature parser mapping a mission and environment to a typed intermediate representation) maps the instruction and environment into a typed intermediate representation, $z = \mathcal{A}_{\text{parse}}(x, \mathcal{E})$. The IR encodes modality type, start and goal references, spatial constraints, goal constraints, using a fixed schema with restricted vocabularies. This constrained representation reduces ambiguity and ensures compatibility with downstream expert code. Since this stage emphasizes reliability rather than diversity, $\mathcal{A}_{\text{parse}}$ uses a low sampling temperature, e.g., temperature $= 0.1$.

Finally, let $\{\mathcal{S}_m\}_{m\in\mathcal{M}}$ denote the family of modality-specific instantiations of the generic solver $\mathcal{S}$ from~\prettyref{sec:training}. A solver module $\mathcal{S}_m$ constructs and solves the corresponding MIP, returning the pair $(\tau, \mathcal{L}) = \mathcal{S}_m(y, \mathcal{E})$, where $\tau$ is the resulting trajectory and $\mathcal{L}$ is the API execution log, and $y$ is the executable program assembled from the IR $z$ together with the modality-$m$ expert code template, using modality-specific expert formulations implemented through a Python library of callable MIP-discretization APIs and Gurobi. For each solved instance, we record the full executable program, the environment parameters, the resulting trajectory $\tau$, and the executed API sequence in \texttt{run\_log.txt}. This log provides an executable trace of the solver-grounded solution and is later used to compute the API reward $R_{\mathrm{API}}$ during GRPO.

Within this section the symbols $z$, $y$, and $\mathcal{L}$ have fixed meanings (typed intermediate representation, executable MIP program, and API execution log, respectively) and should not be confused with same-letter symbols used locally inside the modality subsections of \prettyref{appen:math_detail}. Each single-modality data point is therefore represented as the tuple $(x, \mathcal{E}, z, y, \tau, \mathcal{L})$, where $z$ is the typed intermediate representation, $y$ is the executable MIP program, and $\mathcal{L}$ is the API execution log.

\subsection{Multi-Modal Composition}
For multi-modal motion-planning tasks, direct expert code is not available. Instead, we construct compositional tasks by coupling multiple validated single-modal tasks. Consider a two-stage composition with modalities $m_1$ and $m_2$. Let $\mathbb{T}$ denote the space of feasible trajectories, $\mathbb{E}$ the space of task environments defined above, and $\mathcal{C}$ the space of cross-modal interface constraints (e.g., shared waypoints or contact poses). We first generate and solve a valid single-modal instance, $(\tau_1, \mathcal{L}_1) = \mathcal{S}_{m_1}(y_1, \mathcal{E}_1)$. The terminal state or selected waypoint of the first solution is then used to define the start state, goal state, rendezvous point, or base pose for the second modality via $\iota = \phi(\tau_1, \mathcal{E}_1)$, where $\phi: \mathbb{T}\times\mathbb{E} \to \mathcal{C}$ is an interface-extraction map that selects a coupling waypoint or pose from the first-stage trajectory and environment, and $\iota\in\mathcal{C}$ denotes the resulting cross-modal interface constraint. We then instantiate the second modality conditioned on this interface:
\[
\mathcal{E}_2 = \Psi(\mathcal{E}_1, \iota, m_2),
\qquad
(\tau_2, \mathcal{L}_2) = \mathcal{S}_{m_2}(y_2, \mathcal{E}_2),
\]
where $\Psi: \mathbb{E}\times\mathcal{C}\times\mathcal{M} \to \mathbb{E}$ instantiates a modality-$m_2$ task environment by extending $\mathcal{E}_1$ with the interface $\iota$ and any modality-$m_2$-specific geometry.
Only compositions for which both stages are feasible and the interface constraint is satisfied are retained:
\[
\mathbf{1}[\mathrm{valid}] =
\mathcal{V}(\mathcal{E}_1)
\cdot
\mathcal{V}(\mathcal{E}_2)
\cdot
\mathbf{1}[C_{\text{iface}}(\tau_1, \tau_2, \iota) = 1].
\]
Here $C_{\text{iface}}(\tau_1, \tau_2, \iota) \in \{0,1\}$ is a predicate indicating whether the trajectories $\tau_1$ and $\tau_2$ satisfy the interface constraint $\iota$. $\mathcal{V}$ checks only environment-internal validity (workspace bounds, obstacle non-intersection, distinct start/goal), while cross-modal coupling is enforced by the separate factor $\mathbf{1}[C_{\text{iface}}(\tau_1, \tau_2, \iota) = 1]$.

For example, in a UAV--IK task, the UAV expert first solves a navigation problem and produces a terminal position. This terminal position is then used as the base pose for the IK problem, which is solved by the IK expert code. The final multi-modal solution is obtained by concatenating the two executable traces, $\tau = \tau_1 \oplus \tau_2$ and $\mathcal{L} = \mathcal{L}_1 \oplus \mathcal{L}_2$ (where $\oplus$ denotes sequential concatenation of trajectories, resp.\ of API-call logs), where the combined \texttt{run\_log.txt} contains the API calls and trajectory-generation steps for both modalities. This procedure guarantees feasibility of the constructed multi-modal instance, which is sufficient for fine-tuning and solver-grounded evaluation. The concatenated trajectory $\tau$ need not be globally optimal for the full composed task, since the model is not rewarded for matching the reference trajectory exactly, but for generating an executable program that satisfies the mission constraints. To balance diversity and solve efficiency, we sample moderate-sized environments with 1--3 agents and 2--8 obstacles.

Lastly, human annotators verify that every data sample matches the textual constraints in the mission instruction, ensuring data validity and reducing the effects of possible LLM hallucinations. The resulting dataset contains natural-language missions, validated environments, executable MIP programs, and API execution logs for both single- and multi-modal M$^3$P tasks. We showcase data samples in LLMM$^3$P Bench in \prettyref{appen:data_example}.

\section{More Quantitative Results}\label{appen:more_quantitative_results}

\subsection{Effect of In-Context Examples}

\begin{figure}[t]
    \centering
    \includegraphics[width=\linewidth]{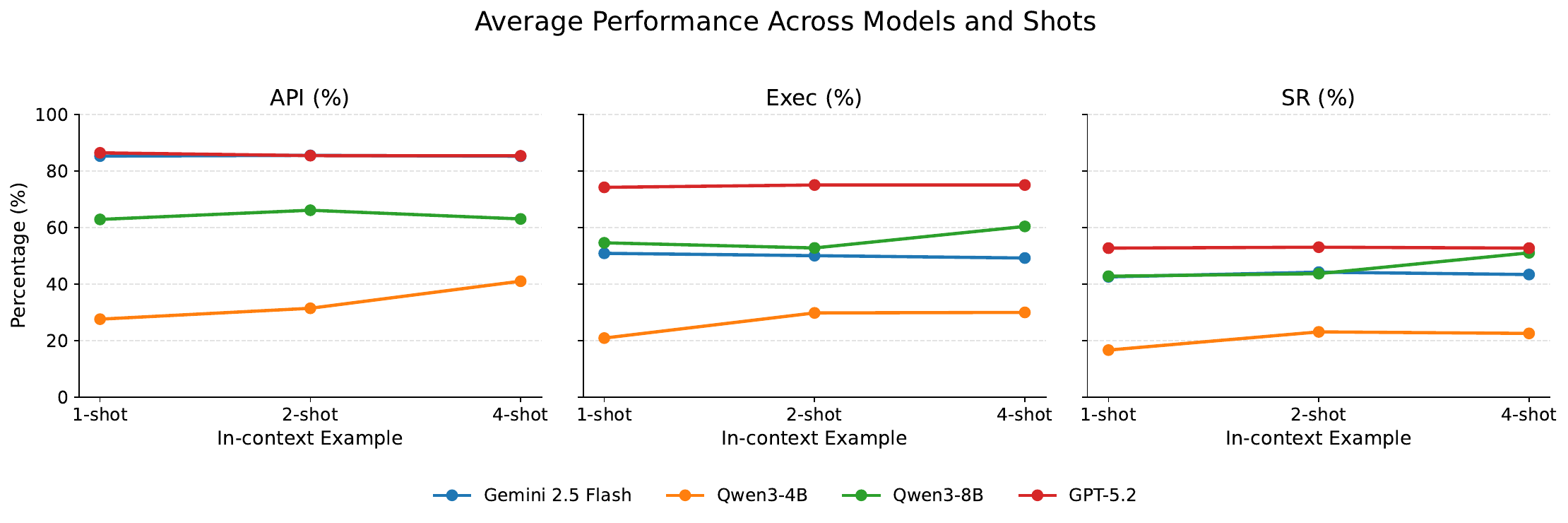}
    \caption{
    Effect of increasing the number of in-context examples for direct prompting. We evaluate 1/2/4-shot prompting using only single-modal demonstrations and test the resulting prompts on both single- and multi-modality M$^3$P tasks. Each panel reports one metric (API \%, Exec \%, or SR \%) on the y-axis; the x-axis sweeps the number of in-context shots (1/2/4), and colors denote different LLMs. Across different LLMs, increasing the number of demonstrations does not consistently improve API accuracy, execution success, or task correctness, indicating that M$^3$P reasoning cannot be reliably induced through prompting alone.
    }
    \label{fig:icl_shot_comparison}
\end{figure}

\prettyref{fig:icl_shot_comparison} compares the performance of different LLMs under varying numbers of in-context demonstrations. We compare 1/2/4 shots of in-context examples (single-modality) and observe that in-context learning does not scale for our tasks. While larger proprietary models such as GPT-5.2 and Gemini-2.5-Flash achieve relatively strong performance, additional examples provide only marginal or inconsistent improvements. Smaller open-source models exhibit larger variance and limited gains from prompting alone. Overall, the results suggest that M$^3$P tasks require structured solver-grounded reasoning beyond simple in-context learning, motivating the need for dedicated fine-tuning.

\subsection{Effect of GRPO Training Steps}

\begin{figure}[t]
    \centering
    \includegraphics[width=0.75\linewidth]{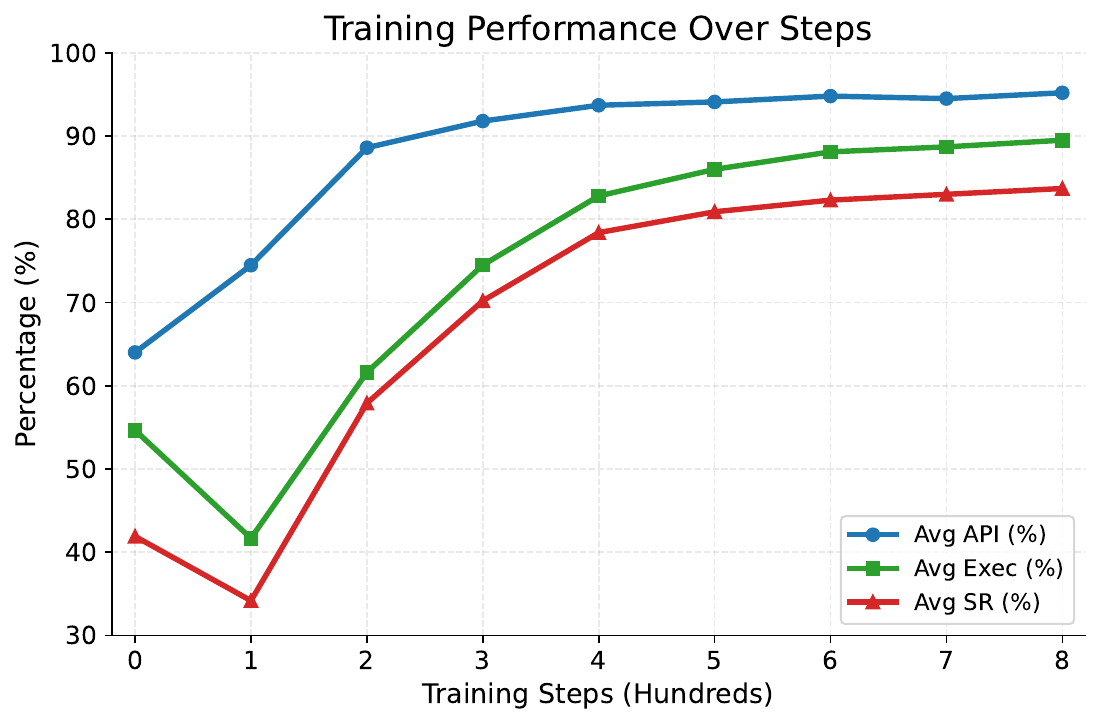}
    \caption{
    Performance improvement across different GRPO training steps. API accuracy, execution success, and task correctness improve rapidly during the early stages of training and gradually saturate after approximately 500 steps, with only minor fluctuations afterward.
    }
    \label{fig:rft_training_trend}
\end{figure}

\prettyref{fig:rft_training_trend} illustrates the effect of GRPO optimization on M$^3$P performance. The model shows substantial gains in API accuracy, execution success, and task correctness during the first several hundred training steps, demonstrating that GRPO effectively improves solver-grounded reasoning abilities. After around 500 training steps, the performance begins to stabilize and approaches saturation, indicating convergence toward a robust policy for structured robotic planning.

\subsection{Detailed Per-Modality Performance}
\begin{figure}[t]
    \centering
    \includegraphics[width=\linewidth]{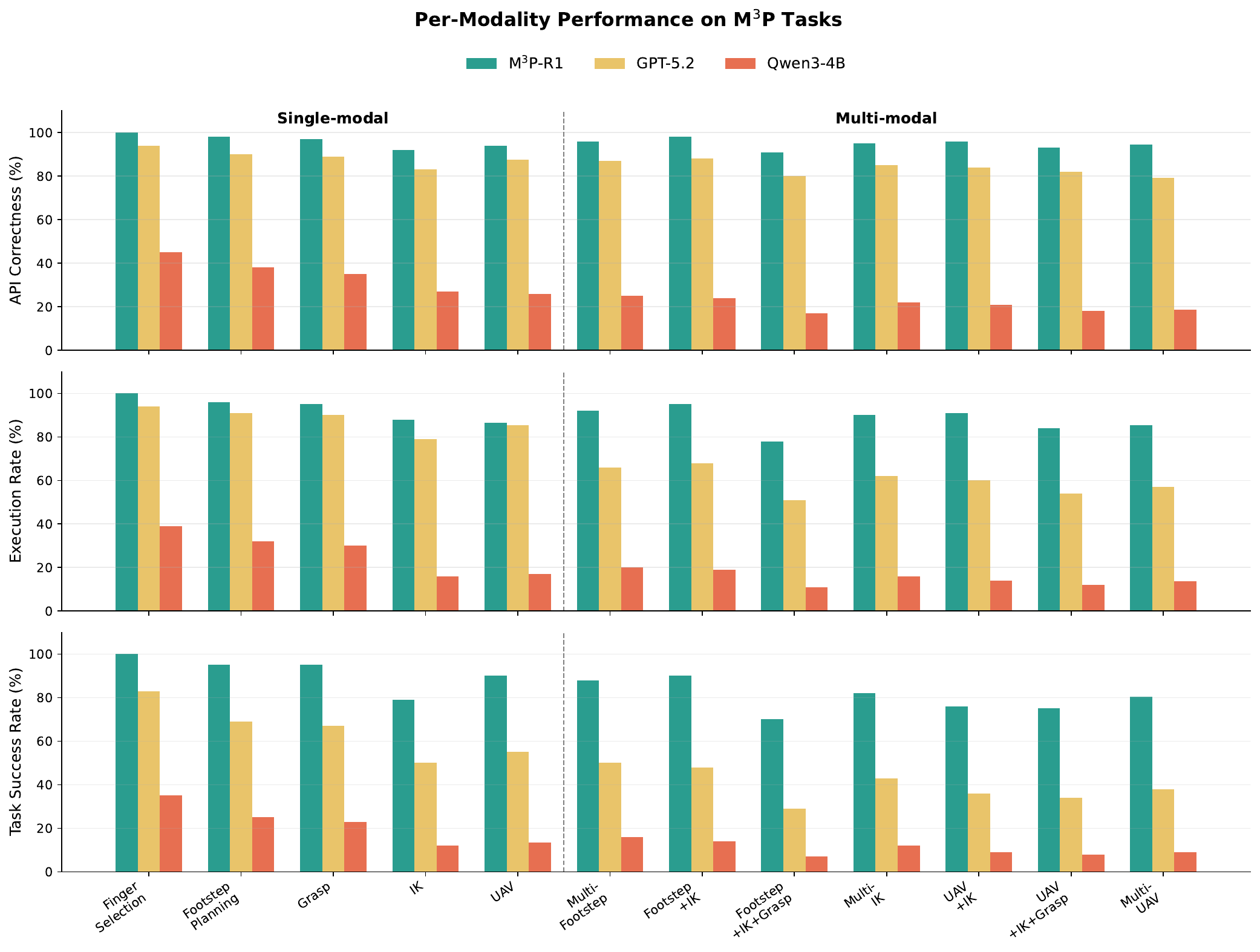}
    \caption{\small Per-modality performance on single- and multi-modality M$^3$P tasks. We compare M$^3$P-R1 with GPT-5.2 and Qwen3-4B across API correctness, execution rate, and task success rate. M$^3$P-R1 maintains strong performance across both single-modal and multi-modal planning domains, with the largest gains appearing on complex multi-modal tasks.}
    \label{fig:per_modality_results}
    \vspace{-1.5em}
\end{figure}

\prettyref{fig:per_modality_results} provides a detailed performance breakdown across different single- and multi-modality M$^3$P tasks. Overall, M$^3$P-R1 achieves strong and consistent performance on single-modality planning problems, with near-perfect results on \texttt{finger selection}, \texttt{footstep planning}, and \texttt{grasp}. More challenging continuous optimization tasks, such as \texttt{IK} and \texttt{UAV}, show lower execution and task success rates due to the increased difficulty of geometric reasoning, collision avoidance, and constraint satisfaction.

Performance on multi-modality tasks remains strong but decreases compared with single-modality settings, especially for tasks requiring joint reasoning over multiple coupled decision spaces. In particular, \texttt{footstep+ik+grasp} and \texttt{uav+ik+grasp} produce the lowest execution and task success rates because the model must simultaneously coordinate locomotion, inverse kinematics, grasp feasibility, and collision-free planning within a unified MIP formulation. Nevertheless, M$^3$P-R1 consistently outperforms GPT-5.2 and Qwen3-4B across all modalities. The gap becomes larger as task complexity increases, especially in execution success and task correctness for multi-modal planning problems. As shown in \prettyref{fig:per_modality_results}, general-purpose LLM baselines experience substantial degradation when solving complex multi-modal tasks, whereas M$^3$P-R1 maintains relatively stable API correctness and execution performance across diverse planning domains. These results suggest that RFT enables the model to learn transferable solver-grounded planning strategies rather than modality-specific program templates.

\subsection{Structured Held-Out Generalization}\label{appen:heldout-1}

\paragraph{Note on independent runs and ID test reconciliation.}
Each held-out split below entails a fully independent SFT+GRPO training run with the corresponding subset of tasks excluded; the \emph{Multi-modal Seen} rows therefore vary across \prettyref{tab:heldout_composition}, \ref{tab:heldout_grasp}, and \ref{tab:heldout_uav} because each row reflects a separate training run. For the same reason, the in-distribution (ID) test single-/multi-modal SR reported in this section (91.8\%/80.2\%; \prettyref{tab:heldout_llm_metrics}) differs slightly from the headline 91.7\%/80.9\% in \prettyref{tab:llm_avg_performance}: the headline number is averaged over three independent training runs, while the held-out tables report a single training run per row. Differences smaller than the ID-test run-to-run std (approximately 1.5 pp) should therefore not be over-interpreted.

To assess whether M$^3$P-R1 learns reusable solver-grounded planning strategies rather than memorizing fixed program templates, we evaluate three structured held-out settings. In the first setting, the complex compositions \texttt{uav+ik+grasp} and \texttt{footstep+ik+grasp} are excluded during RFT training. In the second and third settings, we remove all grasp-related or UAV-related tasks from fine-tuning, respectively, and evaluate them only at test time. These splits are substantially harder than random test splits because they require the model to compose MIP primitives and cross-modal interfaces in unseen configurations. Note that this setting evaluates ID generalization to held-out structural splits, as opposed to the out-of-distribution (OOD) difficulty-axis generalization in \prettyref{appen:heldout_difficulty}.
\paragraph{Metrics.}
We evaluate generated programs using both LLM-level and robotics-level metrics. API measures whether the generated code uses valid discretization APIs with correct argument structure. Exec measures whether the generated program is syntactically valid and successfully invokes the optimization solver. SR measures full task success, requiring the executed solution to satisfy the complete task specification. For robotics-level evaluation, Traj. SR measures whether the returned trajectory satisfies start and goal constraints, continuity, region membership, and robot-specific feasibility constraints. Sem. SR measures whether the trajectory satisfies the language-specified spatial relations and task semantics like robot meet. Collision-free rate measures the percentage of executed trajectories that avoid all obstacles. Path length is normalized by the ground-truth expert MIP solution, with $1.0$ corresponding to expert-level path efficiency.

As shown in \prettyref{tab:heldout_composition}, \ref{tab:heldout_grasp}, and \ref{tab:heldout_uav}, performance decreases compared with the fully fine-tuned setting, especially in execution success and task success rate. This degradation is expected because the model must solve unseen modality compositions or reason about held-out robotic primitives without direct fine-tuning examples. Nevertheless, M$^3$P-R1 remains consistently stronger than \textbf{GPT-5.2}, the strongest non-fine-tuned LLM baseline in \prettyref{tab:llm_avg_performance} (the same model is used across all held-out rows). On held-out composition tasks, M$^3$P-R1 improves over GPT-5.2 by $+7.2\%$ API accuracy, $+18.1\%$ execution success, and $+22.8\%$ task success rate. Even in the more challenging held-out primitive settings, it still achieves gains of $+2.6\%$/$+18.8\%$/$+16.0\%$ on grasp-related tasks and $+3.5\%$/$+30.9\%$/$+25.5\%$ on UAV-related tasks. These results suggest that the model is not simply memorizing training templates. Instead, RFT helps it learn transferable robotic MIP formulation patterns, including API composition, solver invocation, and cross-modal constraint grounding.

\begin{table}[t]
    \centering
    \vspace{-0em}
    \caption{\small Held-out composition generalization during RFT. Complex compositions (\texttt{uav+ik+grasp} and \texttt{footstep+ik+grasp}) are excluded from RFT training but evaluated during testing. We compare against GPT-5.2, the strongest non-fine-tuned LLM baseline (same model used across all held-out rows).}
    \label{tab:heldout_composition}
    \resizebox{\linewidth}{!}{
    \begin{tabular}{lccccccccc}
    \toprule
    & \multicolumn{3}{c}{Ours (\%)}
    & \multicolumn{3}{c}{Decrease (\%)}
    & \multicolumn{3}{c}{Gain over Best LLM (\%)} \\
    \cmidrule(lr){2-4} \cmidrule(lr){5-7} \cmidrule(lr){8-10}
    Setting & API & Exec & SR & API & Exec & SR & API & Exec & SR \\
    \midrule
    Single-modal
    & 96.8 & 92.4 & 92.1
    & +0.6 & -0.7 & +0.3
    & +8.1 & +4.5 & +27.3 \\

    Multi-modal Seen
    & 92.4 & 85.4 & 76.9
    & -3.5 & -5.3 & -6.4
    & +8.8 & +25.7 & +37.2 \\

    Held-out Composition
    & 88.6 & 71.8 & 60.2
    & -3.4 & -9.2 & -12.3
    & +7.2 & +18.1 & +22.8 \\
    \bottomrule
    \end{tabular}
    }
    \vspace{-0em}
\end{table}

\begin{table}[t]
    \centering
    \vspace{-0em}
    \caption{\small Held-out grasp generalization. Grasp-related tasks are excluded from both SFT and RFT training and evaluated only during testing. We compare against GPT-5.2, the strongest non-fine-tuned LLM baseline (same model used across all held-out rows).}
    \label{tab:heldout_grasp}
    \resizebox{\linewidth}{!}{
    \begin{tabular}{lccccccccc}
    \toprule
    & \multicolumn{3}{c}{Ours (\%)}
    & \multicolumn{3}{c}{Decrease (\%)}
    & \multicolumn{3}{c}{Gain over Best LLM (\%)} \\
    \cmidrule(lr){2-4} \cmidrule(lr){5-7} \cmidrule(lr){8-10}
    Setting & API & Exec & SR & API & Exec & SR & API & Exec & SR \\
    \midrule
    Single-modal Seen
    & 96.3 & 92.1 & 91.4
    & +0.3 & -0.5 & +0.4
    & +7.6 & +4.2 & +26.6 \\

    Multi-modal Seen
    & 89.1 & 82.6 & 73.8
    & -6.8 & -8.1 & -9.5
    & +5.5 & +22.9 & +34.1 \\

    Held-out Grasp Tasks
    & 84.2 & 63.5 & 50.7
    & -9.5 & -22.2 & -29.3
    & +2.6 & +18.8 & +16.0 \\
    \bottomrule
    \end{tabular}
    }
    \vspace{-0em}
\end{table}

\begin{table}[t]
    \centering
    \vspace{-0em}
    \caption{\small Held-out UAV generalization. UAV-related tasks are excluded from both SFT and RFT training and evaluated only during testing. We compare against GPT-5.2, the strongest non-fine-tuned LLM baseline (same model used across all held-out rows).}
    \label{tab:heldout_uav}
    \resizebox{\linewidth}{!}{
    \begin{tabular}{lccccccccc}
    \toprule
    & \multicolumn{3}{c}{Ours (\%)}
    & \multicolumn{3}{c}{Decrease (\%)}
    & \multicolumn{3}{c}{Gain over Best LLM (\%)} \\
    \cmidrule(lr){2-4} \cmidrule(lr){5-7} \cmidrule(lr){8-10}
    Setting & API & Exec & SR & API & Exec & SR & API & Exec & SR \\
    \midrule
    Single-modal Seen
    & 95.4 & 94.2 & 89.8
    & -1.4 & -0.6 & -2.5
    & +6.7 & +6.3 & +25.0 \\

    Multi-modal Seen
    & 91.8 & 82.4 & 74.1
    & -3.2 & -6.4 & -8.4
    & +8.2 & +22.7 & +34.4 \\

    Held-out UAV Tasks
    & 84.1 & 62.6 & 55.2
    & -10.3 & -24.1 & -25.2
    & +3.5 & +30.9 & +25.5 \\
    \bottomrule
    \end{tabular}
    }
    \vspace{-0em}
\end{table}

\prettyref{tab:heldout_robotics_metrics_modality} further evaluates whether the generated programs remain physically feasible and semantically aligned under structured held-out modality shifts. Among the three settings, Held-out Composition is the most challenging, achieving $82.0\%$ trajectory success, $75.3\%$ semantic success, and $80.0\%$ collision-free rate, while also producing the longest normalized paths ($1.68$). This degradation is expected because the model must compose unseen cross-modal interfaces and MIP primitives without direct RFT supervision. Held-out Grasp Tasks achieve $78.2\%$ trajectory success and $76.9\%$ collision-free rate, indicating that the model can still construct physically valid grasp-related optimization programs despite never observing grasp tasks during fine-tuning. Held-out UAV Tasks produce the lowest trajectory success and collision-free rate, with $76.1\%$ and $74.3\%$, respectively, reflecting the additional geometric and motion-planning complexity of UAV dynamics. Nevertheless, all held-out settings maintain relatively high semantic success and feasible trajectory generation, demonstrating that M$^3$P-R1 learns transferable solver-grounded planning strategies rather than memorizing fixed modality-specific templates.

\begin{table}[th]
\centering
\caption{\small Robotics-level metrics on held-out modality splits. Traj. SR measures physical trajectory feasibility, while Sem. SR measures satisfaction of language-specified spatial and task-level requirements. Path length is normalized by the expert MIP solution, where lower is better.}
\label{tab:heldout_robotics_metrics_modality}
\resizebox{\linewidth}{!}{
\begin{tabular}{lcccc}
\toprule
Evaluation split
& Traj. SR (\%)
& Sem. SR (\%)
& Collision-free (\%)
& Path length $\downarrow$\\
\midrule
Held-out Composition
& 82.0 & 75.3 & 80.0 & 1.68 \\

Held-out Grasp Tasks
& 78.2 & 74.4 & 76.9 & 1.12 \\

Held-out UAV Tasks
& 76.1 & 75.6 & 74.3 & 1.55 \\
\bottomrule
\end{tabular}
}
\end{table}

\subsection{Held-Out Difficulty Generalization}
\label{appen:heldout_difficulty}

We further evaluate whether M$^3$P-R1 generalizes to task configurations that are systematically harder than those observed during fine-tuning. This study constructs controlled held-out difficulty splits along three axes: obstacle density, agent count, and spatial-relation complexity. These factors directly affect geometric planning difficulty, multi-agent coordination, constraint construction, and language-grounded task satisfaction.

We define obstacle difficulty by the number of obstacles in the environment. Simple tasks contain $0$--$2$ obstacles, medium tasks contain $3$--$5$ obstacles, and hard tasks contain $6$--$8$ obstacles. Agent difficulty is defined by the number of simultaneously planned agents, ranging from one to three. Spatial difficulty is defined by the number and diversity of spatial relations specified in the language instruction. We consider five spatial relation types: \texttt{top}, \texttt{bottom}, \texttt{left}, \texttt{right}, and \texttt{circle}. Each obstacle can be associated with at most one spatial relation.

\prettyref{tab:heldout_setup} summarizes the held-out difficulty splits. The ID test split follows the original benchmark distribution and serves as the in-distribution reference. In Obstacle-OOD, training contains only simple and medium obstacle layouts ($0$--$5$ obstacles), while testing evaluates hard layouts with $6$--$8$ obstacles. Agent-OOD trains on one- and two-agent tasks and evaluates three-agent coordination. Spatial-OOD trains on instructions with at most one spatial constraint and tests denser language-grounded specifications with $2$--$8$ spatial constraints.

\begin{table}[t]
\centering
\caption{\small Held-out difficulty splits for evaluating out-of-distribution generalization. ID denotes the original in-distribution test setting. Each OOD split removes the target difficulty regime from both SFT and RFT training and evaluates only on the held-out regime.}
\label{tab:heldout_setup}
\resizebox{\linewidth}{!}{
\begin{tabular}{lcccccc}
\toprule
Split
& Train obstacles
& Test obstacles
& Train agents
& Test agents
& Train spatial constraints
& Test spatial constraints \\
\midrule
ID test
& $0$--$8$
& $0$--$8$
& $1$--$3$
& $1$--$3$
& $0$--$8$ constraints
& $0$--$8$ constraints \\

Obstacle-OOD
& $0$--$5$
& $6$--$8$
& $1$--$3$
& $1$--$3$
& $0$--$8$ constraints
& $0$--$8$ constraints \\

Agent-OOD
& $0$--$8$
& $0$--$8$
& $1$--$2$
& $3$
& $0$--$8$ constraints
& $0$--$8$ constraints \\

Spatial-OOD
& $0$--$8$
& $0$--$8$
& $1$--$3$
& $1$--$3$
& $\leq 1$ constraint
& $2$--$8$ constraints \\
\bottomrule
\end{tabular}
}
\end{table}

\begin{table}[t]
\centering
\caption{\small Held-out difficulty generalization of M$^3$P-R1. ID denotes the original in-distribution test set. OOD splits evaluate extrapolation to harder obstacle, agent, spatial, and compositional settings. Gain is computed over GPT-5.2, the strongest non-fine-tuned LLM baseline (same model used across all rows).}
\label{tab:heldout_llm_metrics}
\resizebox{\linewidth}{!}{
\begin{tabular}{lcccccccccccc}
\toprule
& \multicolumn{3}{c}{Single-modal Ours (\%)}
& \multicolumn{3}{c}{Multi-modal Ours (\%)}
& \multicolumn{3}{c}{Single-modal Gain (\%)}
& \multicolumn{3}{c}{Multi-modal Gain (\%)} \\
\cmidrule(lr){2-4} \cmidrule(lr){5-7}
\cmidrule(lr){8-10} \cmidrule(lr){11-13}
Evaluation split
& API & Exec & SR
& API & Exec & SR
& API & Exec & SR
& API & Exec & SR \\
\midrule
ID test
& 96.2 & 93.1 & 91.8
& 94.8 & 87.9 & 80.2
& +7.5 & +5.2 & +27.0
& +11.2 & +28.2 & +40.5 \\

Obstacle-OOD
& 93.8 & 89.6 & 85.7
& 91.8 & 83.4 & 70.5
& +5.1 & +1.7 & +20.9
& +8.2 & +23.7 & +30.8 \\

Agent-OOD
& 94.3 & 90.8 & 89.9
& 92.6 & 84.7 & 75.9
& +5.6 & +2.9 & +25.1
& +9.0 & +25.0 & +36.2 \\

Spatial-OOD
& 94.6 & 91.2 & 87.4
& 93.1 & 85.6 & 76.8
& +5.9 & +3.3 & +22.6
& +9.5 & +25.9 & +37.1 \\

\bottomrule
\end{tabular}
}
\end{table}

\prettyref{tab:heldout_llm_metrics} reports LLM-level program-generation performance under these held-out shifts. On the ID split, M$^3$P-R1 achieves $91.8\%$ and $80.2\%$ SR for single-modal and multi-modal tasks, respectively. Performance decreases gradually under held-out settings rather than collapsing. Obstacle-OOD is the most challenging split, reducing SR to $85.7\%$ and $70.5\%$, reflecting the increased difficulty of dense geometric reasoning. Agent-OOD achieves $89.9\%$ and $75.9\%$ SR, while Spatial-OOD achieves $87.4\%$ and $76.8\%$ SR, demonstrating robust generalization to harder coordination and denser spatial constraints. Across all held-out splits, M$^3$P-R1 consistently outperforms GPT-5.2 (the strongest non-fine-tuned LLM baseline; same model used across all rows) by large margins.

\begin{table}[th]
\centering
\caption{\small Robotics-level metrics on held-out difficulty splits. Traj. SR measures physical trajectory feasibility, while Sem. SR measures satisfaction of language-specified spatial and task-level requirements. Path length is normalized by the expert MIP solution, where lower is better.}
\label{tab:heldout_robotics_metrics}
\resizebox{\linewidth}{!}{
\begin{tabular}{lcccc}
\toprule
Evaluation split
& Traj. SR (\%)
& Sem. SR (\%)
& Collision-free (\%)
& Path length $\downarrow$\\
\midrule
ID test
& 87.9 & 89.3 & 95.8 & 1.11 \\

Obstacle-OOD
& 78.2 & 86.4 & 81.5 & 1.31 \\

Agent-OOD
& 83.1 & 85.6 & 89.8 & 1.20 \\

Spatial-OOD
& 86.1 & 84.7 & 92.3 & 1.07 \\
\bottomrule
\end{tabular}
}
\end{table}

\prettyref{tab:heldout_robotics_metrics} evaluates whether the executed solutions remain physically meaningful under the same shifts. On the ID split, M$^3$P-R1 achieves $87.9\%$ trajectory success, $89.3\%$ semantic success, and $95.8\%$ collision-free rate. Obstacle-OOD produces the largest degradation, reducing trajectory success to $78.2\%$ and increasing normalized path length to $1.31$, indicating that dense obstacle layouts mainly challenge geometric feasibility and path efficiency. Agent-OOD primarily stresses multi-agent coordination, while Spatial-OOD mainly affects semantic alignment. Nevertheless, the generated programs remain largely feasible, safe, and semantically aligned across all held-out settings.

Taken together, \prettyref{tab:heldout_llm_metrics} and \ref{tab:heldout_robotics_metrics} show that M$^3$P-R1 generalizes beyond the difficulty regimes observed during training. Although held-out obstacle, agent, and spatial shifts reduce performance, the model maintains high program validity, execution success, trajectory feasibility, and collision-free performance. These results suggest that solver-grounded fine-tuning enables the model to learn reusable MIP-programming abstractions that transfer to harder planning configurations.

\subsection{Failure Mode Analysis}

\begin{wrapfigure}{r}{0.52\linewidth}
    \centering
    \includegraphics[
        width=\linewidth,
        trim={0.55in 0.95in 0.17in 0.95in},
        clip
    ]{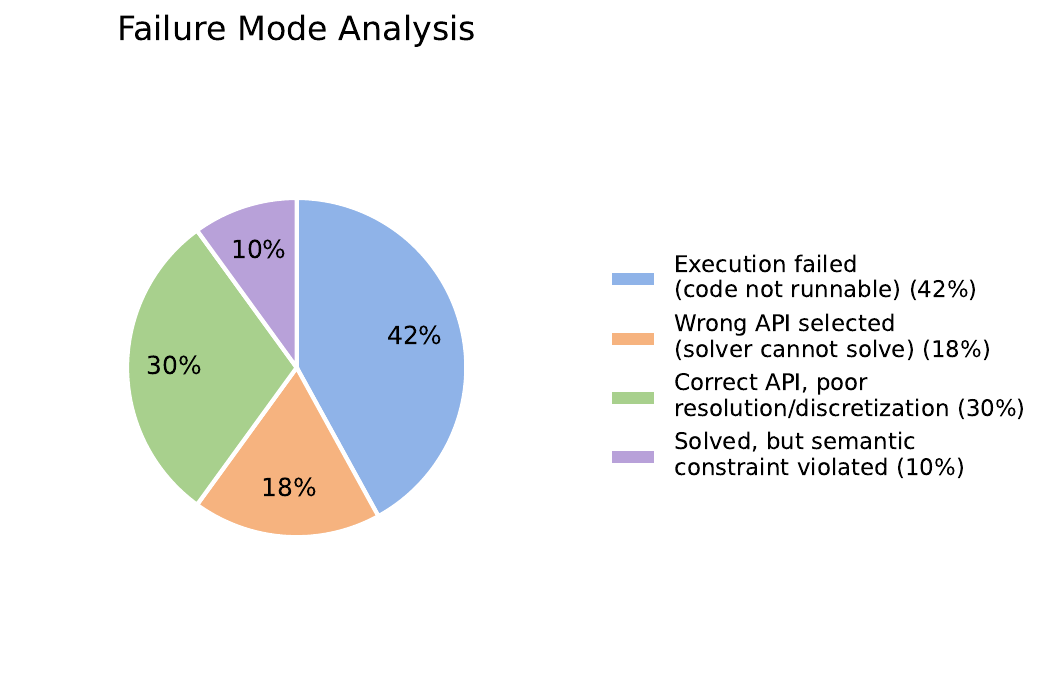}
    \caption{\small Failure mode analysis of unsuccessful M$^3$P-R1 outputs across all single- and multi-modal tasks. Most failures occur before obtaining a valid task solution, either because the generated code is not executable or because the resulting optimization problem is not solver-compatible. Semantic failures after successful execution and solving are comparatively rare.}
    \label{fig:failure_mode_analysis}
\end{wrapfigure}

To better understand the remaining limitations of M$^3$P-R1, we categorize unsuccessful outputs into four mutually exclusive failure modes, as shown in \prettyref{fig:failure_mode_analysis}. The largest category is execution failure (42\%), where the generated program is not runnable due to syntax errors, undefined variables, malformed solver calls, or incompatible API signatures. The second major category is correct API selection with insufficient discretization resolution (30\%), where the model selects the appropriate planning primitive but provides suboptimal arguments, such as too few IRIS regions or an overly coarse convex decomposition, making the resulting MIP difficult or impossible to solve. Another 18\% of failures come from incorrect API selection, for example choosing a rotation-discretization constraint for a UAV navigation task, which leads to an invalid or unsolvable optimization formulation. Finally, only 10\% of failures are semantic failures after successful execution and solving, such as satisfying collision constraints but violating the intended spatial style, e.g., flying around the wrong side of an obstacle or mismatching the specified start/end relation. These four categories are scoped to the simulation evaluation; for the orthogonal real-world failure taxonomy (sim-to-real mesh mismatch, Tello drift, wrist singularities), see the per-failure breakdown in \prettyref{appen:real_world_results}.

This analysis suggests that the dominant remaining challenges are formulation robustness and solver compatibility rather than high-level task understanding. In particular, 90\% of failures occur before a semantically valid solved trajectory can be produced, while only a small fraction correspond to incorrect task intent after successful solving. This indicates that once M$^3$P-R1 constructs an executable and solvable MIP program, it usually preserves the intended task constraints. These findings also point to concrete directions for improving robustness, including stricter API signature checking, automatic repair of non-runnable code, and adaptive discretization or region-refinement strategies for difficult planning instances.

\section{More Qualitative Results}\label{appen:more_qualitative_results}
\subsection{2D Visualization}
\begin{figure}[ht]
    \centering
    \includegraphics[
        width=\textwidth,
        trim=0.8cm 5.3cm 0.8cm 0cm,
        clip
    ]{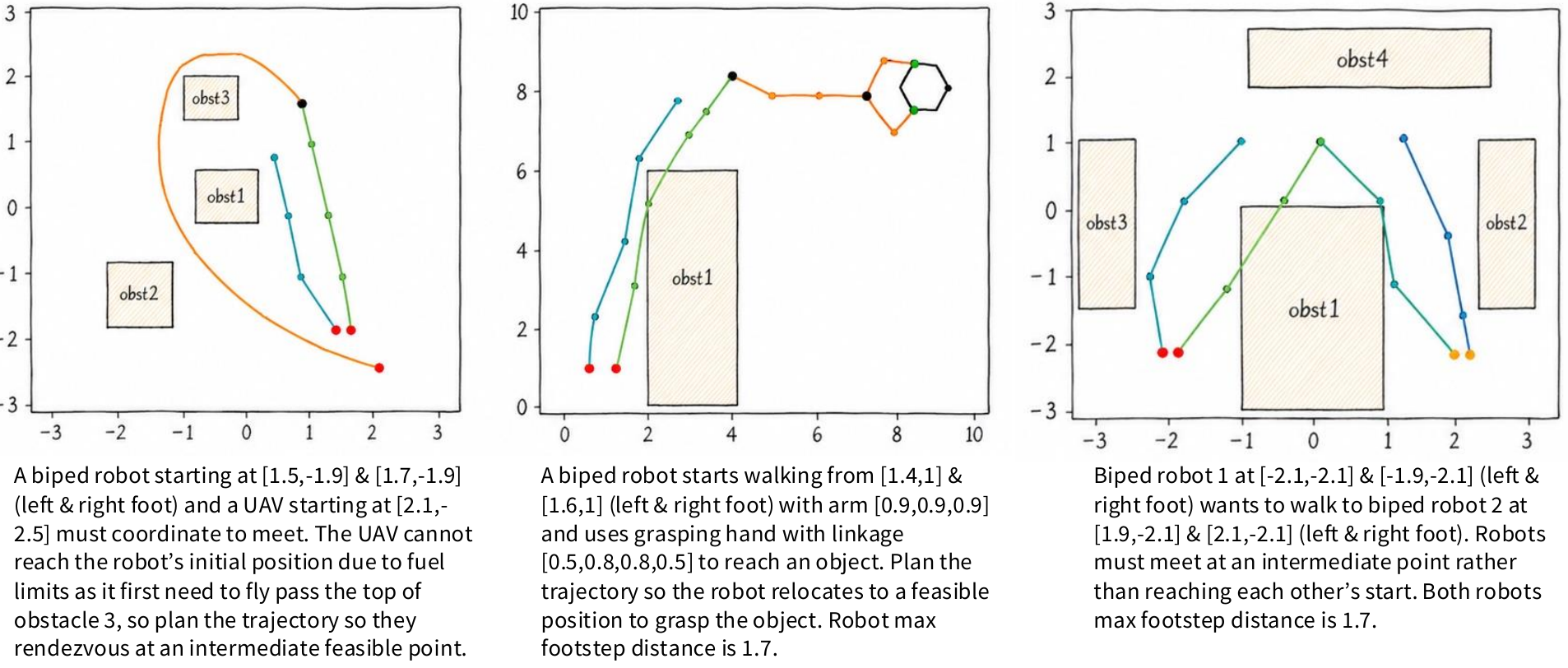}
    \caption{2D qualitative results for the \texttt{Multi-UAV} task. Sub-panels show the IRIS-region-decomposed workspace with two UAV trajectories (orange: UAV1; blue: UAV2), obstacles as gray rectangles, and the shared rendezvous point as a green star. Spatial-style constraints (e.g., \emph{circle} the central obstacle) are visualized via colored region overlays.}
    \label{fig:2d_1}
\end{figure}

\begin{figure}[ht]
    \centering
    \includegraphics[
        width=\textwidth,
        trim=0.8cm 3.8cm 0.8cm 0cm,
        clip
    ]{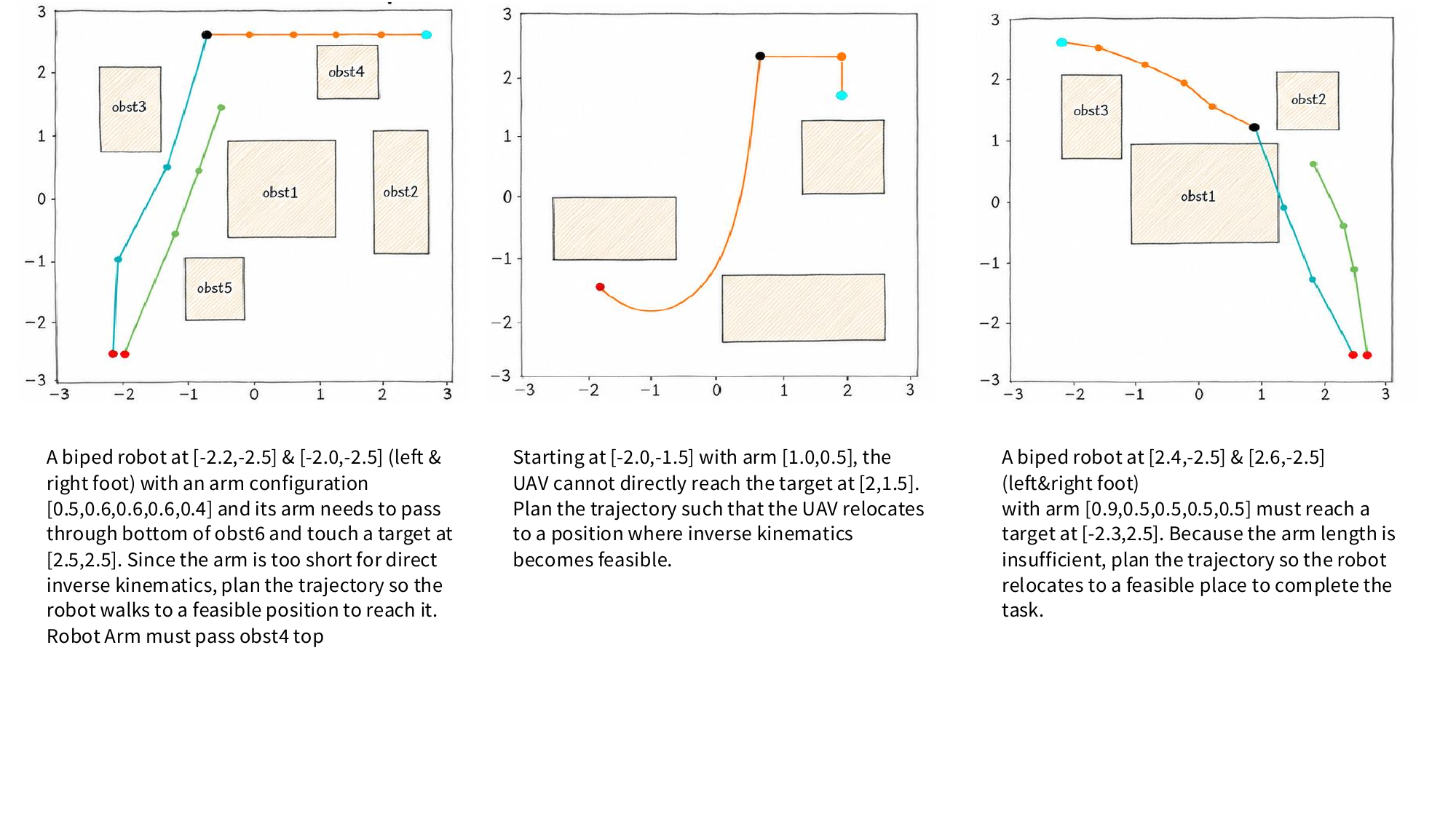}
    \caption{2D qualitative results for the \texttt{UAV+IK} task. Sub-panels show the UAV trajectory (orange) reaching a handoff position (green star), followed by the articulated arm (gray links) executing inverse kinematics to the target (red marker). Obstacles are shown as gray rectangles; IRIS regions overlay the workspace.}
    \label{fig:2d_2}
\end{figure}

\begin{figure}[ht]
    \centering
    \includegraphics[
        width=\textwidth,
        trim=0.8cm 3.5cm 0.8cm 0cm,
        clip
    ]{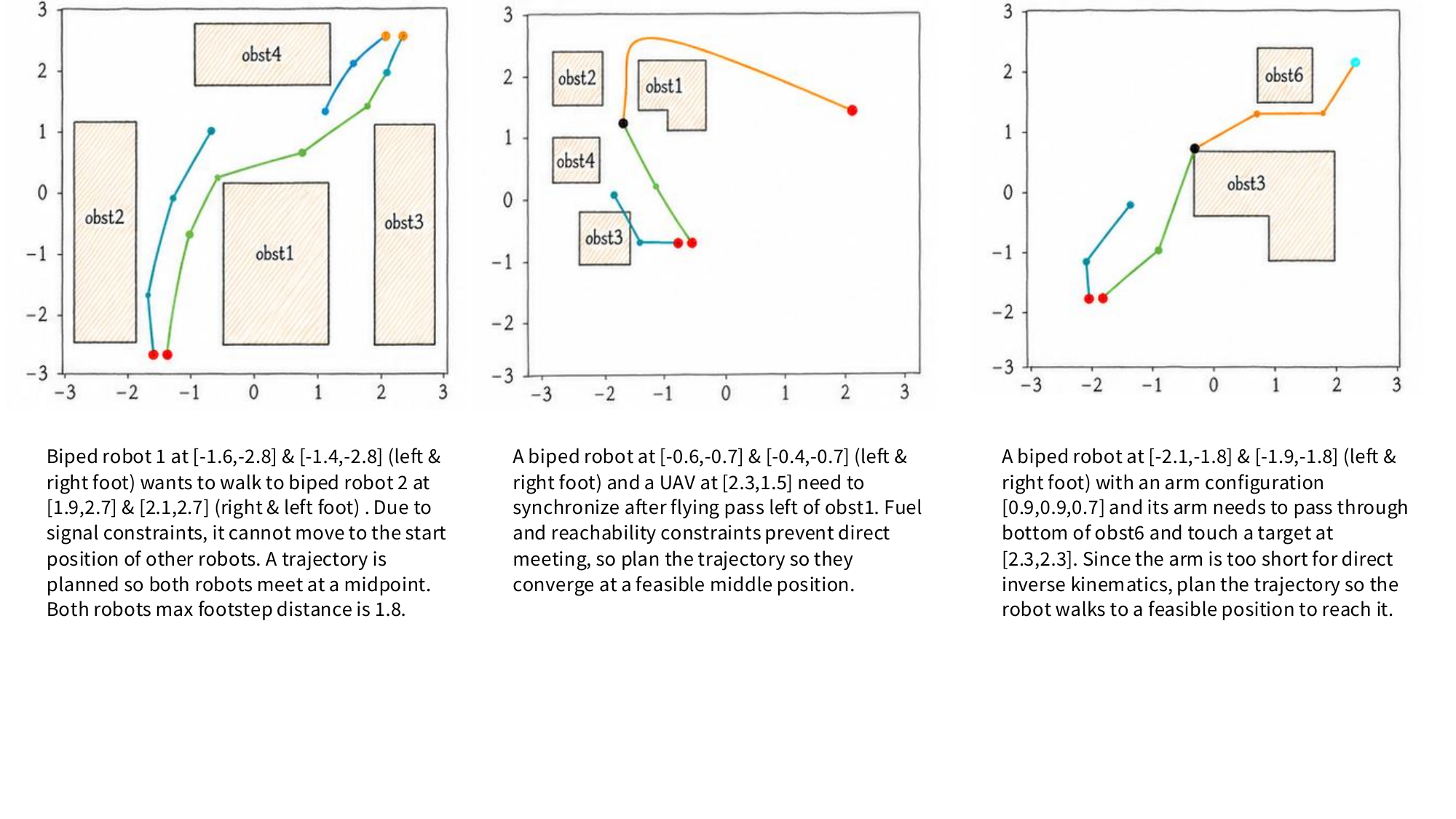}
    \caption{2D qualitative results for the \texttt{Footstep+IK} task. Sub-panels show the bipedal footstep trajectory (alternating left/right contacts as red and blue squares) walking to a stance position, followed by the articulated arm reaching the target (red marker). The shared base location couples the two stages.}
    \label{fig:2d_3}
\end{figure}

\begin{figure}[ht]
    \centering
    \includegraphics[
        width=\textwidth,
        trim=0.8cm 5.3cm 0.8cm 0cm,
        clip
    ]{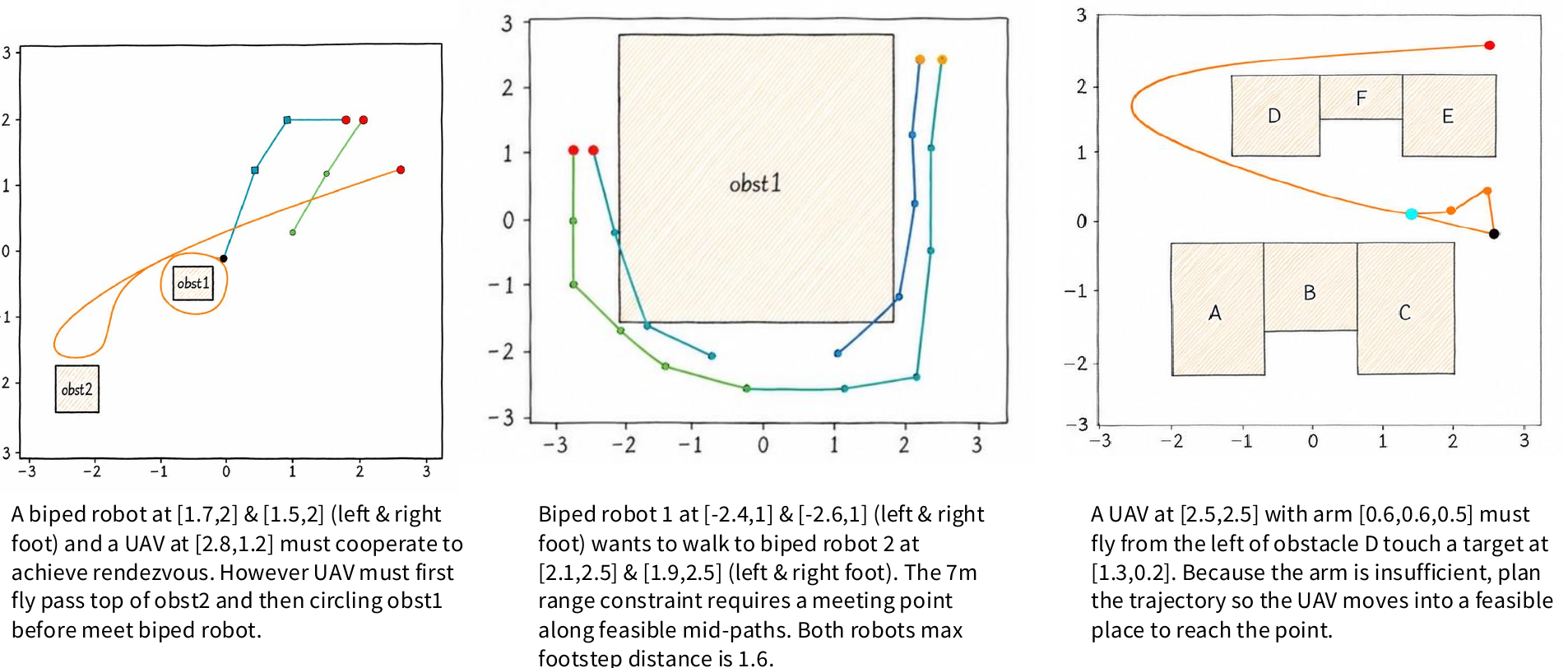}
    \caption{2D qualitative results for the \texttt{Multi-Footstep} task. Sub-panels show two bipedal robots planned jointly (left robot: red/blue contacts; right robot: orange/cyan contacts) meeting at a shared rendezvous point (green star) while avoiding obstacles.}
    \label{fig:2d_4}
\end{figure}

\begin{figure}[ht]
    \centering
    \includegraphics[
        width=\textwidth,
        trim=0.8cm 4.3cm 0.8cm 0cm,
        clip
    ]{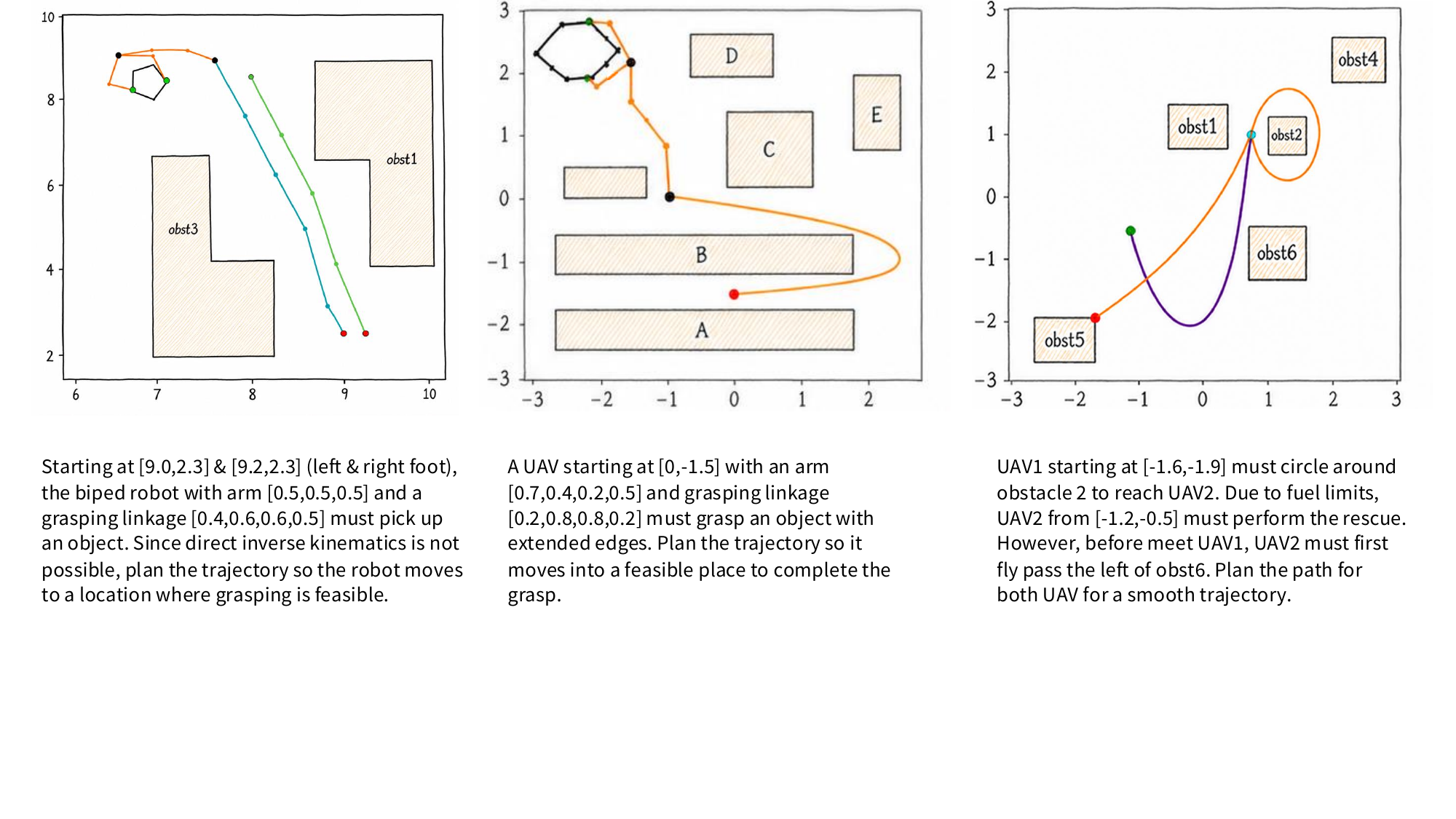}
    \caption{2D qualitative results for the \texttt{Multi-IK} task. Sub-panels show two articulated arms (gray and brown links) reaching a shared target configuration (red marker) while keeping their links collision-free; per-arm pivots are marked with colored dots.}
    \label{fig:2d_5}
\end{figure}

\begin{figure}[ht]
    \centering
    \includegraphics[
        width=\textwidth,
        trim=0.8cm 5.3cm 0.4cm 0cm,
        clip
    ]{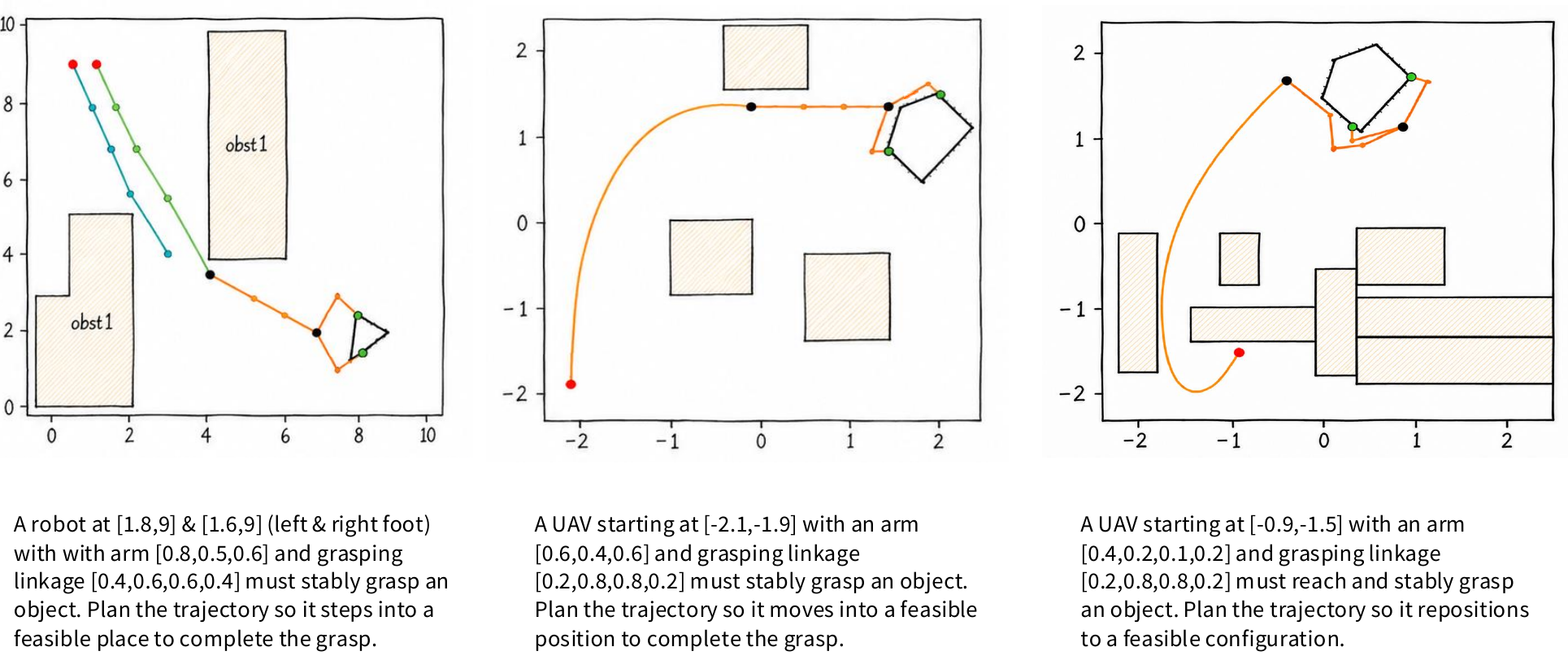}
    \caption{2D qualitative results for the \texttt{Footstep+IK+Grasp} task. Sub-panels show the bipedal footstep trajectory (red/blue contacts) walking to a stance pose, the articulated arm executing IK (gray links), and the selected grasp contact points on the target object (blue dots). All three stages are coupled through shared base/target positions.}
    \label{fig:2d_6}
\end{figure}

\begin{figure}[ht]
    \centering
    \includegraphics[
        width=\textwidth,
        trim=0.4cm 4.0cm 0.8cm 0cm,
        clip
    ]{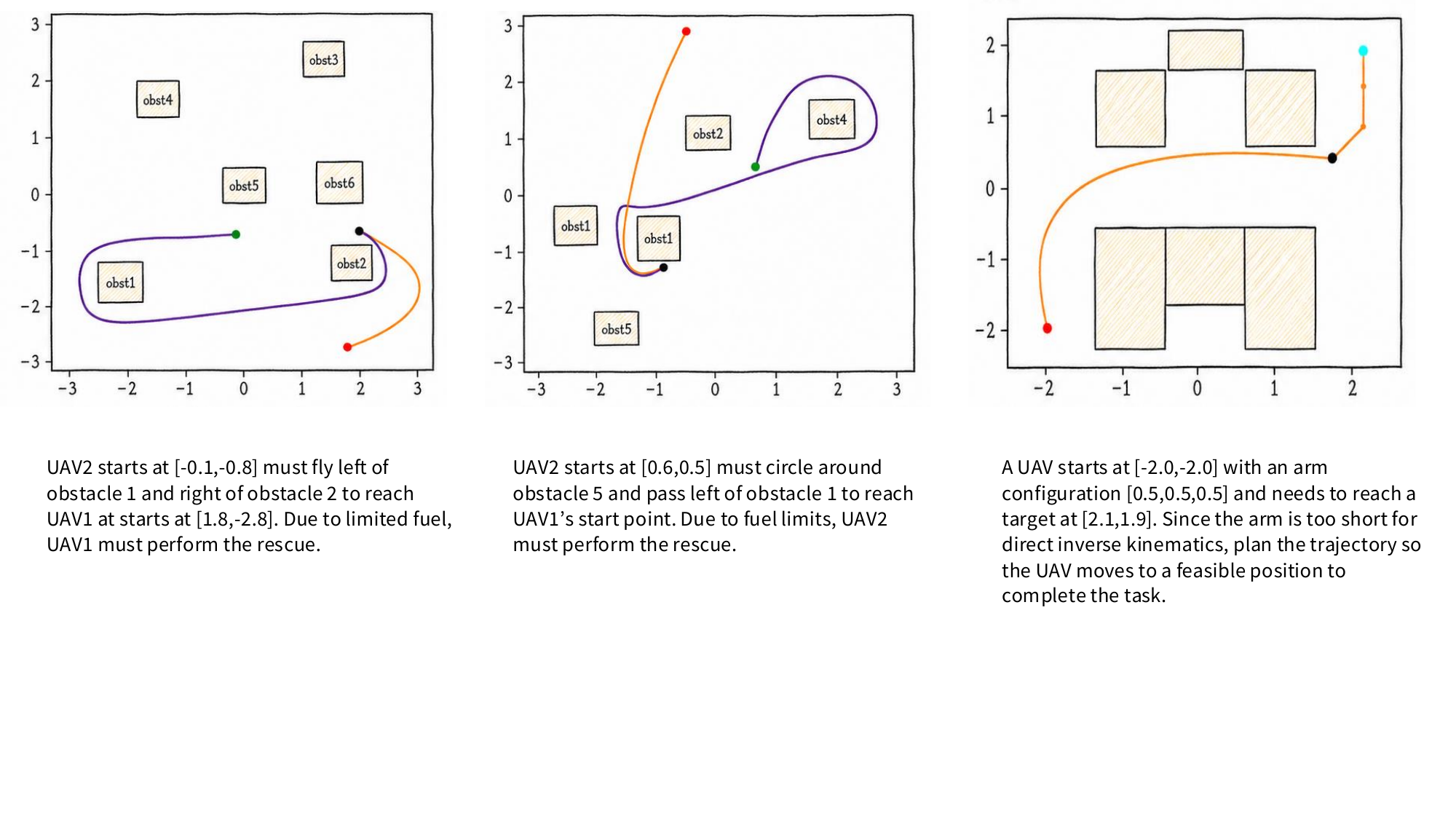}
    \caption{2D qualitative results for the \texttt{UAV+IK+Grasp} task. Sub-panels show the UAV trajectory (orange) reaching a handoff position (green star), the articulated arm performing IK (gray links) attached at the UAV terminal, and the selected grasp contact points on the target object (blue dots). All stages are coupled through shared UAV terminal and IK base/target positions.}
    \label{fig:2d_7}
\end{figure}

\prettyref{fig:2d_1}, \ref{fig:2d_2}, \ref{fig:2d_3}, \ref{fig:2d_4}, \ref{fig:2d_5}, \ref{fig:2d_6}, and \ref{fig:2d_7} show 2D qualitative results for all multi-modal motion planning tasks.
These 2D visualizations provide a clear view of the geometric structure of each solution, including obstacle avoidance, spatial coordination, IK linkages, grasp configurations, and trajectory feasibility.


\subsection{3D Visualization}
Here is our qualitative results in MuJoCo physical simulator. We deploy representative platforms for each modality, including Unitree G1 for footstep planning, Franka FR3 for inverse kinematics, Robotiq 2F-85 for grasping, Skydio X2 for UAV navigation, and Google Robot for UAV+Grasp due to the lack of real UAV-arm platforms. For each figure, the mission instruction prompt is shown below in the white space.

\begin{figure}[ht]
    \centering
    \includegraphics[width=\textwidth]{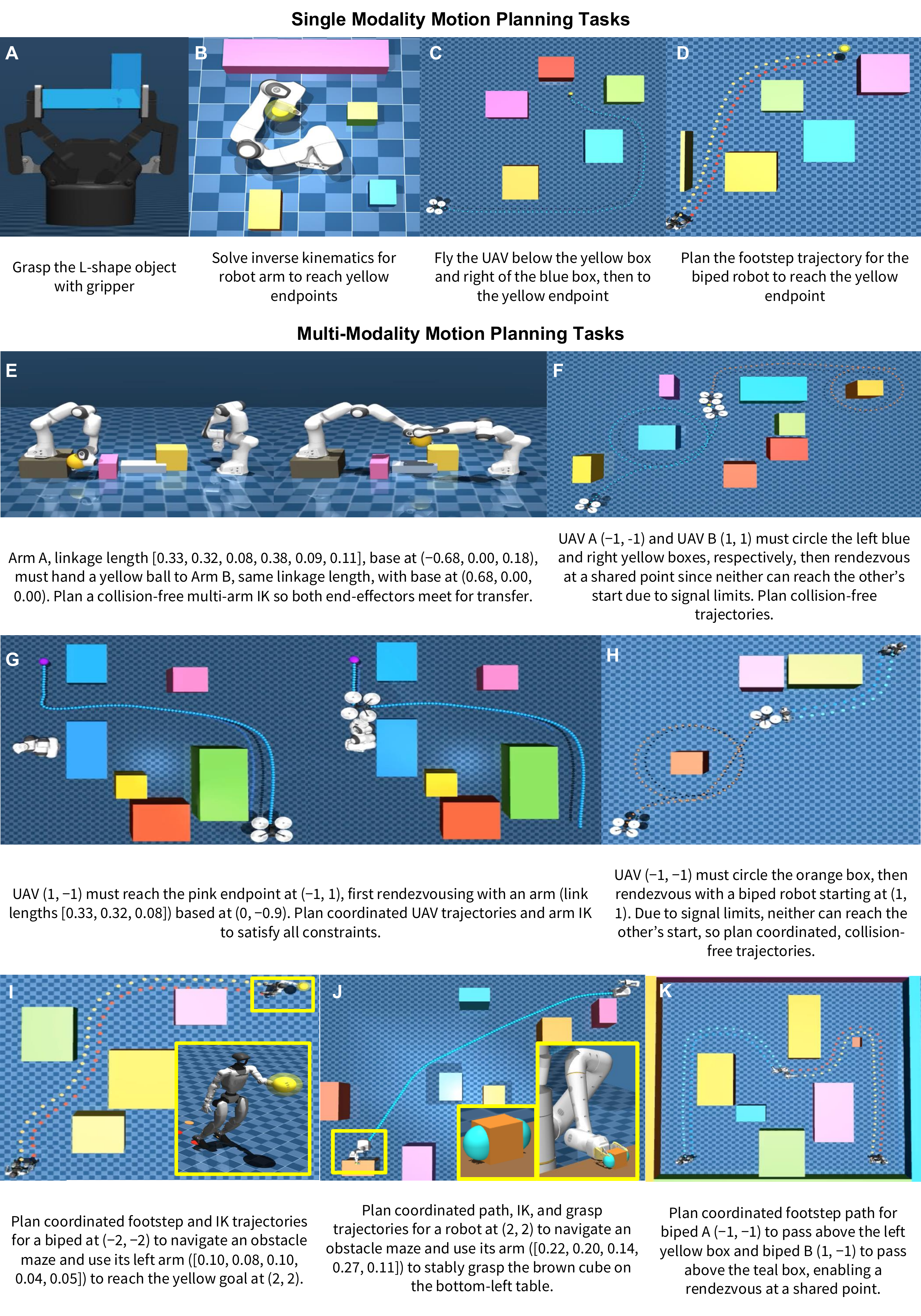}
    \caption{Qualitative results in MuJoCo simulation across single- and multi-modal M$^3$P tasks. (A--D) Single-modal tasks: grasping, inverse kinematics (IK), UAV trajectory planning, and biped footstep planning. (E--K) Multi-modal tasks, including multi-arm coordination, multi-UAV rendezvous, UAV--arm cooperation, UAV--biped coordination, footstep+IK, full pipeline (path+IK+grasp), and multi-biped coordination. Panels B, I, and J plot IK reach points and grasp contact points in yellow and blue dots, respectively. Our method generates feasible, collision-free, and task-consistent trajectories that satisfy geometric, kinematic, and interaction constraints across all settings.}
    \label{fig:result}
\end{figure}

\begin{figure}[ht]
    \centering
    \includegraphics[width=\textwidth]{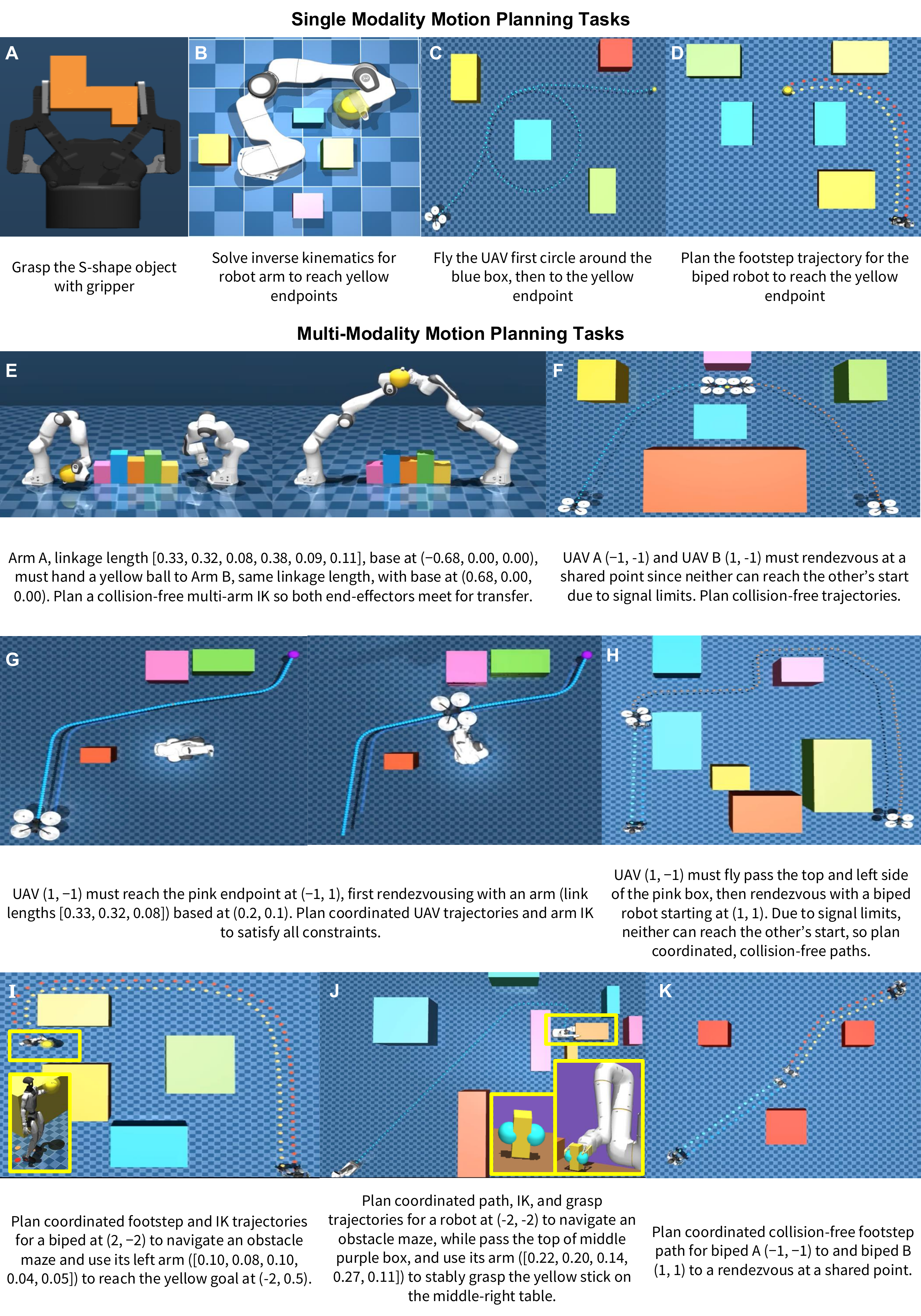}
    \caption{More qualitative results in MuJoCo simulation across single- and multi-modal M$^3$P tasks. (A--D) Single-modal tasks: grasping, inverse kinematics (IK), UAV trajectory planning, and biped footstep planning. (E--K) Multi-modal tasks, including multi-arm coordination, multi-UAV rendezvous, UAV--arm cooperation, UAV--biped coordination, footstep+IK, full pipeline (path+IK+grasp), and multi-biped coordination. Our method generates feasible, collision-free, and task-consistent trajectories that satisfy geometric, kinematic, and interaction constraints across all settings.}
    \label{fig:supp-result}
\end{figure}

\prettyref{fig:result} and \ref{fig:supp-result} show 3D qualitative results in MuJoCo simulation, demonstrating that the generated plans are feasible, collision-free, and consistent with the task constraints in full 3D environments.

\section{Additional Real-World Robot Results}
\label{appen:real_world_results}

We further evaluate M$^3$P-R1 in real-world robotic environments using Franka Panda arms and DJI Tello UAVs.
For each scene, we reconstruct the laboratory setup with the RealityScan app on an iPhone, simplify the scan into object poses, workspace bounds, and obstacle regions, and run M$^3$P planning on a laptop.
For robot-arm tasks, M$^3$P-R1 predicts the final end-effector pose used as the target for inverse kinematics; we then use the Open Motion Planning Library (OMPL)~\citep{6377468} to compute a collision-aware joint-space trajectory from the initial configuration to the selected IK solution.
The resulting robot-arm trajectories and UAV waypoints are executed using \texttt{franky} and \texttt{DJTelloPy}, respectively.
Because our footstep formulation does not model full-body dynamics and lack of access to a physical bipedal platform and a closed-loop tactile gripper for safe contact-rich experiments, the real-world validation focuses on UAV planning, robot-arm inverse kinematics, and their composition. Fourth, real-world validation covers only IK, UAV, Multi-UAV, and UAV+IK tasks. Extending real-world coverage to these modalities is an important direction for future work.

\prettyref{tab:real_world_results} summarizes the real-world results.
For each modality, we construct 10 scenarios, including 5 with object-relative spatial constraints and 5 without spatial constraints.
We report collision-free execution, spatial-constraint alignment, and overall task success.
Most failures are caused by sim-to-real mismatch between the coarse reconstructed mesh and the physical scene, which can lead to missing feasible paths, unexpected real-world collisions, or robot-arm configurations that are feasible in simulation but fall near kinematic singularities during execution.
For Tello UAV experiments, rendezvous is evaluated within a tolerance radius of $0.5\,\mathrm{m}$ rather than requiring physical co-location; this threshold is chosen to match the onboard positioning accuracy of the Tello platform (drift on the order of $0.3$--$0.5\,\mathrm{m}$ during indoor station-keeping) and to maintain safe inter-UAV separation in close-proximity flight.

\begin{table}[t]
    \centering
    \small
    \caption{
    Real-world execution results across 40 hardware trials.
    Each modality contains 10 scenarios, including 5 spatially constrained and 5 non-spatial tasks.
    Spatial alignment is evaluated only on the five spatially constrained scenarios. \textit{Success} is defined as collision-free execution that additionally satisfies any specified spatial constraint and, for composition tasks, achieves the inter-platform rendezvous within the $0.5\,\mathrm{m}$ Tello tolerance.
    }
    \label{tab:real_world_results}
    \setlength{\tabcolsep}{5pt}
    \begin{tabular}{lccccc}
        \toprule
        Modality & \# Scenarios & Spatial / Non-spatial & Collision-free & Spatial alignment & Success \\
        \midrule
        IK        & 10 & 5 / 5 & 8 / 10  & 4 / 5 & 8 / 10 \\
        UAV       & 10 & 5 / 5 & 10 / 10 & 5 / 5 & 10 / 10 \\
        Multi-UAV & 10 & 5 / 5 & 9 / 10  & 5 / 5 & 9 / 10 \\
        UAV+IK    & 10 & 5 / 5 & 8 / 10  & 4 / 5 & 8 / 10 \\
        \bottomrule
    \end{tabular}
\end{table}

\paragraph{Per-failure breakdown.}
We summarize the five failed trials below. (1) \emph{IK--spatial (cube near pitcher).} The reconstructed pitcher mesh under-approximated the handle volume; the executed joint trajectory clipped the handle, triggering the real-arm collision monitor. (2) \emph{IK--non-spatial (beam reach).} The selected IK solution lay near a wrist singularity; OMPL planning succeeded in joint space, but the real arm aborted with a velocity-limit fault during the last segment. (3) \emph{Multi-UAV--non-spatial.} Tello~2 drifted $\sim 0.7\,\mathrm{m}$ from the planned waypoint before rendezvous; this exceeded the $0.5\,\mathrm{m}$ tolerance and triggered an emergency hover. (4) \emph{UAV+IK--spatial.} The UAV reached the handoff position, but the meshed table edge was offset by $\sim 4\,\mathrm{cm}$, causing the IK end-effector pose to graze the table on descent. (5) \emph{UAV+IK--non-spatial.} The Tello lost its visual lock briefly during the descent toward the handoff point and ended $\sim 0.6\,\mathrm{m}$ off the planned location, exceeding the tolerance. Three of the five failures are therefore directly attributable to coarse mesh reconstruction (sim-to-real mismatch), and two to onboard sensing limitations of the Tello platform.

\begin{figure*}[t]
    \centering
    \includegraphics[width=\textwidth, trim={0 1.85in 0 0}, clip]{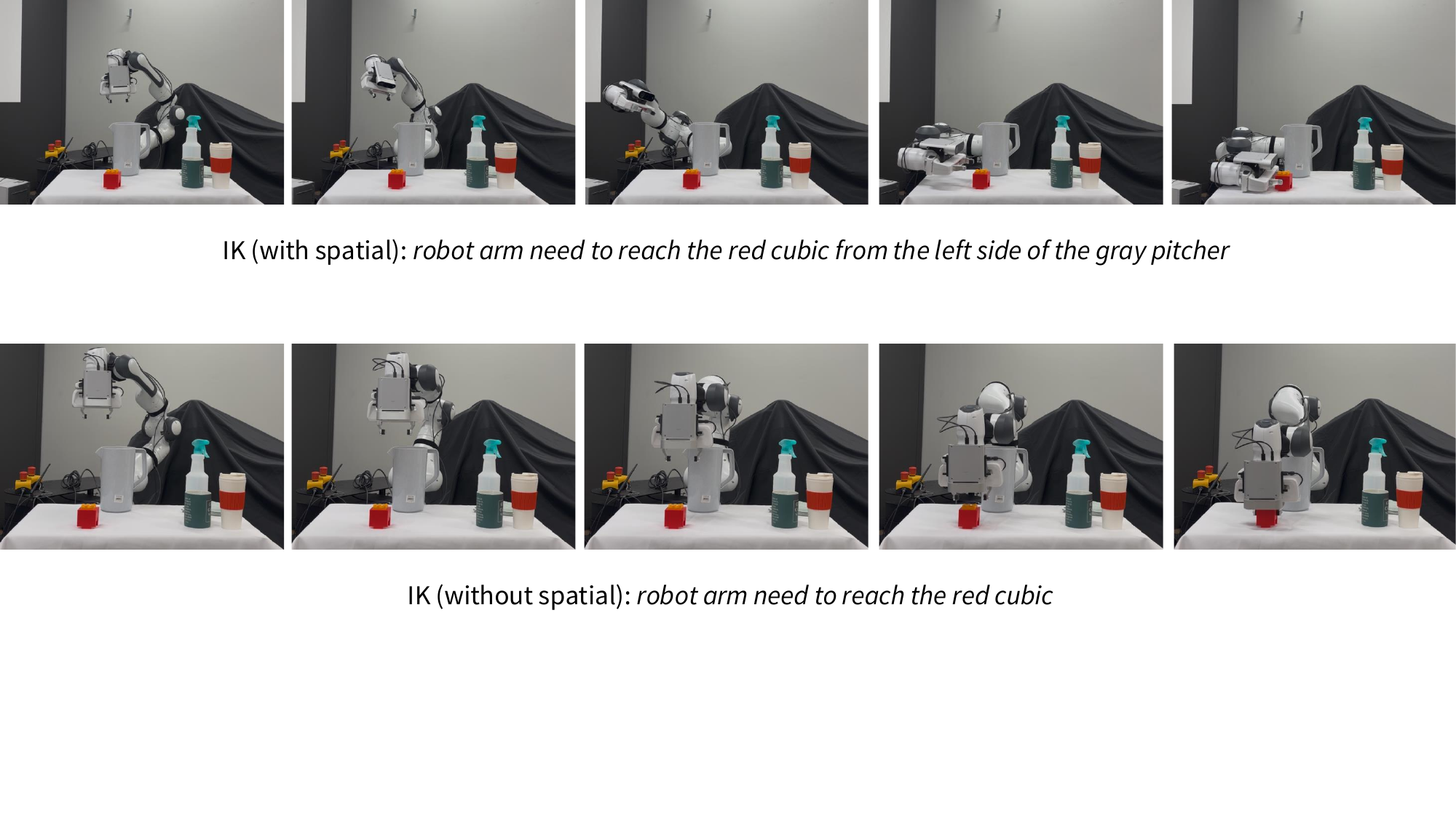}
    \caption{
    Real-world Franka Panda reaching task involving a red cube and a gray pitcher.
    Top: reaching the red cube with a spatial constraint relative to the gray pitcher.
    Bottom: reaching the red cube without the spatial constraint.
    }
    \label{fig:real_ik_pitcher}
\end{figure*}

\begin{figure*}[t]
    \centering
    \includegraphics[width=\textwidth, trim={0 1.85in 0 0}, clip]{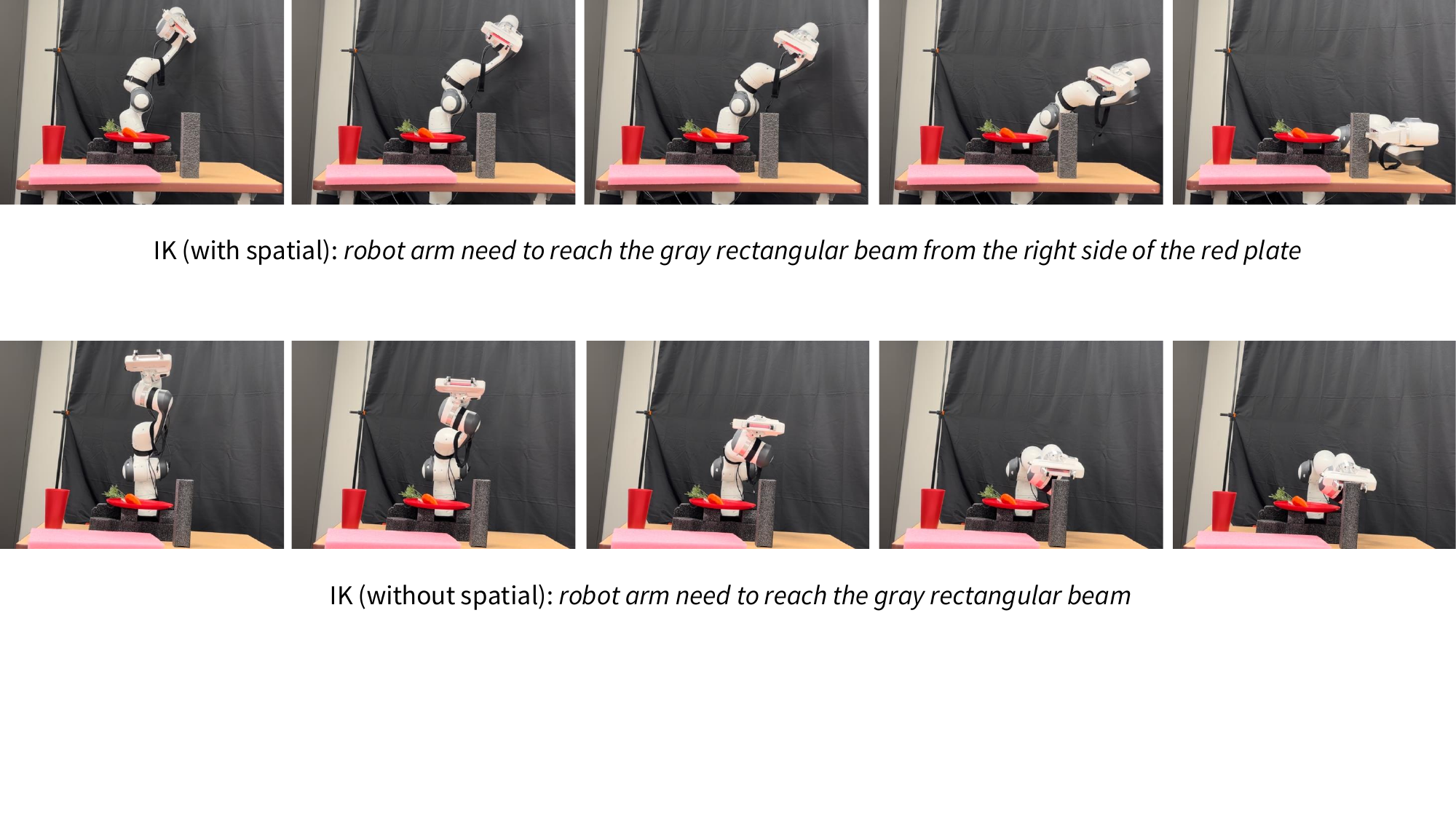}
    \caption{
    Real-world Franka Panda reaching task involving a gray rectangular beam and a red plate.
    Top: reaching the beam with an object-relative spatial constraint.
    Bottom: reaching the beam without the spatial constraint.
    }
    \label{fig:real_ik_beam}
\end{figure*}

\begin{figure*}[t]
    \centering
    \includegraphics[width=\textwidth, trim={0 1.85in 0 0}, clip]{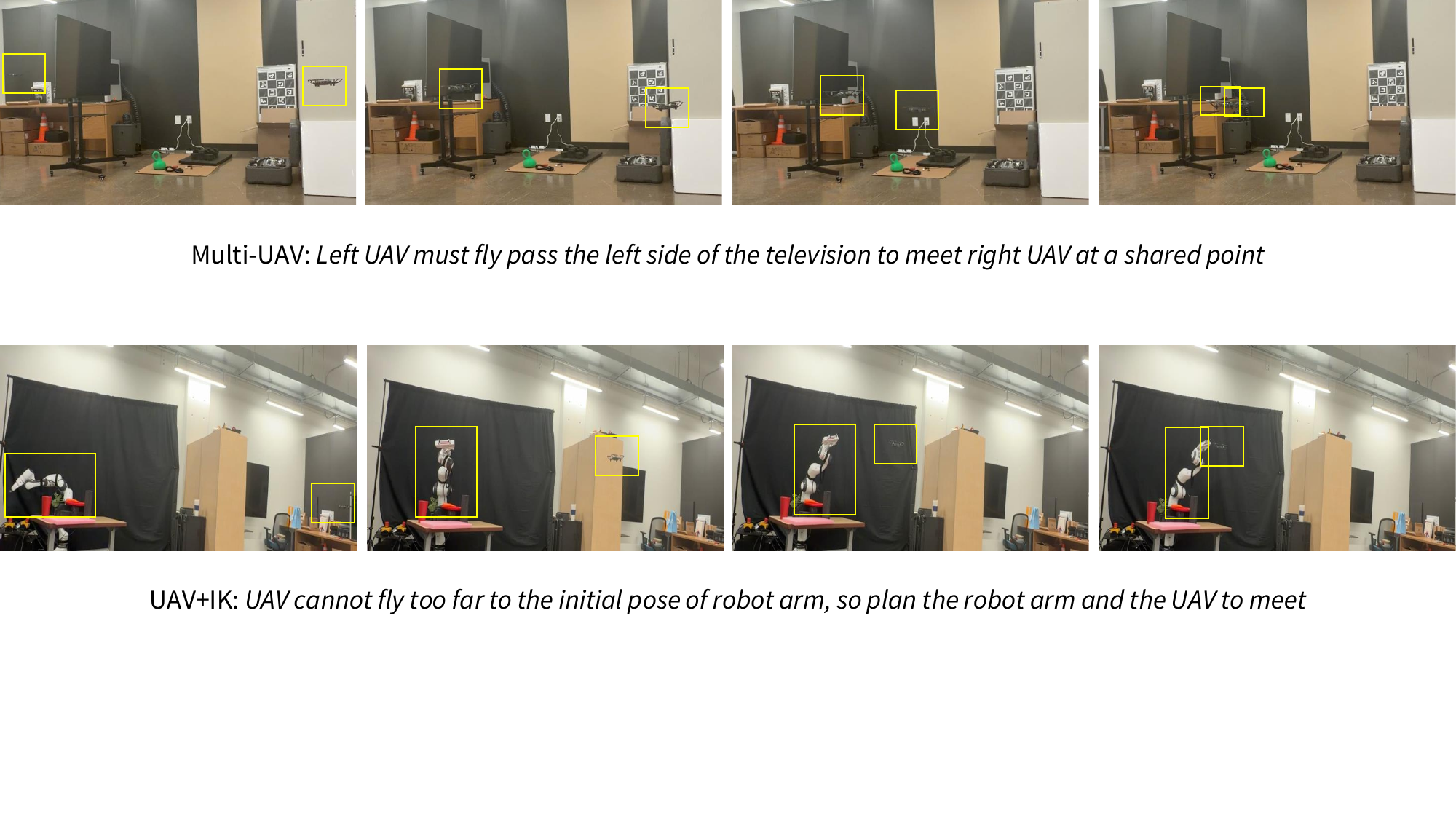}
    \caption{
    Real-world composition results.
    Top: multi-UAV planning with an object-relative spatial constraint and a shared meeting point.
    Bottom: UAV+IK planning, where the UAV and robot arm are jointly planned to meet at a feasible shared point.
    }
    \label{fig:real_composition}
\end{figure*}

\prettyref{fig:real_ik_pitcher} and \ref{fig:real_ik_beam} show Franka Panda reaching tasks with and without object-relative spatial constraints.
The constrained cases require the robot to approach the target from a specified side of a nearby object, while the unconstrained cases only require reaching the target.
\prettyref{fig:real_composition} shows multi-UAV and UAV+IK composition tasks, demonstrating that M$^3$P-R1 can coordinate shared meeting points and spatial constraints across heterogeneous platforms.

\clearpage

\section{LLM-Callable API}\label{appen:api}
In this section, we list the set of LLM-callable APIs for discretizing the general non-convex constraints into disjointly convex sets.

\subsection{Initialization APIs}

Initialization APIs construct safe regions, optimization variables, and object representations. These functions define the problem structure before constraints are added.

\begin{apibox}{generate\_iris\_regions(bounding\_lb, bounding\_ub, endpoints, num\_regions, obstacle\_map)}
\textbf{Inputs.} Workspace bounds, task endpoints (start/goal or base/target), number of regions, and obstacle map.

\textbf{Description.}
Generates convex collision-free regions using IRIS. These regions serve as the feasible domain for all geometric constraints and enable efficient mixed-integer formulations.
\end{apibox}

\begin{apibox}{augment\_iris\_regions(iris\_regions, obstacle\_map, styles, side\_width)}
\textbf{Inputs.} Base region set, obstacle metadata, spatial relations, and side width.

\textbf{Description.}
Augments regions with auxiliary side boxes to encode spatial instructions such as directional passing or obstacle circling.
\end{apibox}

\begin{apibox}{initialize\_trajectory\_problem(num\_segments, num\_regions)}
\textbf{Inputs.} Number of trajectory segments and regions.

\textbf{Description.}
Creates B\'ezier control points and binary region assignments, forming the core variables for trajectory optimization.
\end{apibox}

\begin{apibox}{initialize\_kinematic\_problem(link\_lengths, n\_poly, num\_intermediate, base)}
\textbf{Inputs.} Link lengths, rotation discretization level, collision sampling resolution, and base position.

\textbf{Description.}
Initializes articulated chain variables, including pivots, rotations, and collision samples, for kinematic reasoning.
\end{apibox}

\begin{apibox}{initialize\_footstep\_problem(num\_regions, steps)}
\textbf{Inputs.} Number of regions and planning horizon.

\textbf{Description.}
Defines footstep positions, activation variables, and region assignments for biped locomotion.
\end{apibox}

\begin{apibox}{initialize\_finger\_selection\_problem(n\_fingers, num\_points)}
\textbf{Inputs.} Number of fingers and candidate contact points.

\textbf{Description.}
Initializes binary selection variables and robustness objective for contact selection without kinematics.
\end{apibox}

\begin{apibox}{initialize\_grasp\_problem(link\_lengths, n\_poly, num\_intermediate, base)}
\textbf{Inputs.} Gripper link lengths, rotation discretization, collision resolution, and base.

\textbf{Description.}
Initializes the kinematic chain that connects the robot base to grasp contact points.
\end{apibox}

\begin{apibox}{build\_finger\_select\_object(vertices\_list)}
\textbf{Inputs.} Object vertex list.

\textbf{Description.}
Constructs geometric representation of the object used for contact sampling in finger selection.
\end{apibox}

\begin{apibox}{sample\_finger\_contact\_candidates(obj, delta, delta\_n)}
\textbf{Inputs.} Object geometry and spatial/normal sampling resolution.

\textbf{Description.}
Samples candidate contact points and outward normals on the object surface.
\end{apibox}

\begin{apibox}{build\_grasp\_object(vertices\_list)}
\textbf{Inputs.} Object vertices.

\textbf{Description.}
Constructs object geometry for grasp planning, used for contact selection and collision reasoning.
\end{apibox}

\subsection{Constraint APIs}

Constraint APIs translate geometric, kinematic, and contact requirements into solver-verifiable constraints.

\begin{apibox}{add\_segment\_assignment\_constraints(H)}
\textbf{Inputs.} Binary region assignment variables $H$.

\textbf{Description.}
Ensures each segment selects exactly one region, enforcing discrete structure in trajectory planning.
\end{apibox}

\begin{apibox}{add\_region\_membership\_constraints(C, H, iris\_regions, M)}
\textbf{Inputs.} Points $C$, assignment variables $H$, regions, and Big-$M$ (a large constant used for mixed-integer feasibility relaxation; see \prettyref{appen:math_detail}).

\textbf{Description.}
Constrains points to lie within assigned regions, linking discrete assignments to continuous geometry.
\end{apibox}

\begin{apibox}{add\_collision\_free\_constraints(points, iris\_regions)}
\textbf{Inputs.} Geometry samples and region set.

\textbf{Description.}
Ensures all robot geometry remains in obstacle-free space through region assignment.
\end{apibox}

\begin{apibox}{add\_spatial\_relation\_constraints(H, styles)}
\textbf{Inputs.} Assignment variables and spatial relations.

\textbf{Description.}
Encodes high-level spatial instructions such as directional passing or obstacle circling.
\end{apibox}

\begin{apibox}{add\_trajectory\_continuity\_constraints(C)}
\textbf{Inputs.} Control points $C$.

\textbf{Description.}
Enforces smooth transitions between trajectory segments, ensuring physically realistic motion.
\end{apibox}

\begin{apibox}{add\_trajectory\_endpoint\_constraints(C, start, goal)}
\textbf{Inputs.} Control points and endpoints.

\textbf{Description.}
Anchors trajectory start and end positions to fixed or shared targets.
\end{apibox}

\begin{apibox}{add\_reachability\_constraints(positions)}
\textbf{Inputs.} Sequence of positions.

\textbf{Description.}
Ensures transitions between consecutive states satisfy motion feasibility constraints.
\end{apibox}

\begin{apibox}{add\_kinematic\_constraints(pivots, rotations, link\_lengths)}
\textbf{Inputs.} Joint pivots $p$, rotations $R$, and link lengths $\ell$.

\textbf{Description.}
Enforces forward kinematics, coupling joint configuration with spatial geometry.
\end{apibox}

\begin{apibox}{add\_base\_target\_constraints(base\_pivot, end\_pivot, start, goal)}
\textbf{Inputs.} Base pivot, end-effector pivot, base position, and target position.

\textbf{Description.}
Fixes the endpoints of a kinematic chain.
\end{apibox}

\begin{apibox}{add\_contact\_selection\_constraints(selection)}
\textbf{Inputs.} Binary selection variables $F$.

\textbf{Description.}
Selects valid and distinct contact points from candidates.
\end{apibox}

\begin{apibox}{add\_end\_effector\_constraints(selection, points)}
\textbf{Inputs.} Selection variables and candidate contact points for grasp tasks.

\textbf{Description.}
Maps selected contacts to end-effector positions.
\end{apibox}

\begin{apibox}{add\_wrench\_constraints(points, normals, friction\_coef)}
\textbf{Inputs.} Contact points, normals, and friction coefficient $\mu$.

\textbf{Description.}
Enforces force-closure conditions in wrench space for stable grasping.
\end{apibox}

\begin{apibox}{add\_objective(objective\_terms)}
\textbf{Inputs.} Task-dependent objective terms.

\textbf{Description.}
Defines the optimization objective, such as trajectory smoothness or grasp robustness.
\end{apibox}

\section{MIP Formulation}
\label{appen:math_detail}
In this section, we provide the detailed formulation of each of our single-modal tasks.

Throughout the paper we use MIP as the umbrella term; the specific formulations below are mixed-integer quadratic programs (MIQPs)~\citep{deits2015efficient}, and one variant is a mixed-integer semidefinite program (MISDP)~\citep{dai2017synthesis}.

Throughout this section, the modality subsections use \emph{local notation}: symbols such as $N$, $R$, $p_i$, $n_i$, $a_i$, $J$, $b$, $r$, $\rho$, $c_i$, $\ell$, $\sigma_i$, $\Lambda$, $n$ (denoting reachability halfspace normals $n_j$ and support-plane normals $n_r$), and the Cartesian coordinates $y$, $z$ are locally redefined in each subsection (UAV, IK, Grasp, Finger Selection, Footstep), and across-modality reuse should be read as an abuse of notation with the meaning fixed by the subsection in which the symbol appears. By contrast, $s$ and $g$ always denote start and goal states inherited from the mission specification.

\subsection{UAV}
\emph{All symbols introduced below are scoped to this subsection unless stated otherwise.}
We formalize the UAV trajectory planning problem as an MIQP~\citep{deits2015efficient}.  Let $N$ denote the number of trajectory segments, $R$ the number of IRIS regions, and $C\in\mathbb{R}^{4N\times 3}$ the cubic B\'ezier control points across all segments. We index $C_{j,k}$ for the $k$-th control point ($k=0,1,2,3$) of the $j$-th segment, and $C_{j,k,d}$ for its $d$-th coordinate ($d\in\{x,y,z\}$). Binary variables $H_{r,j}\!\in\!\{0,1\}$ assign segment $j$ to region $r$. We aggregate the region-assignment, control-point feasibility (Big-$M$ relaxed with region $r$ described by $A_r x \le b_r$ and $M>0$ a sufficiently large constant), continuity, and start/goal constraints into a single labeled block:
\begin{equation}
\begin{aligned}
\sum_{r=1}^R H_{r,j} &= 1, && \forall j=1,\dots,N, \\
A_r C_{j,k} &\le b_r + M(1-H_{r,j}), && \forall r,j,k, \\
C_{j,3} &= C_{j+1,0}, && \forall j=1,\dots,N-1, \\
C_{j,3}-C_{j,2} &= C_{j+1,1}-C_{j+1,0}, && \forall j=1,\dots,N-1, \\
C_{j,3}-2C_{j,2}+C_{j,1} &= C_{j+1,0}-2C_{j+1,1}+C_{j+1,2}, \\
&\quad \forall j=1,\dots,N-1, \\
C_{0,0} &= s, \qquad C_{N-1,3}=g.
\end{aligned}
\label{eq:uav-constraints}
\end{equation}
The continuity rows are added by API call \texttt{add\_continuity\_constraints} and the start/goal row by \texttt{add\_start\_goal\_constraints}.

\textbf{Objective:} We minimize the trajectory's integrated jerk, computed for each segment $j$ and dimension $d\in\{x,y,z\}$ as:
\[
J_{j,d}=6(C_{j,3,d}-3C_{j,2,d}+3C_{j,1,d}-C_{j,0,d}).
\]
The overall objective is then:
\begin{equation}
\min_{C,H} \; \sum_{j=1}^N \sum_{d\in\{x,y,z\}} J_{j,d}^2 \quad \text{s.t.\ \prettyref{eq:uav-constraints}}.
\label{eq:objective}
\end{equation}
The binary assignments $H$ are integer feasibility variables and enter the program only through the region-assignment and Big-$M$ rows of \prettyref{eq:uav-constraints}; they do not appear in the objective body.

\prettyref{eq:uav-constraints} together with \ref{eq:objective} define the UAV planning MIQP. This formulation tightly couples discrete region assignments with continuous B\'ezier control points, ensuring trajectories are collision-free, dynamically smooth, and solver-executable.

\subsection{Extension to Our Novel Spatial Constraints}\label{appen:novel_spatial}
We propose a novel type of constraint to enforce additional constraints on UAV, enabling more expressive trajectory descriptions. Specifically, we allow UAV to fly from the ``left/right/top/bottom/front/back'' of an obstacle, for a given number of times. We further allow UAV to circle around the obstacle, also for a given number of times. These additional requirements can be achieved by so-called crossing constraints. Suppose there are two IRIS regions indexed by $p$ and $q$, where each region is defined in $\mathbb{R}^3$ as $A_r x \le b_r$, and we want to check whether UAV crosses the boundary of $p$ and $q$ at the $a$th trajectory segment. Such a check can be achieved by introducing the continuous indicator variable $X_{pq}^a \in [0,1]$ and the following constraints:
\[\begin{aligned}
X_{pq}^a &\le H_{p,a}, \\
X_{pq}^a &\le H_{q,a+1}, \\
X_{pq}^a &\ge H_{p,a}+H_{q,a+1}-1.
\end{aligned}\]
We can further ensure that only one crossing happens by the constraint $\sum_{a=1}^{N-1}(X_{pq}^a+X_{qp}^a)=1$.
Now, suppose we want the UAV to circle around an obstacle exactly once, then we can generate IRIS regions surrounding the obstacle, as indexed by $r_{\text{bot}}, r_{\text{rgt}}, r_{\text{top}}, r_{\text{lft}}$ (short for bottom, right, top, and left); we illustrate the construction in the horizontal plane, and the same idea generalizes to additional regions for fully 3D circling. We then enforce:
\[
\sum_{a=1}^{N-1}X_{r_{\text{bot}}\, r_{\text{rgt}}}^a=
\sum_{a=1}^{N-1}X_{r_{\text{rgt}}\, r_{\text{top}}}^a=
\sum_{a=1}^{N-1}X_{r_{\text{top}}\, r_{\text{lft}}}^a=
\sum_{a=1}^{N-1}X_{r_{\text{lft}}\, r_{\text{bot}}}^a=1.
\]
Further suppose we want the UAV to fly from the left of an obstacle. Then we can generate a rectangular IRIS region on the left, as indexed by $r_{\text{lft}}$. We then create two other IRIS regions, as indexed $r_{\text{lft,top}}$ and $r_{\text{lft,bot}}$. $r_{\text{lft,top}}$ intersects with $r_{\text{lft}}$ from the top and $r_{\text{lft,bot}}$ intersects with $r_{\text{lft}}$ from the bottom. With these three IRIS regions, we only need to enforce that the UAV fly $r_{\text{lft,top}}$ to $r_{\text{lft}}$ then to $r_{\text{lft,bot}}$:
\[
\sum_{a=1}^{N-1}X_{r_{\text{lft,top}}\, r_{\text{lft}}}^a=
\sum_{a=1}^{N-1}X_{r_{\text{lft}}\, r_{\text{lft,bot}}}^a=1.
\]

\subsection{IK}
\emph{All symbols introduced below are scoped to this subsection unless stated otherwise.}
We present the simplified, 3D variant of~\citet{dai2019global}, modeling a spatial manipulator with $L$ links of fixed lengths $\{\ell_1,\dots,\ell_L\}$. Let $p_i \in \mathbb{R}^3$ denote the pivot of joint $i$, with base $p_0$ and end-effector $p_L$. Rotations are approximated via polyhedral discretization with $N_\text{poly}$ facets. The MIP aggregates the following constraints. \textbf{Base anchoring} fixes the base at a given location $s$. \textbf{Rotation discretization:} each link's local-to-global transform uses a rotation matrix $\mathbf{R}_i \in \mathbb{R}^{3\times 3}$ constrained to lie in a polyhedral approximation of $SO(3)$; each $\mathbf{R}_i$ is constructed via piecewise linear relaxation of $SO(3)$ by calling~\texttt{add\_chain\_rotation\_constraints}, a construction also used in footstep planning~\citep{deits2014footstep}, yielding an outer approximation of $SO(3)$ with bounded approximation error that decreases with $N_\text{poly}$. The \textbf{end-effector constraint} forces the final position to coincide with the goal $g$. \textbf{Collision discretization} keeps each link collision-free (formulated similarly to the UAV case): each link $[p_{i-1},p_i]$ is discretized into $K$ intermediate points $q_{i,k}$, and all $\{p_i,q_{i,k}\}$ are concatenated into a collision-point matrix indexed by $j=1,\ldots,L(K+1)+1$, with the $j$-th column denoted $Q_j$. We again construct $N_R$ IRIS regions (in $\mathbb{R}^3$) with binary assignment variables $H_{r,j}\in\{0,1\}$ and Big-$M$ feasibility (safety distance $d_\text{safe}$), where each IRIS region $r$ is described by $\{x\in\mathbb{R}^3 : A_r x \le b_r\}$ (same convention as the UAV subsection) and $M>0$ is a sufficiently large constant:
\begin{equation}
\begin{aligned}
p_0 &= s, \\
p_i &= p_{i-1} + \mathbf{R}_i \begin{bmatrix}\ell_i \\ 0 \\ 0\end{bmatrix}, && i=1,\dots,L, \\
p_L &= g, \\
q_{i,k} &= \big(1-\tfrac{k}{K+1}\big)p_{i-1} + \tfrac{k}{K+1}p_i, && k=1,\dots,K, \\
\sum_{r=1}^{N_R} H_{r,j} &= 1, && \forall j=1,\ldots,L(K+1)+1, \\
A_r Q_j &\le (b_r - d_\text{safe}) + M (1 - H_{r,j}), && \forall r,j=1,\ldots,L(K+1)+1.
\end{aligned}
\label{eq:ik-constraints}
\end{equation}
The constraints in \prettyref{eq:ik-constraints} define our MIQP. The formulation couples discrete rotation approximations, continuous kinematics, and region-based collision avoidance.

\subsection{\label{appen:math_grasp}Grasp}
\emph{All symbols introduced below are scoped to this subsection unless stated otherwise.}
We present a simplified, 3D variant of~\cite{liu2020new} for grasp planning. The grasp formulation extends the IK problem introduced in the previous section. All IK constraints (chain kinematics, collision discretization, IRIS region assignment, and optional spatial relation constraints) are inherited directly. We now introduce additional variables and constraints that model grasp contact selection and force-closure feasibility.

\paragraph{Grasp selection (two fingers):}
We sample $N$ equidistant points on the surface of objects as candidate grasp points $\{c_i\}$, which is achieved by calling~\texttt{get\_object\_surface\_samples}. We further introduce binary matrices $F\in\{0,1\}^{2\times N}$ and binary selection flags $\chi\in\{0,1\}^N$, where $\chi_i=1$ iff candidate $i$ is selected, that select two distinct contact points from the $N$ surface samples, via the following constraints:
\[\begin{aligned}
\sum_{i=1}^N F_{1i} &= 1, && \sum_{i=1}^N F_{2i} = 1, \\
F_{1i}+F_{2i} &\le 1, && \chi_i = F_{1i}+F_{2i},\quad i=1,\dots,N.
\end{aligned}\]
The selected end-effector (contact) positions are:
\[
\text{ee}_1=\sum_{i=1}^N F_{1i}\,c_i,\quad \text{ee}_2=\sum_{i=1}^N F_{2i}\,c_i.
\]
The above two variables replace the fixed start and goal constraints from the IK problem by anchoring
the kinematic chain to two contact points on the object.

\paragraph{Force-closure (wrench) constraints:}
For each candidate contact $c_i\in\mathbb{R}^3$ with outward unit normal $n_i$ and two orthonormal tangent directions $\mathbf{t}_i^{(1)}, \mathbf{t}_i^{(2)}$, we define a polyhedral approximation of the friction cone with friction coefficient $\mu$ using four generators:
\[\begin{aligned}
f_i^{(1)} &= n_i + \mu\,\mathbf{t}_i^{(1)}, \\
f_i^{(2)} &= n_i - \mu\,\mathbf{t}_i^{(1)}, \\
f_i^{(3)} &= n_i + \mu\,\mathbf{t}_i^{(2)}, \\
f_i^{(4)} &= n_i - \mu\,\mathbf{t}_i^{(2)}.
\end{aligned}\]
with corresponding spatial wrenches:
\[
w_i^{(k)} = \begin{bmatrix} f_i^{(k)} \\ c_i \times f_i^{(k)} \end{bmatrix} \in \mathbb{R}^6,\quad k=1,\dots,4.
\]
Let $\{\hat w_d\}_{d=1}^D\subset\mathbb{R}^6$ be a fixed set of $D$ test wrench directions, and let $\rho\ge 0$ denote the common inscribed sphere radius in wrench space (the grasp robustness margin to be maximized below). Introducing nonnegatives $\alpha_{i,k,d}$ and slacks $\gamma_d$ for each test wrench direction, we impose:
\[
\sum_{i=1}^N \sum_{k=1}^4 w_i^{(k)} \alpha_{i,k,d} = \gamma_d\,\hat w_d, \quad \forall d=1,\dots,D.
\]
\[\begin{aligned}
\sum_{k=1}^4 \alpha_{i,k,d} &\le 1, && \forall i,d, \\
\sum_{k=1}^4 \alpha_{i,k,d} &\le \chi_i, && \forall i,d, \\
\alpha_{i,k,d} &\ge 0, && \forall i,k,d, \\
\gamma_d &\ge \rho, && \forall d=1,\dots,D.
\end{aligned}\]
A 4-generator pyramid yields a coarse but sufficient polyhedral approximation of the friction cone, capturing its convex geometry while maintaining a minimal number of linear constraints for efficient MIQP optimization.

\paragraph{Objective:} Our objective is to maximize the inscribed sphere radius:
\[
\max_{\rho, F, \chi, \alpha, \gamma}\; \rho \quad \text{s.t.\ (force-closure and selection constraints above)}.
\]
Here $F$, $\chi$, $\alpha$, $\gamma$ are auxiliary variables constrained by the equations above. Note that this inscribed sphere radius can be approximated using a discretization technique~\citep{liu2020new}, leading to an MIQP, or the lower-bound relaxation~\citep{dai2017synthesis}, leading to an MISDP. We use the MIQP formulation since Gurobi does not solve MISDP. Put together, the grasp MIP augments the IK formulation with the above constraints, which jointly enforce valid finger selection and force-closure robustness.

\subsection{Finger Selection}
\emph{All symbols introduced below are scoped to this subsection unless stated otherwise.}
Finger selection can be viewed as a simplified version of the grasp problem (\prettyref{appen:math_grasp}), where we do not consider the reachability of fingers, and directly select $k$ contact points from $N$ sampled candidate grasp points to maximize the inscribed sphere radius, where $k\le N$ is a user-specified contact count.

\subsection{Footstep}
\emph{All symbols introduced below are scoped to this subsection unless stated otherwise.}
Footstep planning~\citep{deits2014footstep} builds upon the general spatial and region constraints introduced above, but specializes them for biped locomotion.

Let $N$ denote the maximum number of footsteps. Each step $i=0,\dots,N-1$ has a contact position $p_i=(x_i,y_i,z_i)\in\mathbb{R}^3$. Although the decision variable is written in $\mathbb{R}^3$, feasible contacts lie on locally planar support surfaces, yielding a 2.5D terrain representation while retaining height variation. Let $R$ denote the number of convex safe regions generated by IRIS for the footstep problem. Let $a_i\in\{0,1\}$ indicate whether step $i$ is active, $H_{r,i}\in\{0,1\}$ indicate whether step $i$ is assigned to convex safe region $r=1,\dots,R$, and $\delta_i^{\mathrm{term}}\in\{0,1\}$ indicate whether step $i$ is the terminal footstep. We use $M>0$ as a sufficiently large big-$M$ constant.

\paragraph{Region membership.}
Let $\{p\in\mathbb{R}^3:A_rp\le b_r\}$ denote the collision-free convex region generated by IRIS. Each active footstep must lie in exactly one convex safe region:
\[
\begin{aligned}
\sum_{r=1}^R H_{r,i} &= a_i,
&& \forall i, \\
A_r p_i &\le b_r + M(1-H_{r,i})\mathbf{1},
&& \forall r,i.
\end{aligned}
\]
The trajectory between consecutive footsteps is not constrained to remain inside these regions, because the robot can lift its swing foot over obstacles.

\paragraph{Support-surface constraints.}
For 3D or 2.5D terrain, region membership alone does not guarantee that a footstep lies on a valid contact surface. For each support region $r$, let $\Pi_r=\{p\in\mathbb{R}^3:n_r^\top p=h_r\}$ denote the local support plane, where $n_r\in\mathbb{R}^3$ is the plane normal and $h_r\in\mathbb{R}$ is the plane offset. We activate the support-plane constraint using the same region-assignment variable:
\[
-\epsilon_z - M(1-H_{r,i})
\le n_r^\top p_i-h_r
\le \epsilon_z + M(1-H_{r,i}),
\quad \forall r,i,
\]
where $\epsilon_z\ge 0$ is a small contact-height tolerance. For flat-ground experiments, this reduces to $z_i=0$ for all active footsteps.

\paragraph{Start and terminal constraints.}
The first two contacts are fixed to the initial left and right foot positions, $p_0=p_{\mathrm{L}}^{\mathrm{start}}$ and $p_1=p_{\mathrm{R}}^{\mathrm{start}}$, and are always active, $a_0=a_1=1$.
Exactly one active step is selected as the terminal footstep:
\[
\sum_{i=0}^{N-1}\delta_i^{\mathrm{term}}=1,
\qquad
\delta_i^{\mathrm{term}}\le a_i,
\quad \forall i.
\]
We introduce an auxiliary terminal position $p^{\mathrm{term}}\in\mathbb{R}^3$ and enforce:
\[
-M(1-\delta_i^{\mathrm{term}})\mathbf{1}
\le p^{\mathrm{term}}-p_i
\le M(1-\delta_i^{\mathrm{term}})\mathbf{1},
\quad \forall i.
\]
Thus, if $\delta_i^{\mathrm{term}}=1$, then $p^{\mathrm{term}}=p_i$.

\paragraph{Step activation monotonicity.}
Once a step is inactive, no later step can be active: $a_i\ge a_{i+1}$ for all $i=0,\dots,N-2$. This constraint can be added by calling \texttt{add\_monotonicity\_constraints}.

\paragraph{Left-right ordering and lateral separation.}
Footsteps alternate between left and right feet. Let $\sigma_i\in\{-1,+1\}$ denote the side of step $i$, where $\sigma_i=+1$ represents the left foot and $\sigma_i=-1$ represents the right foot. Let $e_{\mathrm{lat}}\in\mathbb{R}^3$ be the lateral unit direction of the walking frame, pointing from the right side to the left side, and let $\Delta_{\min}>0$ be the minimum lateral separation between consecutive feet. For each active transition, we impose:
\[
\sigma_{i+1}\, e_{\mathrm{lat}}^\top (p_{i+1}-p_i)
\ge \Delta_{\min} - M(1-a_{i+1}),
\quad \forall i=0,\dots,N-2.
\]
This enforces each active footstep to lie on the correct lateral side of the previous stance foot. In a planar setting where $x$ is the lateral coordinate, this reduces to the corresponding signed constraint on $x_{i+1}-x_i$.

\paragraph{Reachability constraints.}
From stance foot $p_i$, the next contact $p_{i+1}$ must lie in a shifted reachable polytope. Let the local reachable set be approximated by $N_J$ halfspaces with normals $n_j\in\mathbb{R}^3$ and offsets $b_j^{\mathrm{reach}}\in\mathbb{R}$:
\[
\mathcal{R}_{\text{reach}}=\{q\in\mathbb{R}^3:n_j^\top q\le b_j^{\mathrm{reach}},\; j=1,\dots,N_J\}.
\]
Let $\Delta_i^{\mathrm{nom}}\in\mathbb{R}^3$ denote the nominal offset from stance foot $i$ to the next footstep, which may depend on the left/right side. The reachability constraint is:
\[
n_j^\top\big(p_{i+1}-p_i-\Delta_i^{\mathrm{nom}}\big)
\le b_j^{\mathrm{reach}} + M(1-a_{i+1}),
\quad \forall i=0,\dots,N-2,\; j=1,\dots,N_J.
\]

\paragraph{Goal-error constraints.}
Let $p_g=(x_g,y_g,z_g)\in\mathbb{R}^3$ denote the desired terminal footstep location. We introduce nonnegative slack variables $\varepsilon_k\ge 0$ for $k\in\{x,y,z\}$ to measure terminal goal error:
\[
\varepsilon_k \ge p^{\mathrm{term}}_k-p_{g,k},
\qquad
\varepsilon_k \ge -(p^{\mathrm{term}}_k-p_{g,k}),
\quad k\in\{x,y,z\}.
\]
At optimum, $\sum_{k\in\{x,y,z\}}\varepsilon_k=\|p^{\mathrm{term}}-p_g\|_1$.

\paragraph{Nominal-lane deviation constraints.}
We define a nominal walking lane to regularize footstep placement. Let $\bar p_i\in\mathbb{R}^3$ be the nominal centerline point for step $i$, $\hat n_i^{\mathrm{lat}}\in\mathbb{R}^3$ be the local lateral unit direction, and $w_s>0$ be the nominal half-stance width. The nominal footstep location is $p_i^{\mathrm{nom}} = \bar p_i + \sigma_i\, w_s\, \hat n_i^{\mathrm{lat}}$.
We define $d_i$ as the activated $\ell_1$ deviation of step $i$ from this nominal location. For each coordinate $k\in\{x,y,z\}$, introduce a nonnegative slack variable $d_{i,k}\ge 0$ and impose:
\[
\begin{aligned}
d_{i,k}
&\ge p_{i,k}-p_{i,k}^{\mathrm{nom}} - M(1-a_i),
&& \forall i,k, \\
d_{i,k}
&\ge -(p_{i,k}-p_{i,k}^{\mathrm{nom}}) - M(1-a_i),
&& \forall i,k, \\
0&\le d_{i,k}\le M a_i,
&& \forall i,k.
\end{aligned}
\]
Then $d_i=\sum_{k\in\{x,y,z\}}d_{i,k}$. Thus, $d_i=0$ for inactive footsteps, while for active footsteps it equals the $\ell_1$ deviation from the preferred left/right nominal lane location at optimum.

\paragraph{Step-length constraints.}
To linearize the $\ell_1$ distance between consecutive active footsteps, introduce nonnegative slack variables $\Lambda_{i,k}\ge 0$ for each transition $i=0,\dots,N-2$ and coordinate $k\in\{x,y,z\}$:
\[
\begin{aligned}
\Lambda_{i,k}
&\ge p_{i+1,k}-p_{i,k} - M(1-a_{i+1}),
&& \forall i,k, \\
\Lambda_{i,k}
&\ge -(p_{i+1,k}-p_{i,k}) - M(1-a_{i+1}),
&& \forall i,k, \\
0&\le \Lambda_{i,k}\le M a_{i+1},
&& \forall i,k.
\end{aligned}
\]
We define $\Lambda_i=\sum_{k\in\{x,y,z\}}\Lambda_{i,k}$. Thus, $\Lambda_i=0$ when the transition to step $i+1$ is inactive, and $\Lambda_i=\|p_{i+1}-p_i\|_1$ at optimum when the transition is active.

\paragraph{Objective.}
The footstep planner minimizes a weighted sum of step length, number of active footsteps, terminal goal error, and nominal-lane deviation:
\[
\begin{aligned}
\min_{\substack{\{p_i,\, a_i,\, H_{r,i},\, \delta_i^{\mathrm{term}},\\ p^{\mathrm{term}},\, \varepsilon_k,\, d_{i,k},\, \Lambda_{i,k}\}}}\;
& w_1\sum_{i=0}^{N-2} \Lambda_i
+ w_2\sum_{i=0}^{N-1} a_i \\
&{}+ w_3\sum_{k\in\{x,y,z\}} \varepsilon_k
+ w_4\sum_{i=0}^{N-1} d_i \\
&\text{s.t.\ above constraints},
\end{aligned}
\]
where $w_1,w_2,w_3,w_4\ge 0$ are user-defined weights. The first term encourages short steps, the second encourages using fewer footsteps, the third penalizes terminal goal error, and the fourth keeps active footsteps close to their preferred left/right nominal lane locations. The binary variables $a_i$, $H_{r,i}$, and $\delta_i^{\mathrm{term}}$ enter through the feasibility constraints, while the continuous variables $p_i$ and $p^{\mathrm{term}}$ determine the step locations and terminal contact.

\subsection{Multi-modal Tasks}
\emph{All symbols introduced below are scoped to this subsection unless stated otherwise.}
In multi-modal tasks, subsystems such as UAV, IK, and Finger Selection must coordinate at specific interface points—for example, the UAV endpoint aligning with the IK base, or the IK end-effector aligning with the Grasp target. These \emph{shared meeting points} enforce cross-domain consistency while preserving feasibility within each individual modality.

Rather than relying on predefined coupling mechanisms, we train the LLM via GRPO-based reinforcement learning to autonomously discover and construct such coordination strategies. In particular, the model learns to introduce shared variables and synthesize cross-domain constraints that couple multiple modalities within a unified optimization program, yielding jointly consistent and solver-feasible multi-modal plans.

\section{Method Implementation Detail}\label{appen:implementation}

Here we describe the fine-tuning details of our methods and implementation of baselines.

\subsection{Fine-tuning Detail}\label{appen:finetune_detail}

We fine-tune the model in two stages. Stage~1 uses SFT to teach the model the required output format, discretization API usage, and executable MIP-program structure. We use Low-Rank Adaptation (LoRA)~\citep{hu2022lora}, which freezes the base model weights and updates only the adapter parameters. Stage~2 further optimizes the SFT model with GRPO using solver-grounded verifiable rewards. During both stages, the base model weights are frozen and only the LoRA adapter parameters are updated, which makes the training more memory- and compute-efficient. The LoRA adapter uses rank $r=16$, \texttt{lora\_alpha}$=32$, and is applied to \texttt{q\_proj}, \texttt{k\_proj}, \texttt{v\_proj}, \texttt{o\_proj}, \texttt{gate\_proj}, \texttt{up\_proj}, and \texttt{down\_proj} modules.

Because our rewards are solver-grounded and directly verifiable, policy updates are guided by deterministic execution and feasibility signals rather than noisy human-preference feedback. We therefore follow recent RLVR practice~\citep{yu2026dapo, liu2025understanding} and disable KL regularization by setting the KL coefficient $\beta=0.0$. This design avoids over-constraining the policy toward the SFT initialization and allows broader exploration during solver-grounded optimization. Similar choices have been adopted in recent reasoning-oriented RL systems, such as Decoupled Clip and Dynamic sAmpling Policy Optimization (DAPO)~\citep{yu2026dapo} and Understanding R1-Zero-Like Training~\citep{liu2025understanding}, which report a zero KL coefficient in their training configuration.

The complete two-stage fine-tuning process was conducted on one NVIDIA H100 80GB GPU for approximately 30 hours. At inference time, the M$^3$P-R1 policy is served on a single NVIDIA L4 GPU with vLLM~\citep{kwon2023efficient}; the average end-to-end LLM call latency is approximately $4.5\,\mathrm{s}$ per task at $T_{\text{samp}}=0.6$ sampling (range: $2.1$--$11.4\,\mathrm{s}$ depending on output length). The detailed SFT and GRPO configurations are summarized in \prettyref{tab:sft_details} and \ref{tab:grpo_details}.

\begin{table}[t]
\centering
\small
\renewcommand{\arraystretch}{1.15}
\caption{\small Stage~1 supervised fine-tuning details.}
\label{tab:sft_details}
\begin{tabular}{p{0.36\linewidth} p{0.56\linewidth}}
\toprule
\textbf{Parameter} & \textbf{Value} \\
\midrule
Maximum training epoch & $1$ \\
Learning rate & $2\times10^{-4}$ \\
Per-device batch size & $2$ \\
Gradient accumulation steps & $4$ \\
Effective batch size & $8$ per GPU process \\
Maximum sequence length & $8192$ \\
Optimizer & \texttt{adamw} \\
Learning-rate scheduler & Linear \\
Warmup ratio & $0.1$ \\
Weight decay & $0.01$ \\
\bottomrule
\end{tabular}
\end{table}

\begin{table}[t]
\centering
\small
\renewcommand{\arraystretch}{1.15}
\caption{\small Stage~2 GRPO fine-tuning details.}
\label{tab:grpo_details}
\begin{tabular}{p{0.36\linewidth} p{0.56\linewidth}}
\toprule
\textbf{Parameter} & \textbf{Value} \\
\midrule
Maximum training steps & $800$ \\
Per-device batch size & $4$ \\
Number of generations per prompt & $4$ \\
Maximum sequence length & $8192$ \\
Learning rate & $2\times10^{-6}$ \\
KL coefficient & $\beta=0.0$ \\
Gradient accumulation steps & $1$ \\
Optimizer & \texttt{adamw} \\
Learning-rate scheduler & Linear \\
Warmup ratio & $0.1$ \\
Weight decay & $0.001$ \\
Sampling temperature & $0.6$ \\
Sampling \texttt{min\_p} & $0.1$ \\
Sampling \texttt{top\_p} & $0.95$ \\
Sampling \texttt{top\_k} & $50$ \\
Precision & $bf16$ \\
Stop strings & \texttt{</call>} and end-of-sequence (EOS) token \\
Reward weights &
$\lambda_1=1$, $\lambda_2=2$, $\lambda_3=1$, $\lambda_4=1$ \\
\bottomrule
\end{tabular}
\end{table}

\subsection{Baselines Setup Detail}\label{appen:baselines_setup}

\paragraph{LLM-baseline evaluation protocol.}
For each prompt-only LLM baseline reported in \prettyref{tab:llm_avg_performance} and \ref{tab:combined_comparison}, as well as for the LLM component of LLM+P, LLM+RRT*, LLM+BIT*, and LLM+TrajOpt, we use \textbf{GPT-5.2} as the underlying LLM. Sampling uses temperature $T_{\text{samp}}=0.6$, a maximum output length of 8192 tokens (matched to the training context window in \prettyref{tab:grpo_details}), and one in-context example for format alignment. On parse failure (e.g., missing or malformed \texttt{<think>}/\texttt{<code>} block), we retry the call once before recording the trial as a failure. The GPT-5.2 API endpoint version is pinned to the evaluation date so that all runs in this paper use the same underlying model snapshot. ReAct, in contrast, uses Qwen3-8B as the backbone with the same callable API handbook provided in context (see the ReAct paragraph below). For all settings, the same API handbook, mission instruction, environment JSON, and obstacle map are supplied in the prompt; the only quantity that varies across runs of the same baseline is the sampling seed.

\paragraph{PDDLStream.}
We adopt \textsc{PDDLStream} as a symbolic--geometric planning baseline for M$^3$P. This baseline assumes a fully grounded symbolic specification and does not perform language understanding. In our setup, the environment is parsed from the structured JSON block embedded in \texttt{mission\_instruction.txt}, where task semantics (e.g., obstacle-side constraints) are already encoded via \texttt{obstacle\_map[...]["style"]}. These styles are deterministically compiled into discrete tasks and geometric variants, while continuous feasibility (collision checking and distance evaluation) is handled via external streams over a visibility roadmap. Planning is performed using an incremental solver that jointly selects task variants and motion transitions under a distance-based cost objective. Key parameters include roadmap collision padding ($10^{-7}$), node margin ($10^{-3}$), fixed style-region width (e.g., $0.2$), deterministic milestone sampling, and a planning timeout of $60$s.
\paragraph{LLM+P.}
To enable language grounding, we introduce an LLM-based preprocessing layer that translates free-form mission instructions into a structured symbolic representation. The model takes as input the natural-language instruction and environment JSON, and outputs a constrained JSON intermediate representation that specifies task-relevant obstacles, normalized styles, ordering, and (for multi-agent tasks) coordination variables such as shared meeting points. This IR is strictly validated and deterministically compiled into a PDDLStream-ready problem by assigning styles to \texttt{obstacle\_map}, enforcing task ordering, and filtering geometric variants. Planning is then performed using the same PDDLStream backend with an augmented domain that enforces the LLM-specified structure. We use deterministic decoding (temperature $0.1$) and enforce a fixed schema with restricted style and variant vocabularies. While this baseline enables language-to-symbolic grounding, intermediate variables are generated heuristically and are not jointly optimized with geometric constraints, which can lead to infeasible or suboptimal solutions.

\paragraph{Expert-MIP.}
To determine whether M$^3$P-R1's performance can be explained solely by access to high-quality expert optimization modules, we construct an Expert-MIP baseline. Given the natural-language task and environment description, an LLM first identifies the required planning modalities, selects the corresponding fixed expert MIP modules, and predicts the interface variables needed to connect them. These interfaces include quantities such as shared meeting points, contact locations, target poses, and intermediate states. Each expert module then solves its corresponding subproblem using a hand-designed formulation; the LLM does not generate or modify the internal optimization constraints. We evaluate this baseline using both Qwen3-8B and GPT-5.2. The expert modules achieve 100.0\% success on single-modal tasks, but multi-modal success decreases to 55.6\% and 63.2\%, respectively, because an interface predicted independently of downstream geometric feasibility can render subsequent IK, grasp, or motion-planning stages infeasible. This baseline therefore isolates the benefit of learning coupled optimization formulations beyond selecting and composing fixed expert modules.

\paragraph{LLM+RRT* / LLM+BIT*.}
We implement sampling-based motion planning baselines by combining LLM-predicted waypoint decomposition with classical planners in $\mathbb{R}^3$. The LLM takes as input the mission instruction and environment description, and predicts a sequence of key waypoints and their order to satisfy style constraints (e.g., passing above/below obstacles) and coordination requirements (e.g., meeting points). Each consecutive waypoint pair defines a planning segment, for which a collision-free path is computed using OMPL planners~\cite{6377468}, specifically RRT* or BIT*. The state space is a bounded 3D Euclidean space, and collision checking rejects states inside axis-aligned obstacles inflated by a fixed clearance margin. Planning is performed independently per segment, and resulting paths are concatenated to form the final trajectory. Planner-specific hyperparameters follow OMPL defaults. This baseline evaluates LLM-based task decomposition with sampling-based feasibility, but the decoupled waypoint generation and segment-wise planning can lead to suboptimal or infeasible trajectories under complex multi-modal constraints.
\paragraph{LLM+TrajOpt.}
We include a trajectory optimization baseline using \textsc{TrajOpt}~\citep{schulman2014motion}. The LLM first predicts waypoint sequences from the instruction and environment. Initial trajectories are generated by connecting waypoints with BIT*, and then refined using TrajOpt, which performs sequential convex optimization over a discretized trajectory to minimize smoothness and collision costs. Obstacles are modeled as inflated axis-aligned boxes (clearance $\approx 0.2$), and optimized segments are concatenated and densified (e.g., $\Delta s=0.03$). This improves local smoothness but remains dependent on LLM-generated waypoints and does not achieve global joint optimization.
\paragraph{ReAct.}
We choose Qwen3-8B as the backbone model and also provide the same callable API handbook we used for RFT to Qwen3-8B. ReAct is an agentic workflow that iteratively performs reasoning and tool invocation to solve the task. At each step, the model generates an action (e.g., calling planning or geometry APIs) based on intermediate observations, forming a reasoning--acting loop until a solution is produced or a fixed time budget is reached (180s). However, without fine-tuning, LLMs lack the skills for MIP discretizations and long-horizon reasoning, which often leads to failure under complex multi-modal constraints.

\prettyref{tab:baseline_compare} summarizes the capability comparison.
\paragraph{Scope.}
The sampling-based baselines (LLM+RRT*, LLM+BIT*, and LLM+TrajOpt) are designed for geometric path planning and are not directly applicable to tasks involving discrete contact sequences or kinematic feasibility (e.g., footstep planning, grasp synthesis, or inverse kinematics). Therefore, we evaluate these baselines only on single-agent and multi-agent UAV navigation tasks.
\paragraph{Comparison to ours.}
In contrast, our method directly generates solver-grounded MIP programs and jointly optimizes continuous trajectories and discrete mode transitions within a unified optimization framework, ensuring global feasibility and consistency.

\begin{table}[t]
\centering
\small
\caption{\small Capability comparison. Lang.: language grounding; Feas.: continuous feasibility; Joint: joint discrete--continuous optimization; Multi: multi-modal coordination; Opt.: solution optimality.}
\label{tab:baseline_compare}
\begin{tabular}{lccccc}
\toprule
Method & Lang. & Feas. & Joint & Multi & Opt. \\
\midrule
PDDLStream & No  & Yes & Partial & Limited & Approx \\
LLM+P      & Yes & Yes & No      & Weak    & Subopt \\
Ours (MIP) & Yes & Yes & Yes     & Yes     & Near-opt \\
\bottomrule
\end{tabular}
\end{table}

\section{Additional Related Work}
\label{app:additional_related_work}

\paragraph{Data-driven methods for MIP.}
Beyond hand-designed MIP formulations, recent work has explored learning-based techniques to accelerate mixed-integer optimization, including learned branching heuristics~\citep{zhang2024towards}, node selection~\citep{labassi2022learning}, and cutting-plane selection~\citep{wang2023learning}. These methods improve solver efficiency but typically assume that the MIP instance is already specified. Several datasets and environments, such as Ecole~\citep{prouvost2020ecole}, support benchmarking machine-learning methods for combinatorial optimization. However, these benchmarks primarily focus on exact MIP/MILP instances rather than approximate, geometry-specific formulations used in robot motion planning. In contrast, M$^3$P-R1 addresses the problem of generating executable approximate MIP formulations from natural-language task descriptions, where the model must choose and compose discretization primitives rather than merely accelerate an existing solver.

\paragraph{LLM-based planning variants.}
LLMs have been used for planning in several forms. Direct prompting methods generate high-level action sequences from natural-language goals~\citep{ahn2022can,ding2023task,driess2023palm,huang2022language}, but their outputs are often difficult to verify. ReAct-style methods~\citep{yao2022react} interleave reasoning and acting, improving interactive problem solving but still lacking formal guarantees on feasibility or constraint satisfaction. LLM+P frameworks~\citep{liu2023llm+,wang2024llm,silver2024generalized,agarwal2025l3m+} translate language into symbolic planning representations such as PDDL, which are then solved by classical planners. These methods benefit from symbolic structure but are limited by predefined abstractions and require additional mechanisms for continuous feasibility. Code-generation approaches such as Code as Policies~\citep{liang2023code} and ProgPrompt~\citep{singh2022progprompt} produce executable robot programs, but generally output open-loop code rather than solver-verified optimization formulations.

\paragraph{LLM-guided continuous planning.}
A related class of methods uses LLMs to propose waypoints, subgoals, or intermediate states, followed by classical continuous planners such as RRT*~\citep{karaman2011sampling}, BIT*~\citep{gammell2015batch}, or trajectory optimizers such as TrajOpt~\citep{schulman2014motion}. These approaches are useful when a high-level language instruction can be decomposed into waypoint targets, but the symbolic decomposition and continuous optimization are still decoupled. As a result, they can struggle when task success depends on jointly optimizing shared meeting points, contact interfaces, reachability constraints, or synchronized multi-agent states. Our LLM+MIP formulation instead generates a single solver-executable program in which discrete mode choices and continuous trajectories are optimized together.

\paragraph{LLMs for optimization modeling.}
Recent work has also explored using LLMs for optimization tasks, including diagnosing infeasible optimization models~\citep{chen2024diagnosing}, generating supply-chain or operations-research formulations~\citep{li2023large}, evolving heuristics~\citep{liu2024evolution}, translating natural language into MIP scripts~\citep{li2023synthesizing,ahmaditeshnizi2024optimus}, and building foundation models for MILP~\citep{li2025towards}. These works primarily target exact MIP/MILP problems, where the formulation structure is often algebraic and domain-independent. M$^3$P-R1 differs by targeting approximate MIP for robotics, where valid formulations require geometry-specific discretization, collision constraints, reachability approximations, and cross-modal coupling constraints.

\paragraph{Solver-grounded reinforcement fine-tuning.}
Reinforcement fine-tuning has become a common strategy for improving structured reasoning in LLMs. RLHF~\citep{ouyang2022training}, PPO~\citep{schulman2017proximal}, DPO~\citep{rafailov2023direct}, and GRPO~\citep{shao2024deepseekmath} optimize models using outcome- or preference-based rewards. In verifiable domains, external tools can provide stronger supervision: proof assistants such as Lean validate mathematical reasoning~\citep{ren2025deepseek,ying2024lean}, symbolic solvers support geometry reasoning~\citep{Trinh2024SolvingOG,chervonyi2025gold}, and interpreters or compilers verify generated programs~\citep{feng2025retool,li2025start}. Solver-Informed RL~\citep{chen2025solver} further demonstrates the value of solver feedback for optimization modeling. M$^3$P-R1 follows this solver-grounded paradigm, but applies it to language-conditioned robotic M$^3$P, where rewards evaluate not only code execution but also solver feasibility, task satisfaction, and multi-modal trajectory correctness.

\section{LLMM$^3$P Bench Data Example}\label{appen:data_example}
Here we showcase both single- and multi-modal motion planning data examples generated for our dataset.
\begin{codeexample}{UAV Trajectory Planning}{}
Begin the flight mission at the start position $(1.8, 0.3)$. First, navigate to the left side of the bottom obstacle located between $(0.0, -1.5)$ and $(0.9, -0.7)$. Next, approach the top side of the upper obstacle spanning from $(-1.3, 0.0)$ to $(-0.3, 0.7)$. Finally, smoothly proceed to the goal position $(-0.4, 2.0)$ within the bounded workspace $[-3.0, 3.0] \times [-3.0, 3.0]$.

\tcblower

\begin{lstlisting}[style=mipcode]
# Environment and task setup
bounding_lb = [-3.0, -3.0]
bounding_ub = [3.0, 3.0]
start_pos = [1.8, 0.3]
goal_pos = [-0.4, 2.0]

num_segments = 12
num_regions = 32
big_M = 10.0
side_width = 0.2

obstacle_map = {
    "obst1": {
        "lb": [[0.0], [-1.5]],
        "ub": [[0.9], [-0.7]],
        "style": "bottom",
    },
    "obst2": {
        "lb": [[-1.3], [0.0]],
        "ub": [[-0.3], [0.7]],
        "style": "top",
    },
}

# Region generation
endpoints = [start_pos, goal_pos]
iris_regions, obstacles_list_plt, styles, obstacle_keys = generate_iris_regions(
    bounding_lb,
    bounding_ub,
    endpoints,
    num_regions,
    obstacle_map,
)

iris_regions, styles = augment_iris_regions(
    iris_regions,
    obstacle_map,
    styles,
    side_width,
)

# Problem initialization
C, H, var_dict = initialize_trajectory_problem(
    num_segments,
    len(iris_regions),
)

# Constraints
constraints = []
constraints += add_segment_assignment_constraints(H)
constraints += add_region_membership_constraints(C, H, iris_regions, big_M)
constraints += add_spatial_relation_constraints(H, styles)
constraints += add_trajectory_continuity_constraints(C)
constraints += add_trajectory_endpoint_constraints(C, start_pos, goal_pos)

# Objective and solve
objective = add_objective(["trajectory_smoothness"])
status, cost, solution = solve_problem(
    constraints,
    objective,
    var_dict,
)
\end{lstlisting}
\end{codeexample}

\begin{codeexample}{Inverse Kinematics Planning}{}
The mission is for a robotic arm with five links of lengths $0.9$, $1.0$, $0.4$, $0.5$, and $0.6$ m to move from the base position $(-1.6, -1.4)$ to the end-effector target $(0.6, 0.7)$. While navigating, avoid the top obstacle extending from $(-0.1, -0.2)$ to $(0.7, 0.7)$ and the secondary obstacle from $(0.2, -1.3)$ to $(1.2, -0.6)$.

\tcblower

\begin{lstlisting}[style=mipcode]
# Environment and task setup
bounding_lb = [-2.0, -2.0]
bounding_ub = [2.0, 2.0]
base_location = [-1.6, -1.4]
target_location = [0.6, 0.7]

link_lengths = [0.9, 1.0, 0.4, 0.5, 0.6]
n_poly = 8
num_intermediate = 10
num_regions = 20
big_M = 10.0
side_width = 0.2

obstacle_map = {
    "obst1": {
        "lb": [[-0.1], [-0.2]],
        "ub": [[0.7], [0.7]],
        "style": "top",
    },
    "obst2": {
        "lb": [[0.2], [-1.3]],
        "ub": [[1.2], [-0.6]],
        "style": "none",
    },
}

# Region generation
endpoints = [base_location, target_location]
iris_regions, obstacles_list_plt, styles, obstacle_keys = generate_iris_regions(
    bounding_lb,
    bounding_ub,
    endpoints,
    num_regions,
    obstacle_map,
)

iris_regions, styles = augment_iris_regions(
    iris_regions,
    obstacle_map,
    styles,
    side_width,
)

# Problem initialization
pivots, rotations, collision_points, var_dict = initialize_kinematic_problem(
    link_lengths,
    n_poly,
    num_intermediate,
    base_location,
)

# Constraints
constraints = []
constraints += add_base_target_constraints(
    pivots[0],
    pivots[-1],
    base_location,
    target_location,
)
constraints += add_kinematic_constraints(
    pivots,
    rotations,
    link_lengths,
)
constraints += add_collision_free_constraints(
    collision_points,
    iris_regions,
)
constraints += add_spatial_relation_constraints(
    var_dict["H"],
    styles,
)

# Objective and solve
objective = add_objective(["feasibility"])
status, cost, solution = solve_problem(
    constraints,
    objective,
    var_dict,
)
\end{lstlisting}
\end{codeexample}

\begin{codeexample}{Footstep Planning}{}
Plan a biped locomotion path from the initial left foot position $(0.0, 0.0)$ and right foot position $(0.5, 0.0)$ to the goal position $(14.0, 14.0)$ while avoiding obstacles in the workspace $[-5.0, 15.0] \times [-5.0, 15.0]$.

\tcblower

\begin{lstlisting}[style=mipcode]
# Environment and task setup
bounding_lb = [-5.0, -5.0]
bounding_ub = [15.0, 15.0]

start_left = [0.0, 0.0]
start_right = [0.5, 0.0]
goal_pos = [14.0, 14.0]

steps = 15
step_dist = 1.5
num_regions = 12
big_M = 10.0

obstacle_map = {
    "obst1": {
        "lb": [-0.2, 11.3],
        "ub": [2.1, 13.1],
        "style": "none",
    },
    "obst2": {
        "lb": [2.9, 1.3],
        "ub": [5.4, 3.1],
        "style": "none",
    },
    "obst3": {
        "lb": [11.0, -1.9],
        "ub": [12.7, -0.4],
        "style": "none",
    },
}

# Region generation
endpoints = [start_left, start_right, goal_pos]
iris_regions, obstacles_list_plt, styles, obstacle_keys = generate_iris_regions(
    bounding_lb,
    bounding_ub,
    endpoints,
    num_regions,
    obstacle_map,
)

# Problem initialization
positions, region_assignments, var_dict = initialize_footstep_problem(
    len(iris_regions),
    steps,
)

# Constraints
constraints = []
constraints += add_region_membership_constraints(
    positions,
    region_assignments,
    iris_regions,
    big_M,
)
constraints += add_reachability_constraints(positions)
constraints += add_trajectory_endpoint_constraints(
    positions,
    [start_left, start_right],
    goal_pos,
)

# Objective and solve
objective = add_objective(["feasibility"])
status, cost, solution = solve_problem(
    constraints,
    objective,
    var_dict,
)
\end{lstlisting}
\end{codeexample}

\begin{codeexample}{Finger Selection}{}
To stably grasp an object with four fingers, place each finger at well-separated contact points around the object perimeter so that the contacts distribute force symmetrically and improve grasp robustness. The friction coefficient is $\mu=1.0$.

\tcblower

\begin{lstlisting}[style=mipcode]
# Object geometry and task setup
vertices_list = [
    [-0.3, 0.9],
    [-1.3, -1.0],
    [0.9, -1.0],
    [1.2, -0.3],
    [0.4, 0.3],
]

n_fingers = 4
friction_coef = 1.0
delta = 0.1
delta_n = 0.0

# Object construction and contact sampling
obj = build_finger_select_object(vertices_list)
sampled_points, sampled_normals = sample_finger_contact_candidates(
    obj,
    delta,
    delta_n,
)

# Problem initialization
selection, radius, var_dict = initialize_finger_selection_problem(
    n_fingers,
    len(sampled_points),
)

# Constraints
constraints = []
constraints += add_contact_selection_constraints(selection)
constraints += add_wrench_constraints(
    sampled_points,
    sampled_normals,
    friction_coef,
)

# Objective and solve
objective = add_objective(["feasibility"])
status, cost, solution = solve_problem(
    constraints,
    objective,
    var_dict,
)
\end{lstlisting}
\end{codeexample}

\begin{codeexample}{Multi-UAV Planning}{}
Two UAVs begin from different locations in the workspace. UAV~1 starts at $(2.4, -1.8)$ and should first pass around the right side of the lower obstacle before moving through the open corridor. UAV~2 starts at $(-2.3, 1.7)$ and should pass around the left side of the upper obstacle before entering the same open corridor. The two UAVs must complete a rendezvous task by arriving at the same collision-free meeting location while avoiding all obstacles and maintaining smooth trajectories within the bounded workspace $[-3.0, 3.0] \times [-3.0, 3.0]$.

\tcblower

\begin{lstlisting}[style=mipcode]
# Environment and task setup
bounding_lb = [-3.0, -3.0]
bounding_ub = [3.0, 3.0]

start_pos_uav1 = [2.4, -1.8]
start_pos_uav2 = [-2.3, 1.7]
shared_meet_pos = [0.3, 0.2]

goal_pos_uav1 = shared_meet_pos
goal_pos_uav2 = shared_meet_pos

num_segments = 12
num_regions = 32
big_M = 10.0
side_width = 0.2

obstacle_map = {
    "obst1": {
        "lb": [[0.5], [-2.0]],
        "ub": [[1.4], [-0.8]],
        "style": "right",
    },
    "obst2": {
        "lb": [[-1.5], [0.7]],
        "ub": [[-0.5], [1.8]],
        "style": "left",
    },
    "obst3": {
        "lb": [[-0.4], [-0.6]],
        "ub": [[0.6], [0.0]],
        "style": "none",
    },
}

# ------------------------------------------------------------
# UAV 1: single-UAV planning sequence
# ------------------------------------------------------------
endpoints_uav1 = [start_pos_uav1, goal_pos_uav1]
iris_regions_uav1, obstacles_list_plt_uav1, styles_uav1, obstacle_keys_uav1 = generate_iris_regions(
    bounding_lb,
    bounding_ub,
    endpoints_uav1,
    num_regions,
    obstacle_map,
)

iris_regions_uav1, styles_uav1 = augment_iris_regions(
    iris_regions_uav1,
    obstacle_map,
    styles_uav1,
    side_width,
)

C_uav1, H_uav1, var_dict_uav1 = initialize_trajectory_problem(
    num_segments,
    len(iris_regions_uav1),
)

constraints_uav1 = []
constraints_uav1 += add_segment_assignment_constraints(H_uav1)
constraints_uav1 += add_region_membership_constraints(
    C_uav1,
    H_uav1,
    iris_regions_uav1,
    big_M,
)
constraints_uav1 += add_spatial_relation_constraints(H_uav1, styles_uav1)
constraints_uav1 += add_trajectory_continuity_constraints(C_uav1)
constraints_uav1 += add_trajectory_endpoint_constraints(
    C_uav1,
    start_pos_uav1,
    goal_pos_uav1,
)
objective_uav1 = add_objective(["trajectory_smoothness"])
status, cost, solution = solve_problem(
    constraints_uav1,
    objective_uav1,
    var_dict_uav1,
)
# ------------------------------------------------------------
# UAV 2: single-UAV planning sequence
# ------------------------------------------------------------
endpoints_uav2 = [start_pos_uav2, goal_pos_uav2]
iris_regions_uav2, obstacles_list_plt_uav2, styles_uav2, obstacle_keys_uav2 = generate_iris_regions(
    bounding_lb,
    bounding_ub,
    endpoints_uav2,
    num_regions,
    obstacle_map,
)

iris_regions_uav2, styles_uav2 = augment_iris_regions(
    iris_regions_uav2,
    obstacle_map,
    styles_uav2,
    side_width,
)

C_uav2, H_uav2, var_dict_uav2 = initialize_trajectory_problem(
    num_segments,
    len(iris_regions_uav2),
)

constraints_uav2 = []
constraints_uav2 += add_segment_assignment_constraints(H_uav2)
constraints_uav2 += add_region_membership_constraints(
    C_uav2,
    H_uav2,
    iris_regions_uav2,
    big_M,
)
constraints_uav2 += add_spatial_relation_constraints(H_uav2, styles_uav2)
constraints_uav2 += add_trajectory_continuity_constraints(C_uav2)
constraints_uav2 += add_trajectory_endpoint_constraints(
    C_uav2,
    start_pos_uav2,
    goal_pos_uav2,
)

objective_uav2 = add_objective(["trajectory_smoothness"])
status, cost, solution = solve_problem(
    constraints_uav2,
    objective_uav2,
    var_dict_uav2,
)
\end{lstlisting}
\end{codeexample}

\begin{codeexample}{UAV--IK Planning}{}
A UAV starts from $(2.2, -2.2)$ and flies through a cluttered workspace to support a downstream manipulation task. It should pass around the right side of the lower obstacle and avoid the central obstacle before reaching a collision-free handoff location. From this handoff location, a robotic arm with four links must reach the manipulation target while keeping its links collision-free. The complete plan must jointly coordinate the UAV trajectory and the arm configuration within the bounded workspace $[-3.0, 3.0] \times [-3.0, 3.0]$.

\tcblower

\begin{lstlisting}[style=mipcode]
# Environment and task setup
bounding_lb = [-3.0, -3.0]
bounding_ub = [3.0, 3.0]

uav_start = [2.2, -2.2]
handoff_seed = [0.2, 0.4]
ik_target = [-1.2, 1.5]

num_segments = 12
num_regions = 30
big_M = 10.0
side_width = 0.2

link_lengths = [0.8, 0.7, 0.5, 0.4]
n_poly = 8
num_intermediate = 10

obstacle_map = {
    "obst1": {
        "lb": [[0.6], [-1.8]],
        "ub": [[1.4], [-0.8]],
        "style": "right",
    },
    "obst2": {
        "lb": [[-0.5], [-0.4]],
        "ub": [[0.5], [0.5]],
        "style": "none",
    },
    "obst3": {
        "lb": [[-1.8], [0.8]],
        "ub": [[-0.8], [1.6]],
        "style": "none",
    },
}

shared_meet_pos=[0.3,0.2]

# Region generation
endpoints = [uav_start, handoff_seed, ik_target]
iris_regions, obstacles_list_plt, styles, obstacle_keys = generate_iris_regions(
    bounding_lb,
    bounding_ub,
    endpoints,
    num_regions,
    obstacle_map,
)

iris_regions, styles = augment_iris_regions(
    iris_regions,
    obstacle_map,
    styles,
    side_width,
)

# ------------------------------------------------------------
# UAV trajectory planning
# ------------------------------------------------------------
C_uav, H_uav, var_dict_uav = initialize_trajectory_problem(
    num_segments,
    len(iris_regions),
)

constraints_uav = []
constraints_uav += add_segment_assignment_constraints(H_uav)
constraints_uav += add_region_membership_constraints(
    C_uav,
    H_uav,
    iris_regions,
    big_M,
)
constraints_uav += add_spatial_relation_constraints(H_uav, styles)
constraints_uav += add_trajectory_continuity_constraints(C_uav)
constraints_uav += add_trajectory_endpoint_constraints(
    C_uav,
    uav_start,
    shared_meet_pos,
)

objective_uav = add_objective(["trajectory_smoothness"])
status, cost, solution = solve_problem(
    constraints_uav,
    objective_uav,
    var_dict_uav,
)

# ------------------------------------------------------------
# IK planning from the shared UAV handoff location
# ------------------------------------------------------------
pivots, rotations, collision_points, var_dict_ik = initialize_kinematic_problem(
    link_lengths,
    n_poly,
    num_intermediate,
    shared_meet_pos,
)
constraints_ik=[]
constraints_ik += add_base_target_constraints(
    pivots[0],
    pivots[-1],
    shared_meet_pos,
    ik_target,
)
constraints_ik += add_kinematic_constraints(
    pivots,
    rotations,
    link_lengths,
)
constraints_ik += add_collision_free_constraints(
    collision_points,
    iris_regions,
)

# Objective and solve
objective = add_objective([
    "feasibility",
])

status, cost, solution = solve_problem(
    constraints_ik,
    objective,
    var_dict_ik,
)
\end{lstlisting}
\end{codeexample}

\begin{codeexample}{IK--Footstep Planning}{}
A biped robot starts with its left foot at $(-3.0, -2.6)$ and right foot at $(-2.5, -2.6)$. It must walk through the workspace while avoiding obstacles and stop at a feasible manipulation stance. From this stance, its arm with four links must reach a target point near the upper-right workspace region while keeping the articulated chain collision-free. The locomotion and arm motion must be coordinated through a shared base location within the bounded workspace $[-4.0, 4.0] \times [-4.0, 4.0]$.

\tcblower

\begin{lstlisting}[style=mipcode]
# Environment and task setup
bounding_lb = [-4.0, -4.0]
bounding_ub = [4.0, 4.0]

start_left = [-3.0, -2.6]
start_right = [-2.5, -2.6]
ik_target = [1.3, 1.1]

steps = 12
step_dist = 1.0
num_regions = 24
big_M = 10.0
side_width = 0.2

link_lengths = [0.8, 0.7, 0.5, 0.4]
n_poly = 8
num_intermediate = 10

obstacle_map = {
    "obst1": {
        "lb": [-1.8, -1.4],
        "ub": [-0.7, -0.3],
        "style": "none",
    },
    "obst2": {
        "lb": [0.5, -1.2],
        "ub": [1.6, -0.2],
        "style": "none",
    },
    "obst3": {
        "lb": [-0.4, 0.6],
        "ub": [0.6, 1.5],
        "style": "none",
    },
}

shared_base = [0.3, 0.2]

# Region generation
endpoints = [start_left, start_right, shared_base, ik_target]
iris_regions, obstacles_list_plt, styles, obstacle_keys = generate_iris_regions(
    bounding_lb,
    bounding_ub,
    endpoints,
    num_regions,
    obstacle_map,
)

iris_regions, styles = augment_iris_regions(
    iris_regions,
    obstacle_map,
    styles,
    side_width,
)

# ------------------------------------------------------------
# Footstep planning to the shared manipulation stance
# ------------------------------------------------------------
foot_positions, H_foot, var_dict_foot = initialize_footstep_problem(
    len(iris_regions),
    steps,
)

constraints = []
constraints += add_region_membership_constraints(
    foot_positions,
    H_foot,
    iris_regions,
    big_M,
)
constraints += add_reachability_constraints(foot_positions)
constraints += add_trajectory_endpoint_constraints(
    foot_positions,
    [start_left, start_right],
    shared_base,
)

# Objective and solve
objective = add_objective(["feasibility"])
status, cost, solution = solve_problem(
    constraints,
    objective,
    var_dict_foot,
)

# ------------------------------------------------------------
# IK planning from the shared stance
# ------------------------------------------------------------
pivots, rotations, collision_points, var_dict_ik = initialize_kinematic_problem(
    link_lengths,
    n_poly,
    num_intermediate,
    shared_base,
)
constraints_ik=[]
constraints_ik += add_base_target_constraints(
    pivots[0],
    pivots[-1],
    shared_base,
    ik_target,
)
constraints_ik += add_kinematic_constraints(
    pivots,
    rotations,
    link_lengths,
)
constraints_ik += add_collision_free_constraints(
    collision_points,
    iris_regions,
)

# Objective and solve
objective = add_objective(["feasibility"])

status, cost, solution = solve_problem(
    constraints_ik,
    objective,
    var_dict_ik,
)
\end{lstlisting}
\end{codeexample}

\section{LLM Prompt}\label{appen:llm_prompt}
This section presents the detailed system prompt used in our method, together with a representative model output. In the multi-UAV example, the fine-tuned LLM directly constructs decision variables and constraints for the shared rendezvous point, rather than decomposing the task into two independent single-UAV problems with a manually predicted meeting location. This demonstrates that the model learns to discretize multimodal robot planning tasks into solver-grounded MIP formulations.

\begin{codeexample}{LLM System Prompt}{}
You are an expert in multi-modal mixed-integer programming (MIP) for robotic planning.
Read the natural-language mission instruction, the optimal \texttt{OBJECT\_GEOMETRY} block, and the \texttt{ENVIRONMENT} block.
Identify the correct modality and write a runnable Python program that solves the problem.
You are provided with an API handbook containing MIP discretization functions.
You may use the functions listed in the API handbook below, as well as functions from the \texttt{cvxpy} package.
Note that \texttt{obstacle\_map} entries use the standard axis-aligned box form:
\texttt{\{"lb": [...], "ub": [...], "style": ...\}}.

Your output must contain a \texttt{<think>...</think>} block that concisely explains the planning rationale, followed by a \texttt{<code>...</code>} block containing the complete executable Python code.

Follow the structure below:
\begin{enumerate}
    \item build object geometry or safe regions when the modality requires them,
    \item initialize decision variables,
    \item add constraints stage by stage,
    \item call the modality solver.
\end{enumerate}

Keep all arguments explicit, and write the complete executable Python code in the correct format.
\end{codeexample}

\begin{codeexample}{Multi-UAV Rendezvous Planning}{}
Two UAVs operate in the bounded workspace $[-3.0, 3.0] \times [-3.0, 3.0]$.
UAV~1 starts at $(-2.0, -2.0)$ and UAV~2 starts at $(2.0, 2.0)$.
Because of communication-range limits, neither UAV should travel all the way to the other station.
Instead, they must find an intermediate collision-free shared meeting point and plan smooth trajectories to this rendezvous location.
The environment contains one rectangular obstacle with lower bound $(0.223, 0.729)$ and upper bound $(0.765, 1.377)$, and the route should approach this obstacle from the left side before continuing toward the rendezvous.

\tcblower

\begin{lstlisting}[style=mipcode]
<think>
This is a two-UAV rendezvous problem. I need one shared meeting-point variable, then each UAV should plan from its own station to that same point. I will generate shared safe regions, build one trajectory problem per UAV, enforce region assignment, obstacle avoidance, spatial style constraints, continuity, and start-goal constraints. Finally, I will explicitly constrain the shared meeting point to lie in one collision-free safe region.
</think>

<code>

import cvxpy as cp
import numpy as np

num_uavs = 2
num_segments_list = [12, 12]

bounding_lb = np.array([-3.0, -3.0], dtype=float)
bounding_ub = np.array([3.0, 3.0], dtype=float)

start_1 = np.array([-2.0, -2.0], dtype=float)
goal_last = np.array([2.0, 2.0], dtype=float)

num_regions = 24
big_M = 10.0
side_width = 0.2

obstacle_map = {
    "obst1": {
        "lb": np.array([[0.223], [0.729]], dtype=float),
        "ub": np.array([[0.765], [1.377]], dtype=float),
        "style": "left",
    },
}

# ------------------------------------------------------------
# Shared safe-region generation
# ------------------------------------------------------------
safe_regions, obstacles_list_plt, styles, obstacle_keys = generate_iris_regions(bounding_lb,bounding_ub,start_1,goal_last,
num_regions,obstacle_map)

safe_regions, styles_related_regions = augment_iris_regions(safe_regions,obstacle_map,styles,side_width)

num_safe_regions = len(safe_regions)

# ------------------------------------------------------------
# Multi-UAV trajectory variables
# ------------------------------------------------------------
C_list, H_list, var_dict = initialize_trajectory_problem(num_segments_list,num_safe_regions)

# ------------------------------------------------------------
# LLM generated shared point constraints
# ------------------------------------------------------------
shared_points_list = []
for i in range(max(0, num_uavs - 1)):
    shared_point = cp.Variable(2, name=f"shared_points_{i}")
    var_dict[f"shared_points_{i}"] = shared_point
    shared_points_list.append(shared_point)

# ------------------------------------------------------------
# Per-UAV planning constraints
# ------------------------------------------------------------
constraints = []

for uav_idx in range(num_uavs):
    C = C_list[uav_idx]
    H = H_list[uav_idx]
    num_segments = num_segments_list[uav_idx]

    constraints += add_segment_assignment_constraints(H)

    constraints += add_spatial_relation_constraints(H,styles)

    constraints += add_region_membership_constraints(C,H,safe_regions,big_M=big_M)

    constraints += add_trajectory_continuity_constraints(C)

    constraints += add_trajectory_endpoint_constraints(C,start_pos,goal_pos)

# ------------------------------------------------------------
# Shared rendezvous feasibility constraints
# ------------------------------------------------------------
for sp_idx, shared_point in enumerate(shared_points_list):
    num_r = len(safe_regions)

    sp_z = cp.Variable(num_r, boolean=True, name=f"z_shared_point_{sp_idx}")
    var_dict[f"z_shared_point_{sp_idx}"] = sp_z

    sp_y_vars = [
        cp.Variable(2, name=f"y_shared_point_{sp_idx}_{r}")
        for r in range(num_r)
    ]

    for r in range(num_r):
        var_dict[f"y_shared_point_{sp_idx}_{r}"] = sp_y_vars[r]

    constraints += [
        shared_point[0] == sum(y[0] for y in sp_y_vars),
        shared_point[1] == sum(y[1] for y in sp_y_vars),
        cp.sum(sp_z) == 1,
    ]

    for r in range(num_r):
        A_r = safe_regions[r].A()
        b_r = safe_regions[r].b().flatten()

        for j in range(A_r.shape[0]):
            constraints.append(
                A_r[j, 0] * sp_y_vars[r][0]
                + A_r[j, 1] * sp_y_vars[r][1]
                <= b_r[j] * sp_z[r]
            )
objective = add_objective(["trajectory_smoothness"])
# ------------------------------------------------------------
# Solve
# ------------------------------------------------------------
status, cost, H_vals, C_vals = solve_problem(constraints,objective,var_dict)
</code>
\end{lstlisting}
\end{codeexample}

\end{document}